\documentclass{article}

\usepackage[final]{preamble/neurips_2021}

\usepackage{times}					%
\usepackage[utf8]{inputenc} 		%
\usepackage[T1]{fontenc}    		%
\usepackage[english]{babel} 		%
\usepackage{xargs}                  %

\usepackage{booktabs}       %
\usepackage{tabularx}		%

\usepackage{graphicx} 		%
\usepackage{fancybox}		%
\usepackage{xcolor} 		%
\usepackage{epsfig}
\usepackage{pifont}			%
\usepackage{caption}
\usepackage{subcaption}

\usepackage{tikz}			%
\usepackage{pgfplots}
\pgfplotsset{compat=1.15}	%
\usetikzlibrary{shapes,
                pgfplots.groupplots,
                shadings,
                calc,
                arrows,
                backgrounds,
                colorbrewer,
                shadows.blur}

\makeatletter
\newcommand\Cshadowbox{\VerbBox\@Cshadowbox}
\def\@Cshadowbox#1{%
	\setbox\@fancybox\hbox{\fbox{#1}}%
	\leavevmode\vbox{%
		\offinterlineskip
		\dimen@=\shadowsize
		\advance\dimen@ .5\fboxrule
		\hbox{\copy\@fancybox\kern.5\fboxrule\lower\shadowsize\hbox{%
				\color{ShadowColor}\vrule \@height\ht\@fancybox \@depth\dp\@fancybox \@width\dimen@}}%
		\vskip\dimexpr-\dimen@+0.5\fboxrule\relax
		\moveright\shadowsize\vbox{%
			\color{ShadowColor}\hrule \@width\wd\@fancybox \@height\dimen@}}}
\makeatother

\usepackage[frozencache=true,cachedir=minted-cache]{minted}			%
\usepackage{amsfonts}       %
\usepackage{amsmath} 		%
\usepackage{amssymb}
\usepackage{mathtools} 		%
\usepackage{nicefrac}       %

\usepackage{xspace} 		%
\usepackage{microtype} 		%
\usepackage{url} 			%
\usepackage[pagebackref=true,breaklinks=true,colorlinks,bookmarks=false]{hyperref}
\hypersetup{
	citecolor = sns_orange,
	linkcolor = sns_blue,
	urlcolor = sns_blue
}
\usepackage[]{cleveref}
\setcitestyle{numbers,square} %

\usepackage[colorinlistoftodos,prependcaption,textsize=tiny, shadow]{todonotes} 						%
\newcommandx{\unsure}[1]{\todo[linecolor=red,backgroundcolor=red!25,bordercolor=red]{#1}}				%
\newcommandx{\change}[1]{\todo[linecolor=blue,backgroundcolor=blue!25,bordercolor=blue]{#1}}			%
\newcommandx{\note}[1]{\todo[linecolor=orange,backgroundcolor=orange!25,bordercolor=orange]{#1}}		%
\newcommandx{\improvement}[1]{\todo[linecolor=violet,backgroundcolor=violet!25,bordercolor=violet]{#1}}	%

\newcommand{\PhD}{PhD\xspace}
\newcommand*{\eg}{e.g.\@\xspace}
\newcommand*{\citeeg}{e.g.}		%
\newcommand*{\ie}{i.e.\@\xspace}
\newcommand*{\cf}{cf.\@\xspace}
\newcommand*{\etal}{et. al.\@\xspace}
\makeatletter\newcommand*{\etc}{%
	\@ifnextchar{.}%
	{etc}%
	{etc.\@\xspace}%
}
\makeatother

\definecolor{TUred}{RGB}{141,45,57}
\definecolor{TUdark}{RGB}{55,65,74}
\definecolor{TUgold}{RGB}{174,159,109}
\definecolor{TUgray}{RGB}{175,179,183}
\definecolor{ERC_ora}{RGB}{233,93,15}

\definecolor{sns_orange}{HTML}{ff7f0e}
\definecolor{sns_green}{rgb}{0.17254901960784313, 0.6274509803921569, 0.17254901960784313}

\definecolor{sns_blue}{HTML}{1f77b4}
\definecolor{sns_orange_light}{HTML}{FFE5CE}

\definecolor{ShadowColor}{RGB}{175,179,183}

\newcommand{\cockpiturl}{\url{https://github.com/f-dangel/cockpit}\xspace}
\newcommand{\cockpitexpurl}{\url{https://github.com/fsschneider/cockpit-experiments}\xspace}

\newcommand{\cockpit}{\mbox{\normalfont\textsc{Cockpit}}\xspace}
\newcommand{\cockpittitle}{\textsc{Cockpit}\xspace} %
\newcommand{\deepobs}{\mbox{\normalfont\textsc{DeepOBS}}\xspace}
\newcommand{\backpack}{\mbox{\normalfont\textsc{BackPACK}}\xspace}
\newcommand{\tensorflow}{\mbox{\normalfont\textsc{TensorFlow}}\xspace}
\newcommand{\tensorflowdataset}{\mbox{\normalfont\textsc{TensorFlow Datasets}}\xspace}
\newcommand{\tensorboard}{\mbox{\normalfont\textsc{TensorBoard}}\xspace}
\newcommand{\pytorch}{\mbox{\normalfont\textsc{PyTorch}}\xspace}
\newcommand{\torchvision}{\mbox{\normalfont\textsc{torchvision}}\xspace}
\newcommand{\wandb}{\mbox{\normalfont\textsc{Weights \& Biases}}\xspace}
\newcommand{\numpy}{\mbox{\normalfont\textsc{NumPy}}\xspace}
\newcommand{\matplotlib}{\mbox{\normalfont\textsc{matplotlib}}\xspace}

\newcommand{\sgd}{\textsc{SGD}\xspace}
\newcommand{\gd}{\textsc{GD}\xspace}
\newcommand{\ggn}{\textsc{GGN}\xspace}
\newcommand{\adam}{\textsc{Adam}\xspace}
\newcommand{\adagrad}{\textsc{AdaGrad}\xspace}
\newcommand{\cabs}{\textsc{CABS}\xspace}

\newcommand{\cifarten}{\textsc{CIFAR-10}\xspace}
\newcommand{\cifarhun}{\textsc{CIFAR-100}\xspace}
\newcommand{\cifartenhun}{\textsc{CIFAR-10/100}\xspace}
\newcommand{\mnist}{\textsc{MNIST}\xspace}
\newcommand{\fmnist}{\textsc{Fashion-MNIST}\xspace}
\newcommand{\imagenet}{\textsc{ImageNet}\xspace}
\newcommand{\svhn}{\textsc{SVHN}\xspace}

\newcommand{\allcnnc}{\textsc{All-CNN-C}\xspace}
\newcommand{\threecthreed}{\textsc{3c3d}\xspace}
\newcommand{\twoctwod}{\textsc{2c2d}\xspace}
\newcommand{\lstm}{\textsc{LSTM}\xspace}
\newcommand{\mlp}{\textsc{MLP}\xspace}
\newcommand{\resnetfifty}{\textsc{ResNet-50}\xspace}
\newcommand{\vgg}{\textsc{VGG16}\xspace}

\newcommand{\gsnr}{\textsc{GSNR}\xspace}

\newcommand{\appendixtitle}{
	\vbox{
		{\hrule height 4pt \vskip 0.25in \vskip -\parskip}
		\centering
		{\LARGE\bf \cockpittitle: A Practical Debugging Tool\\for the Training of Deep Neural Networks\par}
		{\Large\bf Supplementary Material\par}
		{\vskip 0.29in \vskip -\parskip \hrule height 1pt \vskip 0.09in}
	}
} %

\usepackage{amsmath,amsfonts,bm}

\newcommand{\figleft}{{\em (Left)}}
\newcommand{\figcenter}{{\em (Center)}}
\newcommand{\figright}{{\em (Right)}}
\newcommand{\figtop}{{\em (Top)}}
\newcommand{\figbottom}{{\em (Bottom)}}
\newcommand{\captiona}{{\em (a)}}
\newcommand{\captionb}{{\em (b)}}
\newcommand{\captionc}{{\em (c)}}
\newcommand{\captiond}{{\em (d)}}

\newcommand{\newterm}[1]{{\bf #1}}

\def\figref#1{figure~\ref{#1}}
\def\Figref#1{Figure~\ref{#1}}
\def\twofigref#1#2{figures \ref{#1} and \ref{#2}}
\def\quadfigref#1#2#3#4{figures \ref{#1}, \ref{#2}, \ref{#3} and \ref{#4}}
\def\secref#1{section~\ref{#1}}
\def\Secref#1{Section~\ref{#1}}
\def\twosecrefs#1#2{sections \ref{#1} and \ref{#2}}
\def\secrefs#1#2#3{sections \ref{#1}, \ref{#2} and \ref{#3}}
\def\eqref#1{equation~\ref{#1}}
\def\Eqref#1{Equation~\ref{#1}}
\def\plaineqref#1{\ref{#1}}
\def\chapref#1{chapter~\ref{#1}}
\def\Chapref#1{Chapter~\ref{#1}}
\def\rangechapref#1#2{chapters\ref{#1}--\ref{#2}}
\def\algref#1{algorithm~\ref{#1}}
\def\Algref#1{Algorithm~\ref{#1}}
\def\twoalgref#1#2{algorithms \ref{#1} and \ref{#2}}
\def\Twoalgref#1#2{Algorithms \ref{#1} and \ref{#2}}
\def\partref#1{part~\ref{#1}}
\def\Partref#1{Part~\ref{#1}}
\def\twopartref#1#2{parts \ref{#1} and \ref{#2}}

\def\ceil#1{\lceil #1 \rceil}
\def\floor#1{\lfloor #1 \rfloor}
\def\1{\bm{1}}
\newcommand{\train}{\mathcal{D}}
\newcommand{\valid}{\mathcal{D_{\mathrm{valid}}}}
\newcommand{\test}{\mathcal{D_{\mathrm{test}}}}

\def\eps{{\epsilon}}

\def\reta{{\textnormal{$\eta$}}}
\def\ra{{\textnormal{a}}}
\def\rb{{\textnormal{b}}}
\def\rc{{\textnormal{c}}}
\def\rd{{\textnormal{d}}}
\def\re{{\textnormal{e}}}
\def\rf{{\textnormal{f}}}
\def\rg{{\textnormal{g}}}
\def\rh{{\textnormal{h}}}
\def\ri{{\textnormal{i}}}
\def\rj{{\textnormal{j}}}
\def\rk{{\textnormal{k}}}
\def\rl{{\textnormal{l}}}
\def\rn{{\textnormal{n}}}
\def\ro{{\textnormal{o}}}
\def\rp{{\textnormal{p}}}
\def\rq{{\textnormal{q}}}
\def\rr{{\textnormal{r}}}
\def\rs{{\textnormal{s}}}
\def\rt{{\textnormal{t}}}
\def\ru{{\textnormal{u}}}
\def\rv{{\textnormal{v}}}
\def\rw{{\textnormal{w}}}
\def\rx{{\textnormal{x}}}
\def\ry{{\textnormal{y}}}
\def\rz{{\textnormal{z}}}

\def\rvepsilon{{\mathbf{\epsilon}}}
\def\rvtheta{{\mathbf{\theta}}}
\def\rva{{\mathbf{a}}}
\def\rvb{{\mathbf{b}}}
\def\rvc{{\mathbf{c}}}
\def\rvd{{\mathbf{d}}}
\def\rve{{\mathbf{e}}}
\def\rvf{{\mathbf{f}}}
\def\rvg{{\mathbf{g}}}
\def\rvh{{\mathbf{h}}}
\def\rvu{{\mathbf{i}}}
\def\rvj{{\mathbf{j}}}
\def\rvk{{\mathbf{k}}}
\def\rvl{{\mathbf{l}}}
\def\rvm{{\mathbf{m}}}
\def\rvn{{\mathbf{n}}}
\def\rvo{{\mathbf{o}}}
\def\rvp{{\mathbf{p}}}
\def\rvq{{\mathbf{q}}}
\def\rvr{{\mathbf{r}}}
\def\rvs{{\mathbf{s}}}
\def\rvt{{\mathbf{t}}}
\def\rvu{{\mathbf{u}}}
\def\rvv{{\mathbf{v}}}
\def\rvw{{\mathbf{w}}}
\def\rvx{{\mathbf{x}}}
\def\rvy{{\mathbf{y}}}
\def\rvz{{\mathbf{z}}}

\def\erva{{\textnormal{a}}}
\def\ervb{{\textnormal{b}}}
\def\ervc{{\textnormal{c}}}
\def\ervd{{\textnormal{d}}}
\def\erve{{\textnormal{e}}}
\def\ervf{{\textnormal{f}}}
\def\ervg{{\textnormal{g}}}
\def\ervh{{\textnormal{h}}}
\def\ervi{{\textnormal{i}}}
\def\ervj{{\textnormal{j}}}
\def\ervk{{\textnormal{k}}}
\def\ervl{{\textnormal{l}}}
\def\ervm{{\textnormal{m}}}
\def\ervn{{\textnormal{n}}}
\def\ervo{{\textnormal{o}}}
\def\ervp{{\textnormal{p}}}
\def\ervq{{\textnormal{q}}}
\def\ervr{{\textnormal{r}}}
\def\ervs{{\textnormal{s}}}
\def\ervt{{\textnormal{t}}}
\def\ervu{{\textnormal{u}}}
\def\ervv{{\textnormal{v}}}
\def\ervw{{\textnormal{w}}}
\def\ervx{{\textnormal{x}}}
\def\ervy{{\textnormal{y}}}
\def\ervz{{\textnormal{z}}}

\def\rmA{{\mathbf{A}}}
\def\rmB{{\mathbf{B}}}
\def\rmC{{\mathbf{C}}}
\def\rmD{{\mathbf{D}}}
\def\rmE{{\mathbf{E}}}
\def\rmF{{\mathbf{F}}}
\def\rmG{{\mathbf{G}}}
\def\rmH{{\mathbf{H}}}
\def\rmI{{\mathbf{I}}}
\def\rmJ{{\mathbf{J}}}
\def\rmK{{\mathbf{K}}}
\def\rmL{{\mathbf{L}}}
\def\rmM{{\mathbf{M}}}
\def\rmN{{\mathbf{N}}}
\def\rmO{{\mathbf{O}}}
\def\rmP{{\mathbf{P}}}
\def\rmQ{{\mathbf{Q}}}
\def\rmR{{\mathbf{R}}}
\def\rmS{{\mathbf{S}}}
\def\rmT{{\mathbf{T}}}
\def\rmU{{\mathbf{U}}}
\def\rmV{{\mathbf{V}}}
\def\rmW{{\mathbf{W}}}
\def\rmX{{\mathbf{X}}}
\def\rmY{{\mathbf{Y}}}
\def\rmZ{{\mathbf{Z}}}

\def\ermA{{\textnormal{A}}}
\def\ermB{{\textnormal{B}}}
\def\ermC{{\textnormal{C}}}
\def\ermD{{\textnormal{D}}}
\def\ermE{{\textnormal{E}}}
\def\ermF{{\textnormal{F}}}
\def\ermG{{\textnormal{G}}}
\def\ermH{{\textnormal{H}}}
\def\ermI{{\textnormal{I}}}
\def\ermJ{{\textnormal{J}}}
\def\ermK{{\textnormal{K}}}
\def\ermL{{\textnormal{L}}}
\def\ermM{{\textnormal{M}}}
\def\ermN{{\textnormal{N}}}
\def\ermO{{\textnormal{O}}}
\def\ermP{{\textnormal{P}}}
\def\ermQ{{\textnormal{Q}}}
\def\ermR{{\textnormal{R}}}
\def\ermS{{\textnormal{S}}}
\def\ermT{{\textnormal{T}}}
\def\ermU{{\textnormal{U}}}
\def\ermV{{\textnormal{V}}}
\def\ermW{{\textnormal{W}}}
\def\ermX{{\textnormal{X}}}
\def\ermY{{\textnormal{Y}}}
\def\ermZ{{\textnormal{Z}}}

\def\vzero{{\bm{0}}}
\def\vone{{\bm{1}}}
\def\vmu{{\bm{\mu}}}
\def\vnu{{\bm{\nu}}}
\def\vtheta{{\bm{\theta}}}
\def\va{{\bm{a}}}
\def\vb{{\bm{b}}}
\def\vc{{\bm{c}}}
\def\vd{{\bm{d}}}
\def\ve{{\bm{e}}}
\def\vf{{\bm{f}}}
\def\vg{{\bm{g}}}
\def\vh{{\bm{h}}}
\def\vi{{\bm{i}}}
\def\vj{{\bm{j}}}
\def\vk{{\bm{k}}}
\def\vl{{\bm{l}}}
\def\vm{{\bm{m}}}
\def\vn{{\bm{n}}}
\def\vo{{\bm{o}}}
\def\vp{{\bm{p}}}
\def\vq{{\bm{q}}}
\def\vr{{\bm{r}}}
\def\vs{{\bm{s}}}
\def\vt{{\bm{t}}}
\def\vu{{\bm{u}}}
\def\vv{{\bm{v}}}
\def\vw{{\bm{w}}}
\def\vx{{\bm{x}}}
\def\vy{{\bm{y}}}
\def\vz{{\bm{z}}}

\def\evalpha{{\alpha}}
\def\evbeta{{\beta}}
\def\evepsilon{{\epsilon}}
\def\evlambda{{\lambda}}
\def\evomega{{\omega}}
\def\evmu{{\mu}}
\def\evpsi{{\psi}}
\def\evsigma{{\sigma}}
\def\evtheta{{\theta}}
\def\eva{{a}}
\def\evb{{b}}
\def\evc{{c}}
\def\evd{{d}}
\def\eve{{e}}
\def\evf{{f}}
\def\evg{{g}}
\def\evh{{h}}
\def\evi{{i}}
\def\evj{{j}}
\def\evk{{k}}
\def\evl{{l}}
\def\evm{{m}}
\def\evn{{n}}
\def\evo{{o}}
\def\evp{{p}}
\def\evq{{q}}
\def\evr{{r}}
\def\evs{{s}}
\def\evt{{t}}
\def\evu{{u}}
\def\evv{{v}}
\def\evw{{w}}
\def\evx{{x}}
\def\evy{{y}}
\def\evz{{z}}

\def\mA{{\bm{A}}}
\def\mB{{\bm{B}}}
\def\mC{{\bm{C}}}
\def\mD{{\bm{D}}}
\def\mE{{\bm{E}}}
\def\mF{{\bm{F}}}
\def\mG{{\bm{G}}}
\def\mH{{\bm{H}}}
\def\mI{{\bm{I}}}
\def\mJ{{\bm{J}}}
\def\mK{{\bm{K}}}
\def\mL{{\bm{L}}}
\def\mM{{\bm{M}}}
\def\mN{{\bm{N}}}
\def\mO{{\bm{O}}}
\def\mP{{\bm{P}}}
\def\mQ{{\bm{Q}}}
\def\mR{{\bm{R}}}
\def\mS{{\bm{S}}}
\def\mT{{\bm{T}}}
\def\mU{{\bm{U}}}
\def\mV{{\bm{V}}}
\def\mW{{\bm{W}}}
\def\mX{{\bm{X}}}
\def\mY{{\bm{Y}}}
\def\mZ{{\bm{Z}}}
\def\mBeta{{\bm{\beta}}}
\def\mPhi{{\bm{\Phi}}}
\def\mLambda{{\bm{\Lambda}}}
\def\mSigma{{\bm{\Sigma}}}

\DeclareMathAlphabet{\mathsfit}{\encodingdefault}{\sfdefault}{m}{sl}
\SetMathAlphabet{\mathsfit}{bold}{\encodingdefault}{\sfdefault}{bx}{n}
\newcommand{\tens}[1]{\bm{\mathsfit{#1}}}
\def\tA{{\tens{A}}}
\def\tB{{\tens{B}}}
\def\tC{{\tens{C}}}
\def\tD{{\tens{D}}}
\def\tE{{\tens{E}}}
\def\tF{{\tens{F}}}
\def\tG{{\tens{G}}}
\def\tH{{\tens{H}}}
\def\tI{{\tens{I}}}
\def\tJ{{\tens{J}}}
\def\tK{{\tens{K}}}
\def\tL{{\tens{L}}}
\def\tM{{\tens{M}}}
\def\tN{{\tens{N}}}
\def\tO{{\tens{O}}}
\def\tP{{\tens{P}}}
\def\tQ{{\tens{Q}}}
\def\tR{{\tens{R}}}
\def\tS{{\tens{S}}}
\def\tT{{\tens{T}}}
\def\tU{{\tens{U}}}
\def\tV{{\tens{V}}}
\def\tW{{\tens{W}}}
\def\tX{{\tens{X}}}
\def\tY{{\tens{Y}}}
\def\tZ{{\tens{Z}}}

\def\gA{{\mathcal{A}}}
\def\gB{{\mathcal{B}}}
\def\gC{{\mathcal{C}}}
\def\gD{{\mathcal{D}}}
\def\gE{{\mathcal{E}}}
\def\gF{{\mathcal{F}}}
\def\gG{{\mathcal{G}}}
\def\gH{{\mathcal{H}}}
\def\gI{{\mathcal{I}}}
\def\gJ{{\mathcal{J}}}
\def\gK{{\mathcal{K}}}
\def\gL{{\mathcal{L}}}
\def\gM{{\mathcal{M}}}
\def\gN{{\mathcal{N}}}
\def\gO{{\mathcal{O}}}
\def\gP{{\mathcal{P}}}
\def\gQ{{\mathcal{Q}}}
\def\gR{{\mathcal{R}}}
\def\gS{{\mathcal{S}}}
\def\gT{{\mathcal{T}}}
\def\gU{{\mathcal{U}}}
\def\gV{{\mathcal{V}}}
\def\gW{{\mathcal{W}}}
\def\gX{{\mathcal{X}}}
\def\gY{{\mathcal{Y}}}
\def\gZ{{\mathcal{Z}}}

\def\sA{{\mathbb{A}}}
\def\sB{{\mathbb{B}}}
\def\sC{{\mathbb{C}}}
\def\sD{{\mathbb{D}}}
\def\sF{{\mathbb{F}}}
\def\sG{{\mathbb{G}}}
\def\sH{{\mathbb{H}}}
\def\sI{{\mathbb{I}}}
\def\sJ{{\mathbb{J}}}
\def\sK{{\mathbb{K}}}
\def\sL{{\mathbb{L}}}
\def\sM{{\mathbb{M}}}
\def\sN{{\mathbb{N}}}
\def\sO{{\mathbb{O}}}
\def\sP{{\mathbb{P}}}
\def\sQ{{\mathbb{Q}}}
\def\sR{{\mathbb{R}}}
\def\sS{{\mathbb{S}}}
\def\sT{{\mathbb{T}}}
\def\sU{{\mathbb{U}}}
\def\sV{{\mathbb{V}}}
\def\sW{{\mathbb{W}}}
\def\sX{{\mathbb{X}}}
\def\sY{{\mathbb{Y}}}
\def\sZ{{\mathbb{Z}}}

\def\emLambda{{\Lambda}}
\def\emA{{A}}
\def\emB{{B}}
\def\emC{{C}}
\def\emD{{D}}
\def\emE{{E}}
\def\emF{{F}}
\def\emG{{G}}
\def\emH{{H}}
\def\emI{{I}}
\def\emJ{{J}}
\def\emK{{K}}
\def\emL{{L}}
\def\emM{{M}}
\def\emN{{N}}
\def\emO{{O}}
\def\emP{{P}}
\def\emQ{{Q}}
\def\emR{{R}}
\def\emS{{S}}
\def\emT{{T}}
\def\emU{{U}}
\def\emV{{V}}
\def\emW{{W}}
\def\emX{{X}}
\def\emY{{Y}}
\def\emZ{{Z}}
\def\emSigma{{\Sigma}}

\newcommand{\etens}[1]{\mathsfit{#1}}
\def\etLambda{{\etens{\Lambda}}}
\def\etA{{\etens{A}}}
\def\etB{{\etens{B}}}
\def\etC{{\etens{C}}}
\def\etD{{\etens{D}}}
\def\etE{{\etens{E}}}
\def\etF{{\etens{F}}}
\def\etG{{\etens{G}}}
\def\etH{{\etens{H}}}
\def\etI{{\etens{I}}}
\def\etJ{{\etens{J}}}
\def\etK{{\etens{K}}}
\def\etL{{\etens{L}}}
\def\etM{{\etens{M}}}
\def\etN{{\etens{N}}}
\def\etO{{\etens{O}}}
\def\etP{{\etens{P}}}
\def\etQ{{\etens{Q}}}
\def\etR{{\etens{R}}}
\def\etS{{\etens{S}}}
\def\etT{{\etens{T}}}
\def\etU{{\etens{U}}}
\def\etV{{\etens{V}}}
\def\etW{{\etens{W}}}
\def\etX{{\etens{X}}}
\def\etY{{\etens{Y}}}
\def\etZ{{\etens{Z}}}

\newcommand{\pdata}{p_{\rm{data}}}
\newcommand{\ptrain}{\hat{p}_{\rm{data}}}
\newcommand{\Ptrain}{\hat{P}_{\rm{data}}}
\newcommand{\pmodel}{p_{\rm{model}}}
\newcommand{\Pmodel}{P_{\rm{model}}}
\newcommand{\ptildemodel}{\tilde{p}_{\rm{model}}}
\newcommand{\pencode}{p_{\rm{encoder}}}
\newcommand{\pdecode}{p_{\rm{decoder}}}
\newcommand{\precons}{p_{\rm{reconstruct}}}

\newcommand{\laplace}{\mathrm{Laplace}} %

\newcommand{\E}{\mathbb{E}}
\newcommand{\Ls}{\mathcal{L}}
\newcommand{\R}{\mathbb{R}}
\newcommand{\emp}{\tilde{p}}
\newcommand{\lr}{\alpha}
\newcommand{\reg}{\lambda}
\newcommand{\rect}{\mathrm{rectifier}}
\newcommand{\softmax}{\mathrm{softmax}}
\newcommand{\sigmoid}{\sigma}
\newcommand{\softplus}{\zeta}
\newcommand{\KL}{D_{\mathrm{KL}}}
\newcommand{\Var}{\mathrm{Var}}
\newcommand{\standarderror}{\mathrm{SE}}
\newcommand{\Cov}{\mathrm{Cov}}
\newcommand{\normlzero}{L^0}
\newcommand{\normlone}{L^1}
\newcommand{\normltwo}{L^2}
\newcommand{\normlp}{L^p}
\newcommand{\normmax}{L^\infty}

\newcommand{\parents}{Pa} %

\DeclareMathOperator*{\argmax}{arg\,max}
\DeclareMathOperator*{\argmin}{arg\,min}

\DeclareMathOperator{\sign}{sign}
\DeclareMathOperator{\Tr}{Tr}
\let\ab\allowbreak

\title{\cockpittitle: A Practical Debugging Tool\\ for the Training of Deep Neural Networks}

\author{
	Frank Schneider\thanks{Equal contribution} \\
	University of T{\"u}bingen \\
	Maria-von-Linden-Stra{\ss}e 6\\
	T{\"u}bingen, Germany\\
	{\tt\small f.schneider@uni-tuebingen.de} \\
	\And
	Felix Dangel${}^{\ast}$ \\
	University of T{\"u}bingen \\
	Maria-von-Linden-Stra{\ss}e 6\\
	T{\"u}bingen, Germany\\
	{\tt\small f.dangel@uni-tuebingen.de} \\
	\And
	Philipp Hennig\\
	University of T{\"u}bingen \& \\
	MPI for Intelligent Systems
	\\ T{\"u}bingen, Germany \\
	{\tt\small philipp.hennig@uni-tuebingen.de} }

\begin{document}

\maketitle

\begin{abstract}
  When engineers train deep learning models, they are very much ``flying
  blind''. Commonly used methods for real-time training diagnostics, such as
  monitoring the train/test loss, are limited. Assessing a network's training
  process solely through these performance indicators is akin to debugging
  software without access to internal states through a debugger. To address
  this, we present \cockpit, a collection of instruments that enable a closer
  look into the inner workings of a learning machine, and a more informative and
  meaningful status report for practitioners. It facilitates the identification
  of learning phases and failure modes, like ill-chosen hyperparameters.
  These instruments leverage novel higher-order information about the gradient
  distribution and curvature, which has only recently become efficiently
  accessible. We believe that such a debugging tool, which we open-source for
  \pytorch, is a valuable help in troubleshooting the training
  process. By revealing new insights, it also more generally contributes to
  explainability and interpretability of deep nets.
\end{abstract}

\section{Introduction and motivation}
\label{sec:intro}
Deep learning represents a new programming paradigm: instead of deterministic programs, users design models and ``simply'' train them with data.
In this metaphor, deep learning is a meta-programming form, where \emph{coding}
is replaced by \emph{training}. Here, we ponder the question how we can 
provide more insight into this process by building a 
\emph{debugger} specifically designed for deep learning.

Debuggers are crucial for traditional software development. When things
fail, they provide access to the internal workings of the code,
allowing a look ``into the box''. This is much more efficient than re-running the
program with different inputs. And yet, deep learning is
arguably closer to the latter. If the attempt to train a deep net fails, 
a machine learning engineer faces various options: 
Should they change the training hyperparameters (how?); the
optimizer (to which one?); the model (how?); or just re-run
with a different seed? Machine learning toolboxes provide scant help to guide
these decisions.

\begin{figure*}
	\centering
	\includegraphics[width=\textwidth]{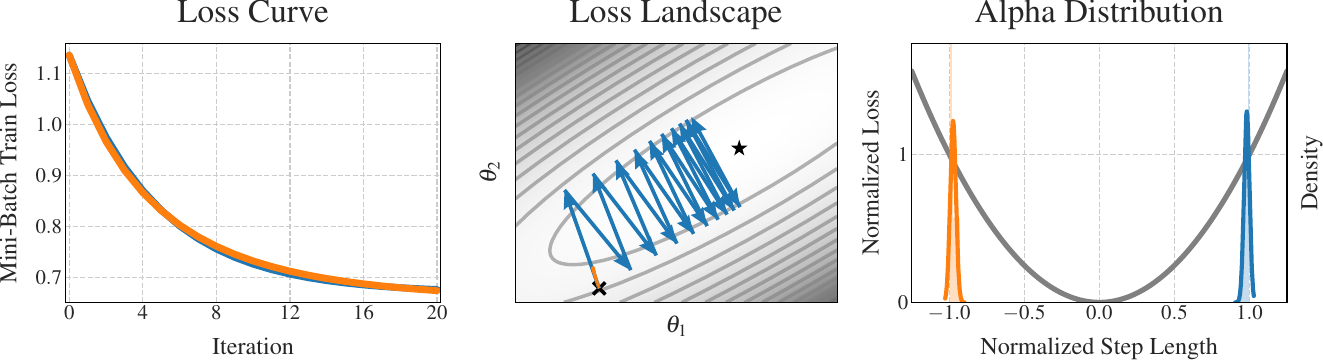}
	\caption{\textbf{Illustrative example: Learning curves do not tell the whole
      story}. Two different optimization runs
    (\textcolor{sns_orange}{\textbf{---}}/\textcolor{sns_blue}{\textbf{---}})
    can lead to virtually the same loss curve (\textit{left}). However, the
    actual optimization trajectories (\textit{middle}), exhibit vastly different
    behaviors. In practice, the trajectories are intractably large and 
    cannot be visualized directly.
    Recommendable actions for both scenarios
    (\textcolor{sns_orange}{increase}/\textcolor{sns_blue}{decrease} the
    learning rate) cannot be inferred from the loss curve. The $\alpha$-distribution, 
    one \cockpit instrument (\textit{right}), not only clearly 
    distinguishes the two scenarios, but also allows for taking decisions
    regarding how the learning rate should be adapted. See \Cref{sec:alpha_exp} 
    for further details.}
	\label{fig:LINE}
\end{figure*}

Of course, traditional debuggers can be applied to deep learning.
They will give access to every single weight of a neural net, or the
individual pixels of its training data.
But this rarely yields insights towards successful training.
Extracting meaningful information requires a statistical approach and
distillation of the bewildering complexity into a manageable summary.
Tools like \tensorboard \citep{Abadi2015} or \wandb \citep{Biewald2020} were built 
in part to streamline this visualization.
Yet, the quantities that are widely monitored (mainly train/test loss \&
accuracy), provide only scant explanation for relative differences between
multiple training runs, because \emph{they do not show the network's internal state}.
\Cref{fig:LINE} illustrates how such established learning curves can describe
the \emph{current} state of the model -- whether it is performing well or not
-- while failing to inform about training state and dynamics.
They tell the user \emph{that} things are going well or badly, but not
\emph{why}.
The situation is similar to flying a plane by sight, without instruments to provide feedback. 
It is not surprising, then, that achieving state-of-the-art performance in deep
learning requires expert intuition, or plain trial \& error.

We aim to enrich the deep learning pipeline with a visual and
statistical debugging tool that uses newly proposed observables as well as several
established ones (\Cref{sec:instruments}).
We leverage and augment recent extensions to automatic differentiation (\ie 
\backpack \citep{Dangel2020} for \pytorch \citep{Paszke2019}) to
efficiently access second-order statistical (\eg gradient variances) and
geometric (\eg Hessian) information.
We show how these quantities can aid the deep learning
engineer in tasks, like learning rate selection, as well as
detecting common bugs with data processing or model architectures
(\Cref{sec:experiments}).

\begin{figure*}
	\centering
	\Cshadowbox{\includegraphics[width=0.97\textwidth, trim={7cm 2.5cm 5cm
	0.5cm}, clip]{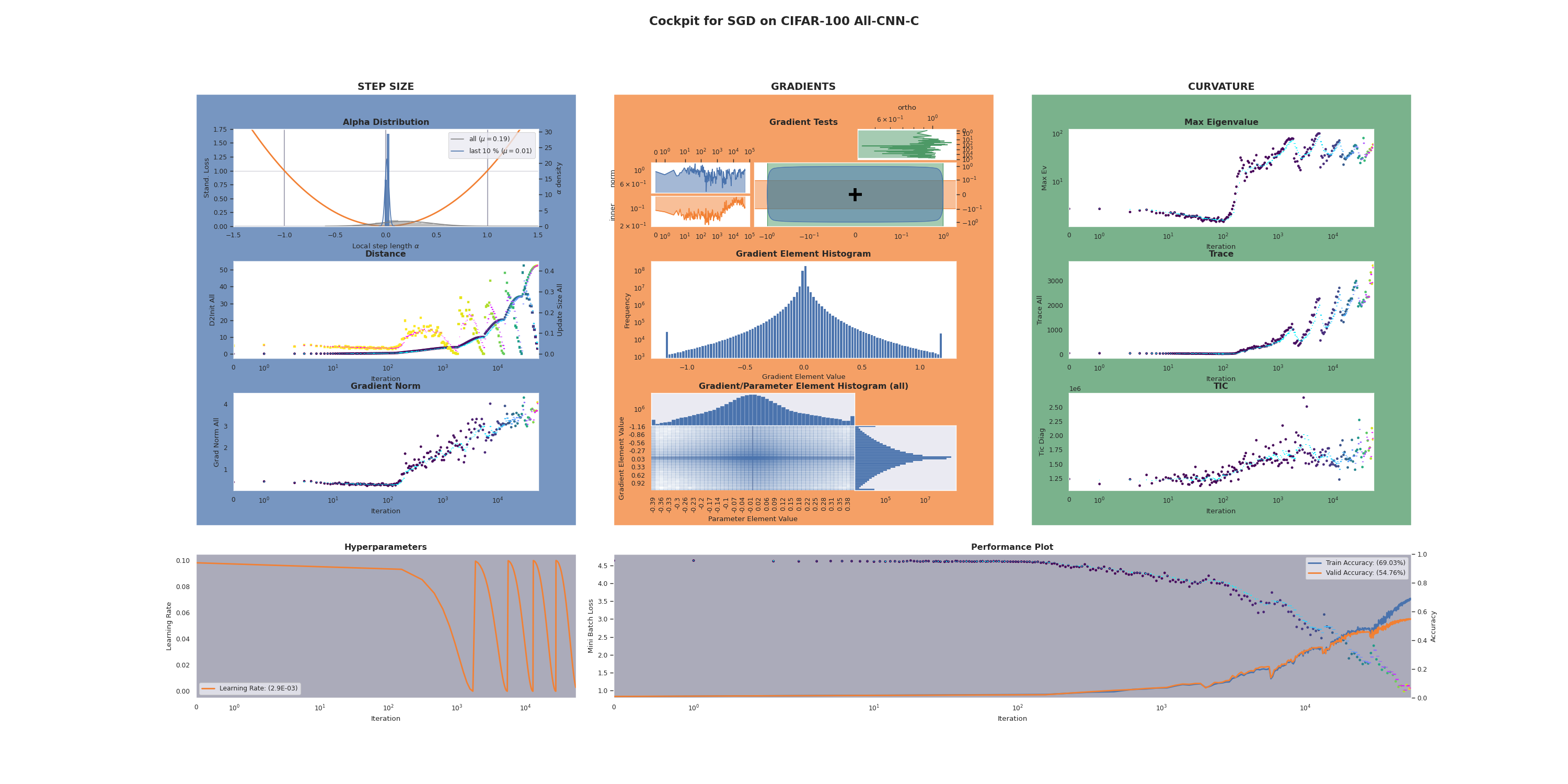}}
  \vspace{0.5ex}
	\caption{\textbf{Screenshot of \cockpittitle's full view} while training the
	\allcnnc \citep{Springenberg2015} on \cifarhun with \sgd using a cyclical
	learning rate schedule. This figure and its labels are not meant to be legible, but rather
	give an impression of \cockpit's user experience. Gray
	panels (bottom row) show the information currently tracked by most
	practitioners. The individual instruments are discussed in
	\Cref{sec:instruments}, and observations are described in
	\Cref{sec:showcase}. An animated version can be found in the accompanying
	GitHub repository.}
	\label{fig:showcase}
\end{figure*}

Concretely, we introduce \cockpit, a flexible and efficient framework
for online-monitoring these observables during training in carefully designed
plots we call ``instruments'' (see \Cref{fig:showcase}).
To be of practical use, such visualization must have a manageable computational
overhead.
We show that \cockpit scales well to real-world deep learning problems (see
\Cref{fig:showcase} and \Cref{sec:showcase}).
We also provide three different configurations of varying computational
complexity and demonstrate that their instruments keep the computational cost
\textit{well below} a factor of $2$ in run time (\Cref{sec:benchmark}).
It is available as
open-source code,\footnote{\cockpiturl} 
extendable, and seamlessly integrates into
existing \pytorch training loops (see \Cref{app:code_example}).

\section{\cockpittitle's instruments}
\label{sec:instruments}
\paragraph{Setting:}
We consider supervised regression/classification with labeled data $(\vx,
\vy) \in \sX \times \sY$ generated by a distribution $P(\vx,
\vy)$. The training set $\gD = \left\{ (\vx_n, \vy_n)\ |\ n = 1, \dots, N
\right\}$ consists of $N$ i.i.d.\,samples from $P$ and the deep model $f:
\Theta \times \sX \rightarrow \sY$ maps inputs $\vx_n$ to predictions
$\hat{\vy}_n$ by parameters $\vtheta \in \sR^D$.
This prediction is evaluated by a loss function
$\ell : \sY \times \sY \rightarrow \R$ which compares to the label $\vy_n$. The
goal is minimizing an inaccessible expected risk $\gL_P(\vtheta) = \int
\ell(f(\vtheta, \vx), \vy) \ \rd P(\vx, \vy)$ by empirical approximation through
$\gL_{\gD}(\vtheta) = \frac{1}{N} \sum_{n=1}^N \ell(f(\vtheta,
\vx_n), \vy_n) := \frac{1}{N} \sum_{n=1}^N \ell_n(\vtheta)$, which in practice
though can only be stochastically sub-sampled on mini-batches $\gB \subseteq
\{1, \dots, N\}$,
\begin{align}
  \label{eq:mini-batch-loss}
  \gL_{\gB}(\vtheta) = \frac{1}{|\gB|} \sum_{n\in\gB} \ell_n(\vtheta)\,.
\end{align}
As is standard practice, we use first- and second-order information of the mini-batch
loss, described by its gradient $\vg_{\gB}(\vtheta)$ and Hessian $\mH_{\gB}(\vtheta)$,
\begin{align}
	\vg_{\gB}(\vtheta) = \frac{1}{|\gB|} \sum_{n \in \gB}
	\underbrace{\nabla_\vtheta\ell_n(\vtheta)}_{\vg_n(\vtheta)} \, ,
	\qquad & \qquad
	\mH_{\gB}(\vtheta) = \frac{1}{|\gB|} \sum_{n \in \gB}
	\nabla^2_\vtheta \ell_n(\vtheta)\,.
\end{align}

\paragraph{Design choices:}
To minimize computational and design overhead, we restrict the metrics to
quantities that require no additional model evaluations. This means that, at
training step $t \to t + 1$ with mini-batches $\gB_t, \gB_{t+1}$ and parameters
$\vtheta_t, \vtheta_{t+1}$, we may access information about the mini-batch
losses $\gL_{\gB_t}(\vtheta_t)$ and $\gL_{\gB_{t+1}}(\vtheta_{t+1})$,
but no cross-terms that would require additional forward passes.

\paragraph{Key point:} $\gL_{\gB}(\vtheta), \vg_{\gB}(\vtheta)$, and
$\mH_{\gB}(\vtheta)$ are just expected values of a \textit{distribution} over
the batch. Only recently, this distribution has begun to attract attention
\citep{Faghri2020} as its computation has become more accessible
\citep{Bradbury2018, Dangel2020}. Contemporary optimizers leverage only the
\emph{mean} gradient and neglect higher moments. One core point of our work is
making extensive use of these distribution properties, trying to visualize them
in various ways.
This out-of-the-box support for the carefully selected and efficiently computed
quantities distinguishes \cockpit from tools like \tensorboard that offer
visualizations as well.
Leveraging these distributional quantities, we create instruments and show how
they can help adapt hyperparameters (\Cref{sec:adapting_hyperparameters}),
analyze the loss landscape (\Cref{sec:curvature}), and track network dynamics
(\Cref{sec:network_dynamics}). Instruments can sometimes be built from
already-computed information or are efficient variants of previously proposed
observables. To keep the presentation concise, we highlight the instruments
shown in \Cref{fig:showcase} and listed in \Cref{tab:overview-quantities}.
\Cref{app:instruments} defines them formally and contains more extensions, such
as the mean \gsnr \citep{Liu2020}, the early stopping \citep{Mahsereci2017} and
\cabs \citep{Balles2017} criterion, which can all be used in \cockpit.

{\def\arraystretch{1.2}
\begin{table*}
	\caption{\textbf{Overview of \cockpittitle quantities}. They range from
		cheap byproducts, to nonlinear transformations of first-order information
		and Hessian-based measures. Some quantities have already been proposed,
		others are first to be considered in this work. They are categorized into
		configurations \textit{economy $\subseteq$ business $\subseteq$ full} based on their
		run time overhead (see \Cref{sec:benchmark} for a detailed evaluation).}
	\label{tab:overview-quantities}
	\begin{center}
    \begin{tabularx}{\textwidth}{lXcr}
    	\toprule
    	\textbf{Name}       & \textbf{Short Description}                                                                                           & \textbf{Config}   & \textbf{Pos. in \Cref{fig:showcase}}  \\
    	\midrule
    	\texttt{Alpha}      & Normalized step size on a noisy quadratic interpolation between two iterates $\vtheta_t$, $\vtheta_{t+1}$            & \textit{economy}  & top \textcolor{sns_blue}{left}        \\
    	\texttt{Distance}   & Distance from initialization $\lVert \vtheta_{t} -  \vtheta_{0}  \rVert_2$                                           & \textit{economy}  & middle \textcolor{sns_blue}{left}     \\
    	\texttt{UpdateSize} & Update size $\lVert \vtheta_{t + 1} - \vtheta_{t} \rVert_2$                                                          & \textit{economy}  & middle \textcolor{sns_blue}{left}     \\
    	\texttt{GradNorm}   & Mini-batch gradient norm $\lVert \vg_{\gB}(\vtheta) \rVert_2$                                                        & \textit{economy}  & bottom \textcolor{sns_blue}{left}     \\
    	\texttt{NormTest}   & Normalized fluctuations of the residual norms ${\lVert  \vg_{\gB} - \vg_n \rVert_2}$, proposed in \citep{Byrd2012}   & \textit{economy}  & top \textcolor{sns_orange}{center}    \\
    	\texttt{InnerTest}  & Normalized fluctuations of the $\vg_n$'s parallel components along $\vg_{\gB}$, proposed in \citep{Bollapragada2017} & \textit{economy}  & top \textcolor{sns_orange}{center}    \\
    	\texttt{OrthoTest}  & Same as \texttt{InnerTest} but using the orthogonal components, proposed in \citep{Bollapragada2017}                 & \textit{economy}  & top \textcolor{sns_orange}{center}    \\
    	\texttt{GradHist1d} & Histogram of individual gradient elements, $\{\vg_n(\vtheta_j)\}_{n\in \gB}^{j=1,\dots,D}$                           & \textit{economy}  & middle \textcolor{sns_orange}{center} \\
    	\texttt{TICDiag}    & Relation between (diagonal) curvature and gradient noise, inspired by \citep{Thomas2020}                             & \textit{business} & bottom \textcolor{sns_green}{right}   \\
    	\texttt{HessTrace}  & Exact or approximate Hessian trace, $\Tr(\mH_{\gB}(\vtheta))$, inspired by \citep{Yao2020}                           & \textit{business} & middle \textcolor{sns_green}{right}   \\
    	\texttt{HessMaxEV}  & Maximum Hessian eigenvalue, $\lambda_{\text{max}}(\mH_{\gB}(\vtheta))$, inspired by \citep{Yao2020}                  & \textit{full}     & top \textcolor{sns_green}{right}      \\
    	\texttt{GradHist2d} & Histogram of weights and individual gradient elements, $\{ (\vtheta_j,\vg_n(\vtheta_j))\}_{n\in \gB}^{j=1,\dots,D}$  & \textit{full}     & bottom \textcolor{sns_orange}{center} \\
    	\bottomrule
    \end{tabularx}
	\end{center}
\end{table*}
}

\paragraph{Bug types:} We distinguish three types of bugs encountered in deep
learning. \emph{Implementation bugs} are low-level software bugs that, for
example, trigger syntax errors. \emph{Training bugs} result in unnecessarily
inefficient or even unsuccessful training. They can, for example, stem from
erroneous data handling (see \Cref{sec:misscaled_data_exp}), the chosen model
architecture (see \Cref{sec:vanishing_gradient_exp}), or ill-chosen
hyperparameters (see \Cref{sec:alpha_exp}). \emph{Prediction bugs}
describe incorrect predictions of a trained model on specific examples.
Traditional debuggers are well-suited to find implementation bugs. \cockpit
focuses on efficiently identifying training bugs instead.

\subsection{Adapting hyperparameters}
\label{sec:adapting_hyperparameters}
One big challenge in deep learning is setting the hyperparameters
correctly, which is currently mostly done by trial \& error through
parameter searches. We aim to augment this process with 
instruments that inform the user about the effect that the chosen 
parameters have on the current training process.

\textbf{Alpha: Are we crossing the valley?}
Using individual loss and gradient observations at the start and end point of
each iteration, we build a noise-informed univariate quadratic approximation
along the step direction (\ie the loss as a function of the step size), and
assess to which point on this parabola our optimizer moves. We standardize this
value $\alpha$ such that stepping to the valley-floor is assigned $\alpha=0$, 
the starting point is at $\alpha=-1$ and updates to the point exactly opposite 
of the starting point have $\alpha=1$ (see \Cref{app:alpha} for a more detailed 
visual and mathematical description of $\alpha$). 
\Cref{fig:LINE} illustrates the scenarios $\alpha=\pm1$ and how monitoring the
$\alpha$-distribution (right panel) can help distinguish between two training 
runs with similar performance but distinct failure sources. 
By default, this \cockpit instrument shows the $\alpha$-distribution for 
the last 10\,\% of training and the entire training process 
(\eg top left plot in \Cref{fig:showcase}).
In \Cref{sec:alpha_exp} we demonstrate empirically that, counter-intuitively, 
it is generally \emph{not} a good idea to choose the step size such that 
$\alpha$ is close to zero.

\textbf{Distances: Are we making progress?}
Another way to discern the trajectories in \Cref{fig:LINE} is by measuring the
$L_2$ \textit{distance from initialization} \citep{Nagarajan2019}
and the \textit{update size} \citep{Agrawal2020,Frankle2020} in parameter space.
Both are shown together in one \cockpit instrument (see also middle left plot in 
\Cref{fig:showcase}) and are far larger for the blue line in \Cref{fig:LINE}.
These distance metrics are also able to disentangle phases for the blue path.
Using the same step size, it will continue to ``jump back and forth''
between the loss valley's walls but at some point cease to make progress.
During this ``surfing of the walls'', the \textit{distance from initialization} 
increases, ultimately though, it will stagnate, with the \textit{update size} 
remaining non-zero, indicating diffusion. While the initial
``surfing the wall''-phase benefits training (see \Cref{sec:alpha_exp}), achieving 
stationarity may require adaptation once the optimizer reaches that diffusion.

\textbf{Gradient norm: How steep is the wall?}
The \textit{update size} will 
show that the orange trajectory is stuck. But why?
Such slow-down can result from both a bad learning rate and from loss landscape
plateaus. The \textit{gradient norm} (bottom left panel in \Cref{fig:showcase})
distinguishes these two causes.

\textbf{Gradient tests: How noisy is the batch?}
The batch size trades off gradient accuracy versus computational cost.
Recently, adaptive sampling strategies based on testing geometric constraints
between mean and individual gradients have been proposed 
\citep{Byrd2012,Bollapragada2017}. The \textit{norm}, \textit{inner product}, 
and \textit{orthogonality tests} use a standardized radius and two
band widths (parallel and orthogonal to the gradient mean) that indicate how
strongly individual gradients scatter around the mean. The original works use
these values to adapt batch sizes. Instead, \cockpit combines all three tests
into a single gauge (top center plot of \Cref{fig:showcase}) using the
standardized noise radius and band widths for visualization.
These noise signals can be used to guide batch size adaptation on- and 
offline, or to probe the influence of gradient alignment on training 
speed \citep{Sankararaman2020} and generalization
\citep{Chatterjee2020a,Chatterjee2020b,Liu2020}.

\subsection{Hessian properties for local loss geometry}
\label{sec:curvature}
An intuition for the local loss landscape helps in many ways. It can
help diagnose whether training is stuck, to adapt the step size, and explain
stability or regularization \citep{Ginsburg2020,Jastrzebski2020}. 
The key challenge is the large number of weights: Low-dimensional 
projections of surfaces can behave unintuitively \citep{Mulayoff2020}, 
but tracking the extreme or average behaviors may help in debugging, 
especially if first-order metrics fail.

\textbf{Hessian eigenvalues: A gorge or a lake?}
In convex optimization, the maximum Hessian eigenvalue crucially determines the
appropriate step size \citep{Schmidt2014}. Many works have studied the
Hessian spectrum in machine learning 
\citep[\eg][]{Ghorbani2019,Ginsburg2020,Mulayoff2020,Sagun2017,Sagun2018,Yao2020}. In
short: curvature matters. Established \citep{Pearlmutter1994} and recent 
autodiff frameworks \citep{Dangel2020} can compute Hessian
properties without requiring the full matrix. \cockpit leverages this
to provide the \textit{Hessian's largest eigenvalue}
and \textit{trace} (top right and middle right plots in \Cref{fig:showcase}).
The former resembles the loss surface's sharpest valley and can thus
hint at training instabilities \citep{Jastrzebski2020}.
The \textit{trace} describes a notion of ``average curvature'', since the 
eigenvalues $\lambda_i$ relate to it by $\sum_i \lambda_i = 
\Tr(\mH_{\gB}(\vtheta))$, which might correlate with generalization 
\citep{Jastrzebski2020a}.

\textbf{TIC: How do curvature and gradient noise interact?}
There is an ongoing debate about curvature's link to
generalization \citep[\eg][]{Dinh2017,Hochreiter1997,Keskar2017}. 
The \emph{Takeuchi Information Criterion (TIC)} \citep{Takeuchi1976,Thomas2020} 
estimates the generalization gap by a ratio between Hessian and non-central 
second gradient moment. It also provides intuition for changes
in the objective function implied by gradient noise. Inspired by the
approximations in \citep{Thomas2020}, \cockpit provides mini-batch TIC
estimates (bottom right plot of \Cref{fig:showcase}).

\subsection{Visualizing internal network dynamics}
\label{sec:network_dynamics}
Histograms are a natural visual compression of the high-dimensional $|\gB| \times D$ 
individual gradient values. They give insights into the gradient
\emph{distribution} and hence offer a more detailed view of the learning
signal. Together with the parameter associated to each
individual gradient, the entire model status and dynamics can be visualized in 
a single plot and be monitored during training.
This provides a more fine-grained view of training compared to tracking parameters 
and gradient norms \citep{Frankle2020}.

\textbf{Gradient and parameter histograms: What is happening in our network?}
\cockpit offers a univariate \textit{histogram of the gradient elements}
$\{\vg_n( \vtheta )_j\}_{n\in \gB}^{j=1,\dots,D}$. Additionally, a combined \textit{histogram of 
parameter-gradient pairs} $\{(\vtheta_j, \vg_n( \vtheta_j )\}_{n\in \gB}^{j=1,\dots,D}$ provides a 
two-dimensional look into the network's gradient and parameter values in a mini-batch.
\Cref{sec:misscaled_data_exp} shows an example use-case of the gradient histogram; 
\Cref{sec:vanishing_gradient_exp} makes the case for the layer-wise variants of the
instruments.

\section{Experiments}
\label{sec:experiments}
The diverse information provided by \cockpit can help users and researchers in
many ways, some of which, just like for a traditional debugger, only become
apparent in practical use. In this section, we present a few motivating examples, 
selecting specific instruments and scenarios in which they are practically useful. 
Specifically, we show that \cockpit can
help the user discern between, and thus fix, common training bugs
(\Cref{sec:misscaled_data_exp,sec:vanishing_gradient_exp}) that are otherwise
hard to distinguish as they lead to the same failure: bad training. We
demonstrate that \cockpit can guide practitioners to choose efficient
hyperparameters \emph{within a single training run}
(\Cref{sec:vanishing_gradient_exp,sec:alpha_exp}). Finally, we highlight that
\cockpit's instruments can provide research insights about the optimization
process (\Cref{sec:alpha_exp}). Our empirical findings are demonstrated on 
problems from the \deepobs \citep{Schneider2019} benchmark collection.

\subsection{Incorrectly scaled data}
\label{sec:misscaled_data_exp}
One prominent source of bugs is the data pipeline.
To pick a relatively simple example: For standard optimizers to work at their
usual learning rates, network inputs must be standardized (\ie~between zero and
one, or have zero mean and unit variance \citep[\eg][]{Bengio2012}).
If the user forgets to do this, optimizer performance is likely to degrade.
It can be difficult to identify the source of this problem as it does not cause
obvious failures, \texttt{NaN} or \texttt{Inf} gradients, \etc.
We now construct a semi-realistic example, to show how using \cockpit can help
diagnose this problem upon observing slow training performance. 

By default\footnote{\url{https://www.cs.toronto.edu/~kriz/cifar.html}}, the popular
image data sets \cifartenhun \citep{Krizhevsky2009} are provided as
\numpy \citep{Harris2020} arrays that consist of integers in the interval
$[0,255]$. This \emph{raw} data, instead of the widely used version with
floats in $[0,1]$, changes the data scale by a factor of $255$ and thus the gradients
as well.  Therefore, the optimizer's optimal learning rate is
scaled as well. In other words, the default parameters of
popular optimization methods may not work well anymore, or good hyperparameters
may take extreme values.
Even if the user directly inspects the training images, this may not be apparent
(see \Cref{fig:data-pre-processing} and \Cref{fig:data-pre-processing_imagenet}
in the appendix for the same experiment with \vgg on \imagenet).
But the gradient histogram instrument of \cockpit, which has a deliberate
default plotting range around $[-1,1]$ to highlight such problems,
immediately and prominently shows that there is an issue.

\begin{figure}
		\pgfkeys{/pgfplots/preprocessingexperimentdefault/.style={
				width=\linewidth,
				height=1.4\linewidth,
				every axis plot/.append style={line width = 1.2pt},
				tick pos = left,
				xmajorticks = true,
				ymajorticks = true,
				ylabel near ticks,
				xlabel near ticks,
				xtick align = inside,
				ytick align = inside,
				legend cell align = left,
				legend columns = 1,
				legend pos = south east,
				legend style = {
					fill opacity = 0.9,
					text opacity = 1,
					font = \small,
				},
				xticklabel style = {font = \small, inner xsep = -5ex},
				xlabel style = {font = \small},
				axis line style = {black},
				yticklabel style = {font = \small, inner ysep = -4ex},
				ylabel style = {font = \small},
				title style = {font = \small, inner ysep = -3ex},
				grid = major,
				grid style = {dashed}
			}
		}
	
	\centering
	\begin{subfigure}[t]{0.46\textwidth}
		\begin{minipage}{.49\textwidth}
			\Cshadowbox{\includegraphics[width = .35\textwidth]{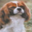}}
			\Cshadowbox{\includegraphics[width = .35\textwidth]{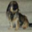}}
	
			\Cshadowbox{\includegraphics[width = .35\textwidth]{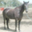}}
			\Cshadowbox{\includegraphics[width = .35\textwidth]{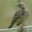}}
		\end{minipage}
		\begin{minipage}{.49\textwidth}
			\centering
			\pgfkeys{/pgfplots/zmystyle/.style={preprocessingexperimentdefault,
					ylabel={Gradient Element}
			}}
			\vspace{1.4\baselineskip}
\begin{tikzpicture}

\definecolor{color0}{rgb}{0.12156862745098,0.466666666666667,0.705882352941177}

\begin{axis}[
axis line style={white},
log basis x={10},
tick align=outside,
xmajorticks=false,
xmin=0.9, xmax=239494401.773689,
xmode=log,
xtick style={color=white!15!black},
ymajorticks=false,
ymin=-1.5, ymax=1.5,
zmystyle
]
\draw[draw=white,fill=color0,line width=0.04pt] (axis cs:0.9,-1.5) rectangle (axis cs:0.9,-1.42499995231628);
\draw[draw=white,fill=color0,line width=0.04pt] (axis cs:0.9,-1.42500007152557) rectangle (axis cs:0.9,-1.35000002384186);
\draw[draw=white,fill=color0,line width=0.04pt] (axis cs:0.9,-1.35000002384186) rectangle (axis cs:0.9,-1.27499997615814);
\draw[draw=white,fill=color0,line width=0.04pt] (axis cs:0.9,-1.27499997615814) rectangle (axis cs:0.9,-1.19999992847443);
\draw[draw=white,fill=color0,line width=0.04pt] (axis cs:0.9,-1.20000004768372) rectangle (axis cs:0.9,-1.125);
\draw[draw=white,fill=color0,line width=0.04pt] (axis cs:0.9,-1.125) rectangle (axis cs:0.9,-1.04999995231628);
\draw[draw=white,fill=color0,line width=0.04pt] (axis cs:0.9,-1.04999995231628) rectangle (axis cs:0.9,-0.974999904632568);
\draw[draw=white,fill=color0,line width=0.04pt] (axis cs:0.9,-0.975000023841858) rectangle (axis cs:0.9,-0.899999976158142);
\draw[draw=white,fill=color0,line width=0.04pt] (axis cs:0.9,-0.899999976158142) rectangle (axis cs:0.9,-0.824999928474426);
\draw[draw=white,fill=color0,line width=0.04pt] (axis cs:0.9,-0.825000047683716) rectangle (axis cs:0.9,-0.75);
\draw[draw=white,fill=color0,line width=0.04pt] (axis cs:0.9,-0.75) rectangle (axis cs:0.9,-0.674999952316284);
\draw[draw=white,fill=color0,line width=0.04pt] (axis cs:0.9,-0.674999952316284) rectangle (axis cs:0.9,-0.599999904632568);
\draw[draw=white,fill=color0,line width=0.04pt] (axis cs:0.9,-0.600000023841858) rectangle (axis cs:0.9,-0.524999976158142);
\draw[draw=white,fill=color0,line width=0.04pt] (axis cs:0.9,-0.524999976158142) rectangle (axis cs:0.9,-0.449999928474426);
\draw[draw=white,fill=color0,line width=0.04pt] (axis cs:0.9,-0.449999988079071) rectangle (axis cs:0.9,-0.374999940395355);
\draw[draw=white,fill=color0,line width=0.04pt] (axis cs:0.9,-0.374999970197678) rectangle (axis cs:0.9,-0.299999922513962);
\draw[draw=white,fill=color0,line width=0.04pt] (axis cs:0.9,-0.299999982118607) rectangle (axis cs:15.9,-0.224999934434891);
\draw[draw=white,fill=color0,line width=0.04pt] (axis cs:0.9,-0.224999964237213) rectangle (axis cs:755.9,-0.149999916553497);
\draw[draw=white,fill=color0,line width=0.04pt] (axis cs:0.9,-0.149999968707561) rectangle (axis cs:12866.9,-0.0749999210238457);
\draw[draw=white,fill=color0,line width=0.04pt] (axis cs:0.9,-0.075000025331974) rectangle (axis cs:95082889.9,2.23517417907715e-08);
\draw[draw=white,fill=color0,line width=0.04pt] (axis cs:0.9,-8.19563865661621e-08) rectangle (axis cs:19475047.9,0.0749999657273293);
\draw[draw=white,fill=color0,line width=0.04pt] (axis cs:0.9,0.0749999210238457) rectangle (axis cs:14378.9,0.149999968707561);
\draw[draw=white,fill=color0,line width=0.04pt] (axis cs:0.9,0.149999916553497) rectangle (axis cs:794.9,0.224999964237213);
\draw[draw=white,fill=color0,line width=0.04pt] (axis cs:0.9,0.224999934434891) rectangle (axis cs:8.9,0.299999982118607);
\draw[draw=white,fill=color0,line width=0.04pt] (axis cs:0.9,0.299999922513962) rectangle (axis cs:0.9,0.374999970197678);
\draw[draw=white,fill=color0,line width=0.04pt] (axis cs:0.9,0.374999940395355) rectangle (axis cs:0.9,0.449999988079071);
\draw[draw=white,fill=color0,line width=0.04pt] (axis cs:0.9,0.449999928474426) rectangle (axis cs:0.9,0.524999976158142);
\draw[draw=white,fill=color0,line width=0.04pt] (axis cs:0.9,0.524999976158142) rectangle (axis cs:0.9,0.600000023841858);
\draw[draw=white,fill=color0,line width=0.04pt] (axis cs:0.9,0.599999904632568) rectangle (axis cs:0.9,0.674999952316284);
\draw[draw=white,fill=color0,line width=0.04pt] (axis cs:0.9,0.674999952316284) rectangle (axis cs:0.9,0.75);
\draw[draw=white,fill=color0,line width=0.04pt] (axis cs:0.9,0.75) rectangle (axis cs:0.9,0.825000047683716);
\draw[draw=white,fill=color0,line width=0.04pt] (axis cs:0.9,0.824999928474426) rectangle (axis cs:30.9,0.899999976158142);
\draw[draw=white,fill=color0,line width=0.04pt] (axis cs:0.9,0.899999976158142) rectangle (axis cs:98.9,0.975000023841858);
\draw[draw=white,fill=color0,line width=0.04pt] (axis cs:0.9,0.974999904632568) rectangle (axis cs:0.9,1.04999995231628);
\draw[draw=white,fill=color0,line width=0.04pt] (axis cs:0.9,1.04999995231628) rectangle (axis cs:0.9,1.125);
\draw[draw=white,fill=color0,line width=0.04pt] (axis cs:0.9,1.125) rectangle (axis cs:0.9,1.20000004768372);
\draw[draw=white,fill=color0,line width=0.04pt] (axis cs:0.9,1.19999992847443) rectangle (axis cs:0.9,1.27499997615814);
\draw[draw=white,fill=color0,line width=0.04pt] (axis cs:0.9,1.27499997615814) rectangle (axis cs:0.9,1.35000002384186);
\draw[draw=white,fill=color0,line width=0.04pt] (axis cs:0.9,1.35000002384186) rectangle (axis cs:0.9,1.42500007152557);
\draw[draw=white,fill=color0,line width=0.04pt] (axis cs:0.9,1.42499995231628) rectangle (axis cs:0.9,1.5);
\end{axis}

\end{tikzpicture}
 		\end{minipage}
		\caption{Normalized Data}
		\label{fig:data-pre-processing_norm}
	\end{subfigure}
	\hfill
	\begin{subfigure}[t]{0.46\textwidth}
		\begin{minipage}{.49\textwidth}
			\Cshadowbox{\includegraphics[width = .35\textwidth]{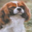}}
			\Cshadowbox{\includegraphics[width = .35\textwidth]{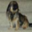}}
			
			\Cshadowbox{\includegraphics[width = .35\textwidth]{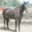}}
			\Cshadowbox{\includegraphics[width = .35\textwidth]{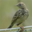}}
		\end{minipage}
		\begin{minipage}{.49\textwidth}
			\centering
			\pgfkeys{/pgfplots/zmystyle/.style={preprocessingexperimentdefault,
					ylabel={Gradient Element}
			}}
			\vspace{1.4\baselineskip}
\begin{tikzpicture}

\definecolor{color0}{rgb}{1,0.498039215686275,0.0549019607843137}

\begin{axis}[
axis line style={white},
log basis x={10},
tick align=outside,
xmajorticks=false,
xmin=0.9, xmax=199565229.367149,
xmode=log,
xtick style={color=white!15!black},
ymajorticks=false,
ymin=-1.5, ymax=1.5,
zmystyle
]
\draw[draw=white,fill=color0,line width=0.04pt] (axis cs:0.9,-1.5) rectangle (axis cs:6792656.9,-1.42499995231628);
\draw[draw=white,fill=color0,line width=0.04pt] (axis cs:0.9,-1.42500007152557) rectangle (axis cs:304537.9,-1.35000002384186);
\draw[draw=white,fill=color0,line width=0.04pt] (axis cs:0.9,-1.35000002384186) rectangle (axis cs:322920.9,-1.27499997615814);
\draw[draw=white,fill=color0,line width=0.04pt] (axis cs:0.9,-1.27499997615814) rectangle (axis cs:342177.9,-1.19999992847443);
\draw[draw=white,fill=color0,line width=0.04pt] (axis cs:0.9,-1.20000004768372) rectangle (axis cs:365697.9,-1.125);
\draw[draw=white,fill=color0,line width=0.04pt] (axis cs:0.9,-1.125) rectangle (axis cs:389812.9,-1.04999995231628);
\draw[draw=white,fill=color0,line width=0.04pt] (axis cs:0.9,-1.04999995231628) rectangle (axis cs:417003.9,-0.974999904632568);
\draw[draw=white,fill=color0,line width=0.04pt] (axis cs:0.9,-0.975000023841858) rectangle (axis cs:446173.9,-0.899999976158142);
\draw[draw=white,fill=color0,line width=0.04pt] (axis cs:0.9,-0.899999976158142) rectangle (axis cs:476860.9,-0.824999928474426);
\draw[draw=white,fill=color0,line width=0.04pt] (axis cs:0.9,-0.825000047683716) rectangle (axis cs:514596.9,-0.75);
\draw[draw=white,fill=color0,line width=0.04pt] (axis cs:0.9,-0.75) rectangle (axis cs:555044.9,-0.674999952316284);
\draw[draw=white,fill=color0,line width=0.04pt] (axis cs:0.9,-0.674999952316284) rectangle (axis cs:601849.9,-0.599999904632568);
\draw[draw=white,fill=color0,line width=0.04pt] (axis cs:0.9,-0.600000023841858) rectangle (axis cs:653348.9,-0.524999976158142);
\draw[draw=white,fill=color0,line width=0.04pt] (axis cs:0.9,-0.524999976158142) rectangle (axis cs:715926.9,-0.449999928474426);
\draw[draw=white,fill=color0,line width=0.04pt] (axis cs:0.9,-0.449999988079071) rectangle (axis cs:790634.9,-0.374999940395355);
\draw[draw=white,fill=color0,line width=0.04pt] (axis cs:0.9,-0.374999970197678) rectangle (axis cs:880821.9,-0.299999922513962);
\draw[draw=white,fill=color0,line width=0.04pt] (axis cs:0.9,-0.299999982118607) rectangle (axis cs:996047.9,-0.224999934434891);
\draw[draw=white,fill=color0,line width=0.04pt] (axis cs:0.9,-0.224999964237213) rectangle (axis cs:1158251.9,-0.149999916553497);
\draw[draw=white,fill=color0,line width=0.04pt] (axis cs:0.9,-0.149999968707561) rectangle (axis cs:1410941.9,-0.0749999210238457);
\draw[draw=white,fill=color0,line width=0.04pt] (axis cs:0.9,-0.075000025331974) rectangle (axis cs:79921533.9,2.23517417907715e-08);
\draw[draw=white,fill=color0,line width=0.04pt] (axis cs:0.9,-8.19563865661621e-08) rectangle (axis cs:1965478.9,0.0749999657273293);
\draw[draw=white,fill=color0,line width=0.04pt] (axis cs:0.9,0.0749999210238457) rectangle (axis cs:1289789.9,0.149999968707561);
\draw[draw=white,fill=color0,line width=0.04pt] (axis cs:0.9,0.149999916553497) rectangle (axis cs:1043170.9,0.224999964237213);
\draw[draw=white,fill=color0,line width=0.04pt] (axis cs:0.9,0.224999934434891) rectangle (axis cs:885990.9,0.299999982118607);
\draw[draw=white,fill=color0,line width=0.04pt] (axis cs:0.9,0.299999922513962) rectangle (axis cs:771669.9,0.374999970197678);
\draw[draw=white,fill=color0,line width=0.04pt] (axis cs:0.9,0.374999940395355) rectangle (axis cs:683281.9,0.449999988079071);
\draw[draw=white,fill=color0,line width=0.04pt] (axis cs:0.9,0.449999928474426) rectangle (axis cs:610831.9,0.524999976158142);
\draw[draw=white,fill=color0,line width=0.04pt] (axis cs:0.9,0.524999976158142) rectangle (axis cs:553292.9,0.600000023841858);
\draw[draw=white,fill=color0,line width=0.04pt] (axis cs:0.9,0.599999904632568) rectangle (axis cs:502804.9,0.674999952316284);
\draw[draw=white,fill=color0,line width=0.04pt] (axis cs:0.9,0.674999952316284) rectangle (axis cs:461496.9,0.75);
\draw[draw=white,fill=color0,line width=0.04pt] (axis cs:0.9,0.75) rectangle (axis cs:423747.9,0.825000047683716);
\draw[draw=white,fill=color0,line width=0.04pt] (axis cs:0.9,0.824999928474426) rectangle (axis cs:392145.9,0.899999976158142);
\draw[draw=white,fill=color0,line width=0.04pt] (axis cs:0.9,0.899999976158142) rectangle (axis cs:363396.9,0.975000023841858);
\draw[draw=white,fill=color0,line width=0.04pt] (axis cs:0.9,0.974999904632568) rectangle (axis cs:335621.9,1.04999995231628);
\draw[draw=white,fill=color0,line width=0.04pt] (axis cs:0.9,1.04999995231628) rectangle (axis cs:312992.9,1.125);
\draw[draw=white,fill=color0,line width=0.04pt] (axis cs:0.9,1.125) rectangle (axis cs:291497.9,1.20000004768372);
\draw[draw=white,fill=color0,line width=0.04pt] (axis cs:0.9,1.19999992847443) rectangle (axis cs:273292.9,1.27499997615814);
\draw[draw=white,fill=color0,line width=0.04pt] (axis cs:0.9,1.27499997615814) rectangle (axis cs:255338.9,1.35000002384186);
\draw[draw=white,fill=color0,line width=0.04pt] (axis cs:0.9,1.35000002384186) rectangle (axis cs:240647.9,1.42500007152557);
\draw[draw=white,fill=color0,line width=0.04pt] (axis cs:0.9,1.42499995231628) rectangle (axis cs:4873580.9,1.5);
\end{axis}

\end{tikzpicture}
 		\end{minipage}
		\caption{Raw Data}
		\label{fig:data-pre-processing_raw}
	\end{subfigure}
	\caption{\textbf{Same inputs, different gradients; Catching data 
	    bugs with \cockpittitle.} (a) \emph{normalized} ($[0, 1]$) and (b)
    	\emph{raw} $([0, 255])$ images look identical in auto-scaled
	    front-ends like \matplotlib's \texttt{imshow}. The gradient distribution on
	    the \threecthreed model, however, is crucially affected by this
	    scaling.}
	\label{fig:data-pre-processing}
\end{figure}

Of course, this particular data is only a placeholder for real practical data sets.
While this problem may not frequently arise in the highly pre-processed,
packaged \cifarten, it is not a rare problem for practitioners who work with
their personal data sets.
This is particularly likely in domains outside standard computer vision, \eg
when working with mixed-type data without obvious natural scales.

\subsection{Vanishing gradients}
\label{sec:vanishing_gradient_exp}

The model architecture itself can be a source of training bugs. As before, such
problems mostly arise with novel data sets, where well-working architectures are
unknown. The following example shows how even small (in terms of code)
architecture modifications may severely harm the training.

\begin{figure}
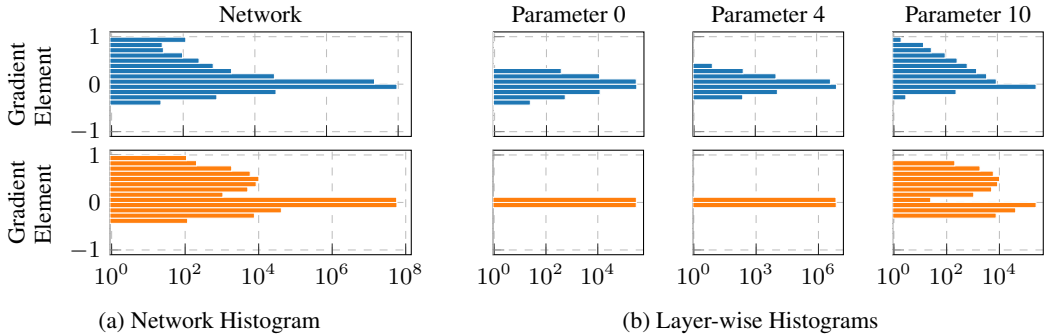

	  \pgfkeys{/pgfplots/layerwiseexperimentdefault/.style={
	      width=\linewidth,
	      height=0.13\textheight,
	      every axis plot/.append style={line width = 1.2pt},
	      tick pos = left,
	      xmajorticks = true,
	      ymajorticks = true,
	      ylabel near ticks,
	      xlabel near ticks,
	      xtick align = inside,
	      ytick align = inside,
	      ytick={-1,0,1},
	      legend cell align = left,
	      legend columns = 1,
	      legend pos = south east,
	      legend style = {
	        fill opacity = 0.9,
	        text opacity = 1,
	        font = \small,
	      },
	      xticklabel style = {font = \small, inner xsep = -5ex},
	      xlabel style = {font = \small},
	      axis line style = {black},
	      yticklabel style = {font = \small, inner ysep = -4ex},
	      ylabel style = {font = \small},
	      title style = {font = \small, inner ysep = -3ex},
	      grid = major,
	      grid style = {dashed}
	    }
	  }
	
	\centering
	\begin{subfigure}[t]{0.4\textwidth}
		\pgfkeys{/pgfplots/zmystyle/.style={layerwiseexperimentdefault,
		   title = {Network}, ylabel=Gradient\\Element, ylabel style={align=left}, xticklabels = {}
		 }}

 		\caption{Network Histogram}
		\label{fig:layerwise-experiment_net}
	\end{subfigure}
	\hfill
	\begin{subfigure}[t]{0.57\textwidth}
		    \pgfkeys{/pgfplots/layerwiseexperimentdefaultparameters/.style={
		        layerwiseexperimentdefault,
		        ylabel={},
		        ylabel style = {inner ysep = -4ex},
		        yticklabels={},
		        width=0.45\textwidth,
		      }
		    }
		    \hfill%
		    \pgfkeys{/pgfplots/zmystyle/.style={layerwiseexperimentdefaultparameters,
		        title={Parameter 0}, xticklabels = {},}}%
%
 		    \hfill
		\caption{Layer-wise Histograms}
		\label{fig:layerwise-experiment_layers}
	\end{subfigure}
	\caption{\textbf{Gradient distributions of two similar architectures on the 
		same problem}. (a) Distribution of individual gradient elements 
		summarized over the entire network. Both seem similar.
		(b) Layer-wise histograms for a subset of layers. Parameter 0 is the layer 
		closest to the network's input, parameter 10 closest to its output. 
		Only the layer-wise view reveals that there are several degenerated gradient 
		distributions for the orange network making training unnecessary hard.}
	\label{fig:layerwise-experiment}
\end{figure}

\Cref{fig:layerwise-experiment_net} shows the distribution of gradient values
of two different network architectures in blue and orange.
Although the blue model trains considerably better than the orange one, their
gradient distributions look quite similar.
The difference becomes evident when inspecting the histogram \emph{layer-wise}.
We can see that multiple layers have a degenerated gradient distribution with many
elements being practically zero (see \Cref{fig:layerwise-experiment_layers}, bottom row).
Since the fully connected layers close to the output have far more parameters 
(a typical pattern of convolutional networks), they dominate
the network-wide histogram. This obscures that a major part of the model
is effectively unable to train.

Both the blue and orange networks follow \deepobs's \threecthreed architecture.
The only difference is the non-linearity: The blue network uses standard ReLU
activations, while the orange one has sigmoid activations. Here, the layer-wise
histogram instrument of \cockpit~highlights which part of the architecture makes
training unnecessarily hard. Accessing information layer-wise is also essential
due to the strong overparameterization in deep models where training can
happen in small subspaces \citep{GurAri2018}. Once again, this is hard to do
with common monitoring tools, such as the loss curve.

\subsection{Tuning learning rates}
\label{sec:alpha_exp}
Once the architecture is defined, the optimizer's learning rate is the most
important hyperparameter to tune.
Getting it right requires extensive hyperparameter searches at high resource
costs.
\cockpit's instruments can provide intuition and information to streamline this process:
In contrast to the raw learning rate, the curvature-standardized step size
$\alpha$-quantity (see \Cref{sec:adapting_hyperparameters}) has a natural scale.

\begin{figure}
	\begin{center}
		\includegraphics[width=\linewidth]{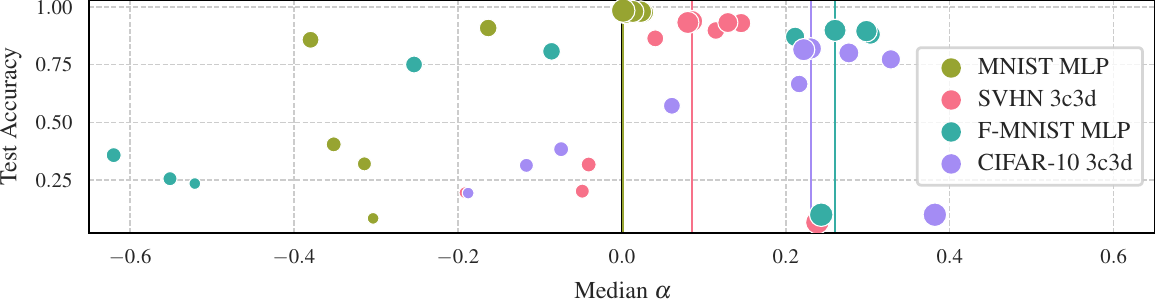}
	\end{center}
	\caption{\textbf{Test accuracy as a function of standardized step size $\alpha$}.
		For four \deepobs problems (see \Cref{app:benchmarks}), final test accuracy is shown versus the median $\alpha$-value over the entire training. Marker size indicates the magnitude of the raw
		learning rate, marker color identifies tasks (see legend). For each problem, the 
		best-performing setting  is highlighted by a vertical colored line.}
	\label{fig:alpha_exp}
\end{figure}

Across multiple optimization problems, we observe, perhaps surprisingly, that
the best runs and indeed all good runs have a median $\alpha>0$
(\Cref{fig:alpha_exp}).
This illustrates a fundamental difference between stochastic
optimization, as is typical for machine learning, and classic deterministic 
optimization.
Instead of locally stepping ``to the valley floor'' (optimal in the
deterministic case), stochastic optimizers should \emph{overshoot} the valley
somewhat.
This need to ``surf the walls'' has been hypothesized before
\citep[e.g.][]{Wu2018a,Xing2018} as a property of neural network training.
Frequently, learning rates are adapted during training, which 
fits with our observation about positive $\alpha$-values:
``Overshooting'' allows fast early progression towards areas of lower loss, but
it does not yield convergence in the end.
Real-time visualizations of the training state, as offered by \cockpit, can augment 
these fine-tuning processes.

\Cref{fig:alpha_exp} also indicates a major challenge preventing simple
automated tuning solutions: The optimal $\alpha$-value is problem-dependent, and
simpler problems, such as a multi-layer perceptron (\mlp) on \mnist
\citep{Lecun1998}, behave much more similar to classic optimization problems.
Algorithmic research on small problems can thus produce misleading conclusions.
The figure also shows that the $\alpha$-gauge is not sufficient by itself:
extreme overshooting with a too-large learning rate leads to poor performance,
which however can be prevented by taking additional instruments into account.
This makes the case for the cockpit metaphor of increasing interpretability from several
instruments in conjunction. By combining the $\alpha$-instrument with other
gauges that capture the local geometry or network dynamics, the user can better
identify good choices of the learning rate and other hyperparameters.

\section{Showcase}
\label{sec:showcase}
Having introduced the tool, we can now return to
\Cref{fig:showcase} for a closer look.
The figure shows a snapshot from training the \allcnnc \citep{Springenberg2015}
on \cifarhun using \sgd with a cyclic learning rate schedule (see
bottom left panel).
Diagonal curvature instruments are configured to use an MC approximation in
order to reduce the run time (here, $C=100$, compare \Cref{sec:benchmark}).

A glance at all panels shows that the learning rate schedule
is reflected in the metrics.
However, the instruments also provide insights into the early phase of
training (first $\sim100$ iterations), where the learning rate is still
unaffected by the schedule:
There, the loss plateaus and the optimizer takes relatively small steps
(compared to later, as can be seen in the small gradient norms, and small
distance from initialization).
Based on these low-cost instruments, one may thus at first suspect that
training was poorly initialized; but training indeed succeeds after iteration
100!
Viewing \cockpit entirely though, it becomes clear that optimization in these
first steps is not stuck at all:
While loss, gradient norms, and distance in parameter space remain almost
constant, curvature changes, which expresses itself in a clear downward trend
of the maximum Hessian eigenvalue (top right panel).

The importance of early training phases has recently been hypothesized
\citep{Frankle2020}, suggesting a logarithmic timeline.
Not only does our showcase support this hypothesis, but it also provides an 
explanation from the curvature-based metrics, which in this particular 
case are the only meaningful feedback in the first few training steps.
It also suggests monitoring training at log-spaced intervals.
\cockpit provides the flexibility to do so, indeed, \Cref{fig:showcase} has
been created with log-scheduled tracking events.

As a final note, we recognize that the approach taken here promotes an amount
of \emph{manual} work (monitoring metrics, deliberately intervening, \etc)
that may seem ironic and at odds with the paradigm of automation that is at the
heart of machine learning.
However, we argue that this might be what is needed at this point in the evolution 
of the field.
Deep learning has been driven notably by scaling compute resources \citep{Thompson2020}, 
and fully automated, one-shot training may still be some way out.
To develop better training methods, researchers, not just users, need 
\emph{algorithmic} interpretability and explainability: direct insights and 
intuition about the processes taking place ``inside'' neural nets.
To highlight how \cockpit might provide this, we contrast in \Cref{app:convex-problems}
the \cockpit view of two convex \deepobs problems: a noisy quadratic and logistic
regression on \mnist. In both cases, the instruments behave differently compared 
to the deep learning problem in \Cref{fig:showcase}.
In particular, the gradient norm increases (left column, bottom panel) during
training, and individual gradients become less scattered (center column, top
panel). This is diametrically opposed to the convex problems and shows that deep
learning differs even qualitatively from well-understood optimization problems.

\section{Benchmark}
\label{sec:benchmark}
\Cref{sec:experiments} made a case for \cockpit as an effective debugging and
tuning tool. To make the library useful in practice, it must 
also have limited computational cost. We now show that it is possible to compute
all quantities at reasonable overhead. The user can
control the absolute cost along two dimensions, by reducing the number of
instruments, or by reducing their update frequency.

All benchmark results show \sgd without momentum. \cockpit's
quantities, however, work for generic optimizers and can mostly be used
identically without increased costs. One current exception is \texttt{Alpha}
which can be computed more efficiently given the optimizer's update
rule.\footnote{This is currently implemented for vanilla \sgd. Otherwise,
	\cockpit falls back to a less efficient scheme.}

\paragraph{Complexity analysis:}
Computing more information adds computational overhead, of course.
However, recent work \citep{Dangel2020} has shown that first-order
information, like distributional statistics on the batch gradients,
can be computed on top of the mean gradient at little extra cost.
Similar savings apply for most quantities in
\Cref{tab:overview-quantities}, as they are \mbox{(non-)linear} transformations
of individual gradients.
A subset of \cockpit's quantities also uses second-order information from the Hessian
diagonal. For ReLU networks on a classification task with $C$ classes, the
additional work is proportional to $C$ gradient backpropagations (\ie $C=10$ for
\cifarten, $C=100$ for \cifarhun). Parallel processing can, to some extent,
process these extra backpropagations in parallel without significant overhead.
If this is no longer possible, we can fall back to a Monte Carlo (MC) sampling
approximation, which reduces the number of extra backprop passes to the number
of samples (1 by default).\footnote{An MC-sampled approximation of
  the Hessian/generalized Gauss-Newton has been used in \Cref{fig:showcase} to
  reduce the prohibitively large number of extra backprops on \cifarhun
  ($C=100$).}

While parallelization is possible for the gradient instruments, computing the
maximum Hessian eigenvalue is inherently sequential.
Similar to \citet{Yao2020}, we use matrix-free Hessian-vector products by 
automatic differentiation \citep{Pearlmutter1994}, where each product's costs 
are proportional to one gradient computation. Regardless of the underlying 
iterative eigensolver, multiple such products must be queried to compute the 
spectral norm (the required number depends on the spectral gap to the 
second-largest eigenvalue).

\begin{figure}	
	\centering
	\begin{subfigure}[t]{0.6\textwidth}
		\centering
		\includegraphics[width=\textwidth]{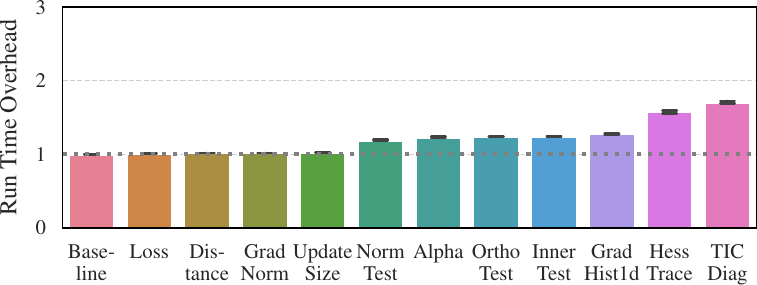}
		\caption{Overhead \cockpit instruments}
		\label{fig:benchmark-instruments}
	\end{subfigure}
	\hfill
	\begin{subfigure}[t]{0.35\textwidth}
		\centering
		\includegraphics[width=\textwidth]{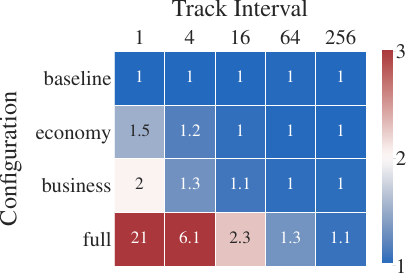}
		\caption{Overhead \cockpit configurations}
		\label{fig:benchmark_heatmap}
	\end{subfigure}
	\caption{\textbf{Run time overhead for individual \cockpittitle instruments and configurations}
		as shown on \cifarten \threecthreed on a GPU.
		\textit{Left:} The run time overheads for individual instruments are 
		shown as multiples of the \emph{baseline} (no tracking). Most instruments 
		add little overhead. This plot shows the overhead in one iteration, 
		determined by averaging over multiple iterations and random seeds.
		\textit{Right:} Overhead for different \cockpit configurations. Adjusting 
		the tracking interval and re-using the computation shared by multiple 
		instruments can make the overhead orders of magnitude smaller. Blue fields 
		mark settings that allow tracking without doubling the training time.}
	\label{fig:benchmark}
\end{figure}

\paragraph{Run time benchmark:}
\Cref{fig:benchmark-instruments} shows the wall-clock computational
overhead for individual instruments (details in
\Cref{app:benchmarks}).\footnote{To improve readability, we exclude 
  \texttt{HessMaxEV} here, because its overhead is large compared to other quantities. 
  Surprisingly, we also observed significant cost for the
  2D histogram on GPU. It is caused by an implementation 
  bottleneck for histogram shapes observed in deep models.
  We thus also omit \texttt{GradHist2d} here, as we expect it 
  to be eliminated with future implementations (see 
  \Cref{app:run-time-benchmarks} for a detailed analysis and
  further benchmarks).  Both quantities, however, are part of the benchmark shown 
  in \Cref{fig:benchmark_heatmap}.} As expected, byproducts are virtually free,
and quantities that rely solely on first-order information add little overhead
(at most roughly 25\,\% on this problem). Thanks to parallelization,
the ten extra backward passes required for Hessian quantities reduce to less 
than 100\,\% overhead. Individual overheads also do not
simply add up when multiple quantities are tracked, because quantities relying
on the same information share computations.

To allow a rough cost control, \cockpit currently offers three configurations,
called \emph{economy}, \emph{business}, and \emph{full}, in increasing order 
of cost (\cf \Cref{tab:overview-quantities}).
As a basic guideline, we consider a factor of two to be an acceptable limit for
the increase in training time and benchmark the configurations' run times for
different tracking intervals.
\Cref{fig:benchmark_heatmap} shows a run time matrix for the \cifarten
\threecthreed problem, where settings that meet this limit are set in blue 
(more problems including \imagenet are shown in \Cref{app:benchmarks}).
Speedups due to shared computations are easy to read off:
Summing all the individual overheads shown in \Cref{fig:benchmark-instruments}
would result in a total overhead larger than 200\,\%, while the joint overhead
(\textit{business}) reduces to 140\,\%.
The \textit{economy} configuration can easily be tracked at every step of this
problem and stay well below our threshold of doubling the execution time.
\cockpit's full view, shown in \Cref{fig:showcase}, can be updated every $64$-th
iteration without a major increase in training time (this corresponds to about
five updates per epoch).
Finally, tracking any configuration about once per epoch -- which is common in 
practice -- adds overhead close to zero (rightmost column).

This good performance is largely due to the efficiency of the \backpack
package \citep{Dangel2020}, which we leverage with custom and optimized modification, that
compacts information layer-wise and then discards unneeded buffers.
Using layer-wise information  (\Cref{sec:vanishing_gradient_exp}) scales better
to large networks, where storing the entire model's individual gradients all at
once becomes increasingly expensive (see \Cref{app:benchmarks}).
To the best of our knowledge, many of the quantities in
\Cref{tab:overview-quantities}, especially those relying on individual
gradients, have only been explored on rather small problems.
With \cockpit they can now be accessed at a reasonable rate for deep learning
models outside the toy problem category.

\section{Conclusion}
\label{sec:conclusion}
Contemporary machine learning, in particular deep learning, remains a craft and an art.
High dimensionality, stochasticity, and non-convexity require constant tracking
and tuning, often resulting in a painful process of trial and
error.
When things fail, popular performance measures, like the
training loss, do not provide enough information by themselves.
These metrics only tell \emph{whether} the model is learning, but not
\emph{why}.
Alternatively, traditional debugging tools can provide access to individual
weights and data.
However, in models whose power only arises from possessing myriad weights, 
this approach is hopeless, like looking for the proverbial needle in a haystack.

To mitigate this, we proposed \cockpit, a practical visual debugging tool 
for deep learning.
It offers instruments to monitor the network's internal dynamics during
training, in real-time.
In its presentation, we focused on two crucial factors affecting user
experience:
Firstly, such a debugger must provide meaningful insights.
To demonstrate \cockpit's utility, we showed how it can identify bugs where
traditional tools fail.
Secondly, it must come at a feasible computational cost.
Although \cockpit uses rich second-order information, efficient
computation keeps the necessary run time overhead cheap.
The open-source \pytorch package can be added to many existing training loops.

Obviously, such a tool is never complete. Just like there is no perfect
universal debugger, the list of current instruments is naturally
incomplete. Further practical experience with the tool, for example in the
form of a future larger user study, could provide additional evidence for its
utility.
However, our analysis shows that \cockpit provides useful tools
and extracts valuable information presently not accessible to the user.
We believe that this improves algorithmic interpretability -- 
helping practitioners understand how to make their models work --
but may also inspire new research.
The code is designed flexibly, deliberately separating the computation and 
visualization. New instruments can be added easily and
also be shown by the user's preferred visualization tool, \eg \tensorboard.
Of course, instead of just showing the data, the same information can be 
used by novel algorithms directly, side-stepping the human in the loop.

\begin{ack}
	The authors gratefully acknowledge financial support by the European Research Council
	through ERC StG Action 757275 / PANAMA; the DFG Cluster of Excellence “Machine
	Learning - New Perspectives for Science”, EXC 2064/1, project number 390727645;
	the German Federal Ministry of Education and Research (BMBF) through the Tübingen
	AI Center (FKZ: 01IS18039A); and funds from the Cyber Valley Initiative of the
	Ministry for Science, Research and Arts of the State of Baden-Württemberg.
	Moreover, the authors thank the International Max Planck Research School for
	Intelligent Systems (IMPRS-IS) for supporting Felix Dangel and Frank Schneider.
	Further, we are grateful to Agustinus Kristiadi, Alexandra Gessner, Christian
	Fröhlich, Filip de Roos, Jonathan Wenger, Julia Grosse, Lukas Tatzel, Marius
	Hobbhahn, and Nicholas Krämer for providing feedback to the manuscript.
\end{ack}

\medskip
{\small
	\bibliographystyle{bibliography/bib_style}

}

\newpage
\appendix
\clearpage

\appendixtitle

\section*{Checklist}
\label{sec:checklist}

\begin{enumerate}
	
	\item For all authors...
	\begin{enumerate}
		\item Do the main claims made in the abstract and introduction accurately reflect the paper's contributions and scope?
		\answerYes{}
		\item Did you describe the limitations of your work?
		\answerYes{} In \Cref{sec:benchmark} we detail the additional costs of each instrument, also showing that two of them come with a large overhead (more details and how it can be mitigated in \Cref{app:run-time-benchmarks}). \Cref{sec:conclusion} acknowledges that while we believe this tool to be an important step, it is understandably incomplete.
		\item Did you discuss any potential negative societal impacts of your work?
		\answerNA{} The paper proposes an algorithmic debugging tool that is of a foundational nature. Ethical questions are thus not sufficiently prominent in this work to warrant a dedicated discussion section. In general, we believe, this work will have an overall positive impact as it can help shed light into the black-box that is deep learning. As a longer-term side-effect of this work, this could help the explainability and interpretability of neural networks.
		\item Have you read the ethics review guidelines and ensured that your paper conforms to them?
		\answerYes{}
	\end{enumerate}
	
	\item If you are including theoretical results...
	\begin{enumerate}
		\item Did you state the full set of assumptions of all theoretical results?
		\answerNA{}
		\item Did you include complete proofs of all theoretical results?
		\answerNA{}
	\end{enumerate}
	
	\item If you ran experiments...
	\begin{enumerate}
		\item Did you include the code, data, and instructions needed to reproduce the main experimental results (either in the supplemental material or as a URL)?
		\answerYes{} All experimental results, as well as the complete code base to reproduce them can be found at the linked GitHub repository at \cockpitexpurl. The \cockpit package is available open source at \cockpiturl.
		\item Did you specify all the training details (e.g., data splits, hyperparameters, how they were chosen)?
		\answerYes{} All training details are given at \cockpitexpurl. If not stated otherwise, we use the defaults suggested by the \deepobs benchmark suite which are summarized in \Cref{app:benchmarks}.
		\item Did you report error bars (e.g., with respect to the random seed after running experiments multiple times)?
		\answerYes{} whenever applicable, we report error bars (\eg left subplot of \Cref{fig:benchmark} shows error bars from averages over ten random seeds).
		\item Did you include the total amount of compute and the type of resources used (e.g., type of GPUs, internal cluster, or cloud provider)?
		\answerYes{} listed in \Cref{app:benchmarks}
	\end{enumerate}
	
	\item If you are using existing assets (e.g., code, data, models) or curating/releasing new assets...
	\begin{enumerate}
		\item If your work uses existing assets, did you cite the creators?
		\answerYes{} We make extensive use of both the \backpack \citep{Dangel2020} and the \deepobs \citep{Schneider2019} packages. Both are cited throughout the text. Whenever applicable, we also cited the used data sets and models. We explicitly mention the authors of the used histogram code in \Cref{app:histograms} and have asked them for permissions.
		\item Did you mention the license of the assets?
		\answerYes{} The library has been released open source under the MIT License. \cockpit's GitHub repository includes the full license.
		\item Did you include any new assets either in the supplemental material or as a URL?
		\answerYes{} The full library can be found at \cockpiturl.
		\item Did you discuss whether and how consent was obtained from people whose data you're using/curating?
		\answerNA{} 
		\item Did you discuss whether the data you are using/curating contains personally identifiable information or offensive content?
		\answerNA{} No new data was collected with our experiments relying on established and published data sets such as \mnist \citep{Lecun1998}.
	\end{enumerate}
	
	\item If you used crowdsourcing or conducted research with human subjects...
	\begin{enumerate}
		\item Did you include the full text of instructions given to participants and screenshots, if applicable?
		\answerNA{}
		\item Did you describe any potential participant risks, with links to Institutional Review Board (IRB) approvals, if applicable?
		\answerNA{}
		\item Did you include the estimated hourly wage paid to participants and the total amount spent on participant compensation?
		\answerNA{}
	\end{enumerate}
	
\end{enumerate}

\section{Code example}
\label{app:code_example}
One design principle of \cockpit is its easy integration with conventional \pytorch
training loops. \Cref{fig:Codeblock} shows a working example of a standard
training loop with \cockpit integration. More examples and tutorials are described 
in \cockpit's documentation.
\cockpit's syntax is inspired by \backpack: It can be used interchangeably with the 
library responsible for most back-end computations. 
Changes to the code are straightforward:
\begin{itemize}
	\item \textbf{Importing} (\textit{Lines 5, 7} and \textit{8}): Besides importing \cockpit
	  we also need to import \backpack which is required for extending (parts of) the
	  model (see next step).
	\item \textbf{Extending} (\textit{Lines 11} and \textit{12}): When defining the model and the
	  loss function, we need to \emph{extend} both of them using \backpack. This is as trivial as
	  wrapping them in the \texttt{extend()} function provided by \backpack and lets \backpack know
	  that additional quantities (such as the individual gradients) should be computed for them.
	  Note, that while applying \backpack is easy, it currently does not support all possible model
	  architectures and layer types. Specifically, \emph{batch norm} layers are not supported
	  since using them results in ill-defined individual gradients.
	\item \textbf{Individual losses} (\textit{Line 13}): For the \texttt{Alpha} quantity, \cockpit
	  also requires the individual loss values (to estimate the variance of the loss estimate).
	  This can be computed cheaply but is not usually part of a conventional training loop. Creating
	  this loss is done analogously to creating any other loss, with the only exception of setting
	  \texttt{reduction="none"}. Since we don't differentiate this loss, we don't need to extend it.
	\item \textbf{Cockpit configuration} (\textit{Line 16} and \textit{17}): Initializing the \cockpit
	  requires passing them (extended) model parameters as well as a list of quantities that should be
	  tracked. \Cref{tab:overview-quantities} provides an overview of all possible quantities. In this
	  example, we use one of the pre-defined configurations offered by \cockpit.
	  Separately, we initialize the plotting part of \cockpit. We deliberately detached the visualization
	  from the tracking to allow greater flexibility.
	\item \textbf{Quantity computation} (\textit{Line 27} and \textit{38}): Performing the training is 
	  very similar to a regular training loop, with the only difference being that the backward pass
	  should be surrounded by the \cockpit context (\texttt{with cockpit():}).
	  Additionally to the \texttt{global\_step} we also pass a few additional information to the \cockpit
	  that are computed anyway and can be re-used by the \cockpit, such as the batch size, the individual
	  losses, or the optimizer itself.
	  After the backward pass (when the context is left) all \cockpit quantities are automatically computed.
	\item \textbf{Logging and visualizing} (\textit{Line 46} and \textit{47}): At any point during the training,
	  here we do it at the end, we can write all quantities to a log file. We can use this log file, or alternatively
	  the \cockpit directly, to visualize all quantities which would result in a status screen similar to
	 \Cref{fig:showcase}.
\end{itemize}

\begin{figure}[ht]
\begin{minted}{python}
"""Example: Training Loop using Cockpit."""

import torch
from _utils_examples import cnn, fmnist_data, get_logpath
from backpack import extend
from cockpit import Cockpit, CockpitPlotter
from cockpit.utils.configuration import configuration as config

fmnist_data = fmnist_data()
model = extend(cnn())
loss_fn = extend(torch.nn.CrossEntropyLoss(reduction="mean"))
losses_fn = torch.nn.CrossEntropyLoss(reduction="none")
opt = torch.optim.SGD(model.parameters(), lr=1e-2)

cockpit = Cockpit(model.parameters(), quantities=config("full"))
plotter = CockpitPlotter()

max_steps, global_step = 50, 0
for inputs, labels in iter(fmnist_data):
    opt.zero_grad()

    outputs = model(inputs)
    loss = loss_fn(outputs, labels)
    losses = losses_fn(outputs, labels)

    with cockpit(
        global_step,
        info={
            "batch_size": inputs.shape[0],
            "individual_losses": losses,
            "loss": loss,
            "optimizer": opt,
        },
    ):
        loss.backward(
            create_graph=cockpit.create_graph(global_step),
        )

    opt.step()
    global_step += 1

    if global_step >= max_steps:
        break

cockpit.write(get_logpath())
plotter.plot(get_logpath())
\end{minted}
	\caption{\textbf{Complete training loop with \cockpittitle} in
    \pytorch. Line changes are highlighted in light orange
    (\textcolor{sns_orange_light}{\ding{122}}).}
	\label{fig:Codeblock}
\end{figure}

\clearpage

\section{\cockpittitle instruments overview}
\label{app:cockpit_features}
\Cref{tab:feature-table} lists all quantities available in the first public
release of \cockpit. If necessary, we provide references to their
mathematical definition. This table contains additional quantities, compared to
\Cref{tab:overview-quantities} in the main text. To improve the presentation of
this work, we decided to not describe every quantity available in \cockpit in
the main part and instead focus on the investigated metrics. Custom quantities
can be added easily without having to understand the inner-workings.

{\def\arraystretch{1.2}
\begin{table*}[ht]
	\caption{\textbf{Overview of all \cockpittitle quantities} with a short
		description and, if necessary, a reference to mathematical definition.}
	\label{tab:feature-table}
	\begin{center}
		\begin{tabularx}{\textwidth}{ lXc }
			\toprule
			\textbf{Name}          & \textbf{Description}                                                                                                         & \textbf{Math}                               \\ \midrule
			\texttt{Loss}          & Mini-batch training loss at current iteration, $\gL_{\gB}(\vtheta)$                                                          & (\ref{eq:mini-batch-loss})                  \\
			\texttt{Parameters}    & Parameter values $\vtheta_{t}$ at the current iteration                                                                      & -                                           \\
			\texttt{Distance}      & $L_2$ distance from initialization $\lVert \vtheta_{t} -  \vtheta_{0} \rVert_2$                                              & -                                           \\
			\texttt{UpdateSize}    & Update size of the current iteration $\lVert \vtheta_{t + 1} -  \vtheta_{t} \rVert_2$                                        & \multicolumn{1}{l}{}                        \\
			\texttt{GradNorm}      & Mini-batch gradient norm $\lVert \vg_{\gB}(\vtheta) \rVert_2$                                                                & -                                           \\
			\texttt{Time}          & Time of the current iteration \newline (\eg used in benchmark of \Cref{app:benchmarks})                                      & -                                           \\
			\texttt{Alpha}         & Normalized step on a noisy quadratic interpolation between two iterates $\vtheta_t, \vtheta_{t+1}$                           & (\ref{eq:alpha-feature-table})              \\
			\texttt{CABS}          & Adaptive batch size for \sgd, optimizes expected objective gain per cost, adapted from \citep{Balles2017}                    & (\ref{eq:cabs-feature-table})               \\
			\texttt{EarlyStopping} & Evidence-based early stopping criterion for \sgd, \newline proposed in \citep{Mahsereci2017}                                 & (\ref{eq:early-stopping-feature-table})     \\
			\texttt{GradHist1d}    & Histogram of individual gradient elements, $\{ \vg_n(\vtheta_j) \}_{n\in \gB}^{j = 1, \dots, D}$                             & (\ref{eq:app-grad-hist-1d})                 \\
			\texttt{GradHist2d}    & Histogram of weights and individual gradient elements, $\{ ( \vtheta_j, \vg_n(\vtheta_j) ) \}_{n\in \gB}^{ j = 1, \dots, D}$ & (\ref{eq:app-grad-hist-2d})                 \\
			\texttt{NormTest}      & Normalized fluctuations of the residual norms $\lVert  \vg_{\gB} - \vg_n \rVert$, \newline proposed in \citep{Byrd2012}      & (\ref{eq:norm-test-feature-table})          \\
			\texttt{InnerTest}     & Normalized fluctuations of $\vg_n$'s parallel components along $\vg_{\gB}$, \newline proposed in \citep{Bollapragada2017}    & (\ref{eq:inner-product-test-feature-table}) \\
			\texttt{OrthoTest}     & Normalized fluctuations of $\vg_n$'s orthogonal components along $\vg_{\gB}$, \newline proposed in \citep{Bollapragada2017}  & (\ref{eq:orthogonality-test-feature-table}) \\
			\texttt{HessMaxEV}     & Maximum Hessian eigenvalue, $ \lambda_{\text{max}}(\mH_{\gB}(\vtheta))$, inspired by \citep{Yao2020}                         & (\ref{eq:hess-max-ev-feature-table})        \\
			\texttt{HessTrace}     & Exact or approximate Hessian trace, $\Tr(\mH_{\gB}(\vtheta))$, inspired by \citep{Yao2020}                                   & -                                           \\
			\texttt{TICDiag}       & Relation between (diagonal) curvature and gradient noise, \newline inspired by \citep{Thomas2020}                            & (\ref{eq:tic-diag-feature-table})           \\
			\texttt{TICTrace}      & Relation between curvature and gradient noise trace, \newline inspired by \citep{Thomas2020}                                 & (\ref{eq:tic-trace-feature-table})          \\
			\texttt{MeanGSNR}      & Average gradient signal-to-noise-ratio (GSNR), inspired by \citep{Liu2020}                                                   & (\ref{eq:mean-gsnr-feature-table})          \\ \bottomrule
		\end{tabularx}
	\end{center}
\end{table*}
}

\clearpage

\section{Mathematical details}
\label{app:instruments}

In this section, we want to provide the mathematical background for each
instrument described in \Cref{tab:feature-table}.
This complements the more informal description presented in
\Cref{sec:instruments} in the main text, which focused more on the
expressiveness of the individual quantities.
We will start by setting up the necessary notation in addition to the one
introduced in \Cref{sec:instruments}. 
\subsection{Additional notation}
\label{app:notation}

\paragraph{Population properties:}
The population risk $\gL_{P}(\vtheta) \in \sR$ and its variance
$\Lambda(\vtheta) \in \sR$ are given by
\begin{subequations}
\begin{align}
  \label{eq:population-risk}
  \gL_{P}(\vtheta)
  &= \E_{(\vx, \vy)\sim P}\left[ \ell(f(\vtheta, \vx), \vy)  \right]
    = \int \ell(f(\vtheta, \vx), \vy) P(\vx, \vy)\:d\vx\:d\vy\,,
  \\
  \label{eq:population-risk-variance}
  \begin{split}
    \Lambda_P(\vtheta)
    &= \Var_{(\vx, \vy)\sim P}\left[ \ell(f(\vtheta, \vx), \vy)  \right]
    = \int \left(
      \ell(f(\vtheta, \vx), \vy) - \gL_{P}(\vtheta)
    \right)^2 P(\vx, \vy) \:d\vx\:d\vy\,.
  \end{split}
\end{align}
\end{subequations}
The population gradient $\vg_{P}(\vtheta) \in \sR^{D}$ and its variance
$\mSigma_{P}(\vtheta) \in \sR^{D\times D}$ are given by
\begin{subequations}
  \begin{align}
    \label{eq:population-gradient}
    \begin{split}
      \vg_{P}(\vtheta)
      &= \E_{(\vx, \vy)\sim P}\left[\nabla_\vtheta \ell(f(\vtheta, \vx), \vy)  \right]
      = \int \nabla_\vtheta\ell(f(\vtheta, \vx), \vy) P(\vx, \vy)\:d\vx\:d\vy\,,
    \end{split}
    \\
    \label{eq:population-gradient-variance}
    \begin{split}
      \mSigma_P(\vtheta)
      &= \Var_{(\vx, \vy)\sim P}\left[\nabla_\vtheta \ell(f(\vtheta, \vx), \vy)  \right]
      \\
      &= \int \left(
        \nabla_\vtheta\ell(f(\vtheta, \vx), \vy) - \vg_{P}(\vtheta)
      \right)
      \left(
        \nabla_\vtheta\ell(f(\vtheta, \vx), \vy) - \vg_{P}(\vtheta)
      \right)^\top P(\vx, \vy)
      \:d\vx\:d\vy\,.
    \end{split}
  \end{align}
\end{subequations}

\paragraph{Empirical approximations:}
Let $\gS$ denote a set of samples drawn i.i.d.\,from $P$, \ie $\gS = \left\{ (\vx_i,
  \vy_i)\:|\:i = 1, \dots, |\gS| \right\}$. With a slight abuse of notation the
empirical risk approximated with $\gS$ is
\begin{subequations}
  \begin{equation}
    \label{eq:empirical-risk-approximation}
    \gL_{\gS}(\vtheta)
    = \frac{1}{|\gS|} \sum_{n \in \gS}  \ell_n(\vtheta)
  \end{equation}
  (later, $\gS$ will represent either a mini-batch $\gB$, or the train set $\gD$).
  The empirical risk gradient $\vg_{\gS}(\vtheta) \in \sR^{D}$ on $\gS$ is
  \begin{equation}
    \label{eq:empirical-risk-gradient-approximation}
    \vg_{\gS}(\vtheta)
    = \nabla_{\vtheta}\gL_{\gS}(\vtheta)
    = \frac{1}{|\gS|} \sum_{n \in \gS}  \nabla_{\vtheta}\ell_n(\vtheta)
    = \frac{1}{|\gS|} \sum_{n \in \gS}  \vg_n(\vtheta)\,,
  \end{equation}

\end{subequations}
with individual gradients $\vg_n(\vtheta) =
\nabla_{\vtheta}\ell_n(\vtheta) \in \sR^D$ implied by a sample $n$.
Population risk and gradient variances $\Lambda_P(\vtheta),
\mSigma_P(\vtheta)$ can be empirically estimated on $\gS$ with the sample
variances $\hat{\Lambda}_\gS(\vtheta) \in \sR, \hat{\mSigma}_\gS(\vtheta) \in
\sR^{D\times D}$, given by
\begin{subequations}
  \label{equ:population-risk-gradient}
  \begin{align}
    \label{eq:population-risk-variance-estimator}
    \Lambda_P(\vtheta)
    &\approx \frac{1}{|S| - 1} \sum_{n\in\gS} \left(
      \ell_n(\vtheta) - \gL_{\gS}(\vtheta)
      \right)^2
      := \hat{\Lambda}_\gS(\vtheta)\,,
    \\
    \label{eq:population-risk-gradient-variance-estimator}
    \begin{split}
      \mSigma_P(\vtheta)
      &\approx \frac{1}{|\gS| - 1} \sum_{n\in\gS} \left(
        \vg_n(\vtheta) - \vg_{\gS}(\vtheta)
      \right)
      \left(
        \vg_n(\vtheta) - \vg_{\gS}(\vtheta)
      \right)^\top
      := \hat{\mSigma}_\gS(\vtheta)
      \\
      &\approx
      \frac{1}{{|\gS| - 1}}
      \left[
        \left(
          \sum_{n\in\gS}
          \vg_n(\vtheta)
          \vg_n(\vtheta)^\top
        \right)
        -
        |\gS|
        \vg_\gS(\vtheta)
        \vg_\gS(\vtheta)^\top
      \right]\,.
    \end{split}
  \end{align}
\end{subequations}
Often, gradient elements are assumed independent and hence their
variance is diagonal ($^{\odot 2}$
denotes element-wise square),
\begin{align}
  \label{eq:cheat-sheet-risk-gradient-diagonal-variance-estimator}
  \mathrm{diag}\!\left(\mSigma_P(\vtheta)\right)
  &\approx \frac{1}{|S| - 1} \sum_{n\in\gS}
    \left(
    \vg_n(\vtheta) - \vg_{\gS}(\vtheta)
    \right)^{\odot 2}
    = \mathrm{diag}\!\left(\hat{\mSigma}_\gS(\vtheta)\right) \in \sR^D\,.
\end{align}

\paragraph{Slicing:} To avoid confusion between $\vtheta_t$ (parameter at
iteration $t$) and $\vtheta_j$ ($j$-th parameter entry), we denote the latter as $[\vtheta]_j$.

\subsection{Normalized Step Length (\texttt{Alpha})}
\label{app:alpha}

\begin{figure}
  \includegraphics[width =
  \textwidth]{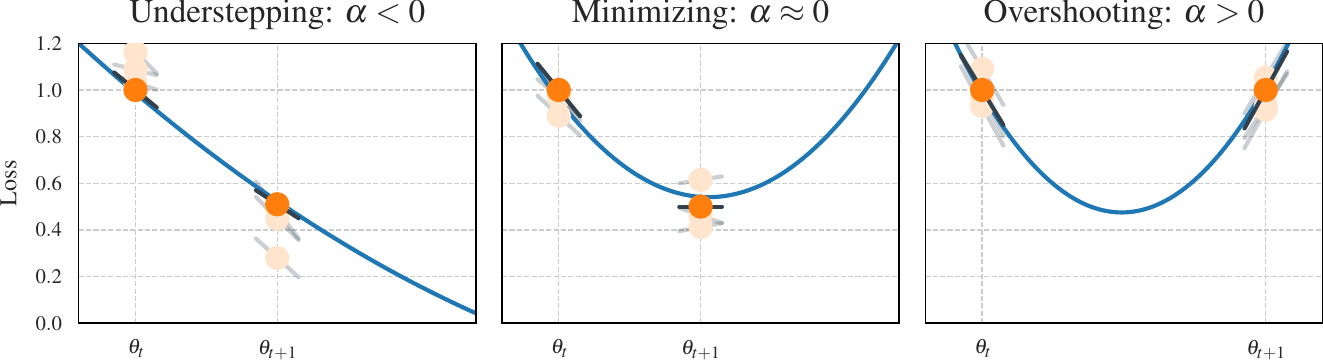}
  \caption{\textbf{Motivational sketch for the $\alpha$ quantity.}
    In each iteration of the optimizer we observe the loss function at two
    positions $\vtheta_{t}$ and $\vtheta_{t+1}$ (shown in
    \textcolor{sns_orange}{\ding{108}}). The black lines
    (\textcolor{TUdark}{\textbf{---}}) show the observed slope at this
    position, which we can get from projecting the gradients onto the current
    step direction $\vtheta_{t+1} - \vtheta_{t}$. Note, that all four
    observations (two loss and two slope values) are noisy, due to being
    computed on a mini-batch. With access to the individual losses and
    gradients (some samples shown in
    \textcolor{sns_orange_light}{\ding{108}}/\textcolor{TUgray}{\textbf{---}}),
    we can estimate their noise level and build a noise-informed
    quadratic fit (\textcolor{sns_blue}{\textbf{---}}). Using this fit, we
    determine whether the optimizer minimizes the local univariate loss
    (\textit{middle plot}), or whether we understep (\textit{left plot}) or
    overshoot (\textit{right plot}) the minimum.}
  \label{fig:alpha_explanation}
\end{figure}

\paragraph{Motivation:} The goal of the $\alpha$-quantity is to estimate and
quantify the effect that a selected learning rate has on the optimizer's steps.
Let's consider the step that the optimizer takes at training iteration $t$. This
parameter update from $\vtheta_{t}$ to $\vtheta_{t+1}$ happens in a
one-dimensional space, defined by the update direction $\vtheta_{t+1} -
\vtheta_{t}=\vs_t$. The update direction depends on the update rule of the
optimizer, \eg for \sgd with learning rate $\eta$ it is simply $\vs_t =- \eta
\vg_{\gB_t}(\vtheta_{t})$.

We build a noise-informed univariate quadratic approximation along this update
step ($\vtheta_{t} \to \vtheta_{t+1}$) based on the two noisy loss function
observations at $\vtheta_{t}$ and $\vtheta_{t+1}$ and the two noisy slope
observation at these two points.
Examining this quadratic fit, we are able to determine where on this parabola
our optimizer steps.
Standardizing this, we express a step to the minimum of the loss in the update
direction as $\alpha=0$.
Analogously, steps that end short of this minimum result in $\alpha<0$, and a
step over the minimum in $\alpha>0$.
These three different scenarios are illustrated in \Cref{fig:alpha_explanation}
also showing the underlying observations that would lead to them.
\Cref{fig:LINE} shows the distribution of $\alpha$-values for two very
different optimization trajectories.

\paragraph{Noisy observations:} In order to build an approximation for the loss
function in the update direction, we leverage the four observations of the
function (and its derivative) that are available in each iteration.
Due to the stochasticity of deep learning optimization, we also take into
account the noise-level of all observations by estimating them.
The first two observations are the mini-batch training losses
$\mathcal{L}_{\gB_t}(\vtheta_t), \mathcal{L}_{\gB_{t+1}}(\vtheta_{t+1})$ at point $\vtheta_{t}$ and $\vtheta_{t+1}$,
which are computed in every standard training loop.
The mini-batch losses are averages over individual losses,
\begin{align*}
  \mathcal{L}_{\gB_t}(\vtheta_t)
  &=
    \E_{\gB_t}\left[ \ell(\vtheta_t) \right] = \frac{1}{|\gB_t|} \sum_{n \in \gB_t}
    \ell_n(\vtheta_t) \,,
  \\
  \mathcal{L}_{\gB_{t+1}}(\vtheta_{t+1})
  &=
    \E_{\gB_{t+1}}\left[ \ell(\vtheta_{t+1}) \right] = \frac{1}{|\gB_{t+1}|} \sum_{n \in \gB_{t+1}}
    \ell_n(\vtheta_{t+1})\,,
\end{align*}
and using these individual losses, we can also compute the variances
to estimate the noise-level of our loss observation,
\begin{align*}
  \Var_{\gB_{t}} \left[ \ell(\vtheta_t) \right]
  =&
     \left( \frac{1}{B_t} \sum_{n \in \gB_t} \ell_n(\vtheta_t)^2 \right)
     -
     \left( \frac{1}{B_t} \sum_{n \in \gB_t} \ell_n(\vtheta_t) \right)^2 \, ,
  \\
  \Var_{\gB_{t+1}} \left[ \ell(\vtheta_{t+1}) \right]
  =&
     \left( \frac{1}{|\gB_{t+1}|} \sum_{n \in \gB_{t+1}} \ell_n(\vtheta_{t+1})^2 \right)
     -
     \left( \frac{1}{|\gB_{t+1}|} \sum_{n \in \gB_{t+1}} \ell_n(\vtheta_{t+1}) \right)^2 \, .
\end{align*}
Similarly, we proceed with the slope in the update direction.
To compute the slope of the loss function in the direction of the optimizer's
update $\vs_t$, we project the current gradient along this update direction
\begin{align*}
  \E_{\gB_t} \left[\frac{\vs_t^\top \vg(\vtheta_t)}{\lVert \vs_t
  \rVert^2}\right]
  =&
  \frac{1}{|\gB_{t}|} \sum_{n\in \gB_t} \frac{\vs_t^\top \vg_n(\vtheta_t)}{\lVert
  \vs_t \rVert^2} \, ,
  \\
  \E_{\gB_{t+1}} \left[\frac{\vs_t^\top \vg(\vtheta_{t+1})}{\lVert \vs_t
  \rVert^2}\right]
  =&
  \frac{1}{|\gB_{t+1}|} \sum_{n\in \gB_{t+1}} \frac{\vs_t^\top \vg_n(\vtheta_{t+1})}{\lVert
  \vs_t \rVert^2} \, .
\end{align*}
Just like before, we can also compute the variance of this slope, by leveraging
individual gradients,
\begin{align*}
  \Var_{\gB_t} \left[\frac{\vs_t^\top \vg(\vtheta_t)}{\lVert \vs_t
  \rVert^2}\right]
  =&
  \frac{1}{|\gB_t|} \sum_{n\in B_t} \left( \frac{\vs_t^\top \vg_n(\vtheta_t)}{\lVert
  \vs_t \rVert^2} \right)^2
  -  \left( \frac{1}{|\gB_t|} \sum_{n\in \gB_t} \frac{\vs_t^\top
  \vg_n(\vtheta_t)}{\lVert \vs_t \rVert^2} \right)^2 \, ,
  \\
  \Var_{\gB_{t+1}} \left[\frac{\vs_t^\top \vg(\vtheta_{t+1})}{\lVert \vs_t
  \rVert^2}\right]
  =&
  \frac{1}{|\gB_{t+1}|} \sum_{n\in \gB_{t+1}} \left( \frac{\vs_t^\top \vg_n(\vtheta_{t+1})}{\lVert
  \vs_t \rVert^2} \right)^2
  -  \left( \frac{1}{|\gB_{t+1}|} \sum_{n\in \gB_{t+1}} \frac{\vs_t^\top
  \vg_n(\vtheta_{t+1})}{\lVert \vs_t \rVert^2} \right)^2 \, .
\end{align*}

\paragraph{Quadratic fit \& normalization:} Using our (noisy) observations, we
are now ready to build an approximation for the loss as a function of the step
size, which we will denote as $f(\tau)$.
We assume a quadratic function for $f$, which follows recent reports for the
loss landscape of neural networks \citep{Xing2018}, \ie a function $f(\tau) =
w_0 + w_1 \tau + w_2 \tau^2$ parameterized by $\vw \in \R^3$.
We further assume a Gaussian likelihood of the form
\begin{align}
  \label{eq:alpha_likelihood}
  p\left(\tilde{\vf} \vert
  \vw, \mPhi\right) = \mathcal{N}\left(\tilde{\vf}; \mPhi^\top \vw,
  \mLambda\right)
\end{align}
for observations $\tilde{\vf}$ of the loss and its slope.
The observation matrix $\mPhi$ and the noise matrix of the observations
$\mLambda$ are
\begin{align*}
  \mPhi = \begin{pmatrix}
    1 & 1 & 0 & 0 \\
    \tau_1 & \tau_2  & 1 & 1 \\
    \tau_1^2 & \tau_2^2 & 2\tau_1 &  2\tau_2
  \end{pmatrix}\,,
                                    \qquad \qquad
                                    \mLambda = \begin{pmatrix}
                                      \sigma_{\tilde{f}_1} & 0 & 0 & 0 \\ 0 & \sigma_{\tilde{f}_2} & 0 & 0  \\ 0
                                      & 0 & \sigma_{\tilde{f}'_1} & 0 \\ 0 & 0 & 0 & \sigma_{\tilde{f}'_2}
                                    \end{pmatrix} \, ,
\end{align*}
where $\tau$ denotes the position and $\sigma$ denotes the noise-level estimate
of the observation.
The maximum likelihood solution of \Cref{eq:alpha_likelihood} for the
parameters of our quadratic fit is given by
\begin{align}
  \label{eq:alpha-feature-table}
  \vw = \left(\mPhi \mLambda^{-1}\mPhi^\top\right)\mPhi \mLambda^{-1}
  \tilde{\vf} \, .
\end{align}
Once we have the quadratic fit of the univariate loss function in the update
direction, we normalize the scales such that the resulting $\alpha$-value
expresses the effective step taken by the optimizer sketched in \Cref{fig:alpha_explanation}.

\paragraph{Usage:}
The $\alpha$-quantity is related to recent line search approaches
\cite{Mahsereci2017a,Vaswani2019}.
However, instead of searching for an acceptable step by repeated attempts, we
instead report the effect of the current step size selection.
This could, for example, be used to disentangle the two optimization runs in
\Cref{fig:LINE}.
Additionally, this information could also be used to automatically adapt the
learning rate during the training process.
But, as discussed in \Cref{sec:alpha_exp}, it isn't trivial what the
``correct'' decision is, as it might depend on the optimization problem, the
training phase, and other factors.
Having this $\alpha$-quantity can, however, provide more insight into what
kind of steps are used in well-tuned runs with traditional optimizers such
as \sgd.

\subsection{\cabs criterion: Coupling adaptive batch sizes with learning rates
  (\texttt{CABS})}
\label{app:cabs}

The \cabs criterion, proposed by \citet{Balles2017}, can be used to adapt the
mini-batch size during training with \sgd. It
relies on the gradient noise and approximately optimizes the objective's
expected gain per cost. The adaptation rule is (with learning rate $\eta$)
\begin{equation}
  \label{eq:cabs-theoretical}
  |\gB| \leftarrow \eta \frac{\Tr(\mSigma_P(\vtheta))}{\gL_{P}(\vtheta)}\,,
\end{equation}
and the practical implementation approximates $ \gL_{P}(\vtheta) \approx
\gL_{\gB}(\vtheta), \Tr(\mSigma_P(\vtheta)) \approx \frac{|\gB|-1}{|\gB|}
\Tr(\hat{\mSigma}_\gB(\vtheta))$ (compare equations (10, 22) and first paragraph
of Section 4 in \citep{Balles2017}). This yields the quantity computed in
cockpit's \texttt{CABS} instrument,
\begin{align}
  \label{eq:cabs-feature-table}
  |\gB| &\leftarrow \eta \frac{
    \frac{1}{|\gB|}
    \sum_{j=1}^D \sum_{n\in\gB}
    \left[
      \vg_n(\vtheta) - \vg_{\gB}(\vtheta)
    \right]_j^2
  }{
    \gL_{\gB}(\vtheta)
  }\,.
\end{align}

\paragraph{Usage:} The \cabs criterion suggests a batch size which is optimal
under certain assumptions. This suggestion can support practitioners in the
batch size selection for their deep learning task.

\subsection{Early-stopping criterion for SGD (\texttt{EarlyStopping})}
\label{app:early-stopping}

The empirical risk $\gL_{\gD}(\vtheta)$, and the mini-batch loss
$\gL_{\gB}(\vtheta)$ are only estimators of the target objective
$\gL_{P}(\vtheta)$. \citet{Mahsereci2017} motivate
$p(\vg_{\gB,\gD}(\vtheta)\:|\:\vg_{P}(\vtheta) = \vzero)$ as a measure for
detecting noise in the finite data sets $\gB, \gD$ due to sampling from $P$. They propose an
evidence-based (EB) criterion for early stopping the training procedure based on
mini-batch statistics, and model $p(\vg_{\gB}(\vtheta))$ with a sampled diagonal
variance approximation (compare
\Cref{eq:cheat-sheet-risk-gradient-diagonal-variance-estimator}),
\begin{equation}
  p(\vg_{\gB}(\vtheta))
  \approx \prod_{j=1}^D  \gN\left(
    \left[\vg_{P}(\vtheta)\right]_j;
    \frac{\left[ \hat{\mSigma}_\gB(\vtheta)\right]_{j,j}}{|\gB|}
  \right)\,.
\end{equation}
Their \sgd stopping criterion is
\begin{subequations}
\begin{align}
  \frac{2}{D} \left[ \log p(\vg_{\gB}(\vtheta))
    - \E_{\vg_{\gB}(\vtheta) \sim p(\vg_{\gB}(\vtheta))} \left[ \log p(\vg_{\gB}(\vtheta))\right]
  \right]
  &> 0\,,
  \intertext{and translates into}
      1 - \frac{|\gB|}{D} \sum_{j=1}^D \frac{
        \left[ \vg_{\gB}(\vtheta)\right]_j^2
      }{
        \left[ \hat{\mSigma}_\gB(\vtheta)\right]_{j,j}
      }
      &> 0\,,
      \\
      1 - \frac{|\gB|}{D} \sum_{d=1}^D \frac{
        \left[ \vg_{\gB}(\vtheta)\right]_d^2
      }{
        \frac{1}{|\gB| - 1}
        \sum_{n\in \gB}
        \left[
          \vg_n(\vtheta) - \vg_{\gB}(\vtheta)
        \right]_d^2
      }
      &> 0\,,
      \\
      \label{eq:early-stopping-feature-table}
      1 - \frac{|\gB| (|\gB| - 1)}{D} \sum_{d=1}^D \frac{
        \left[ \vg_{\gB}(\vtheta)\right]_d^2
      }{
        \left(
          \sum_{n\in \gB}
          \left[
            \vg_n(\vtheta)
          \right]_d^2
        \right)
        - |\gB|
        \left[
          \vg_{\gB}(\vtheta)
        \right]_d^2
      }
      &> 0\,.
  \end{align}
\end{subequations}
\cockpit's \texttt{EarlyStopping} quantity computes the left-hand side of
\Cref{eq:early-stopping-feature-table}.

\paragraph{Usage:} The \texttt{EarlyStopping} quantity of \cockpit can inform
the practitioner that training is about to be completed and the model might be
at risk of overfitting.

\subsection{Individual gradient element histograms (\texttt{GradHist1d},
  \texttt{GradHist2d})}
For the $|\gB| \times D$ individual gradient elements, \cockpit's
\texttt{GradHist1d} instrument displays a histogram of
\begin{equation}
  \label{eq:app-grad-hist-1d}
  \left\{
      \vg_n(\vtheta_j)
  \right\}_{n\in\gB,j=1,\dots, D}\,.
\end{equation}
\cockpit's \texttt{GradHist2d} instrument displays a two-dimensional histogram of the
$|\gB| \times D$ tuples
\begin{equation}
  \label{eq:app-grad-hist-2d}
  \left\{
    \left(
      \vtheta_j,
    \vg_n(\vtheta_j)
    \right)
  \right\}_{n\in\gB,j=1,\dots, D}\,
\end{equation}
and the marginalized one-dimensional histograms over the parameter and gradient axes.

\paragraph{Usage:} \Cref{sec:misscaled_data_exp,sec:vanishing_gradient_exp}
provide use cases (identifying data pre-processing issues and vanishing gradients)
for both the gradient histogram as well as its layer-wise extension.

\subsection{Gradient tests (\texttt{NormTest}, \texttt{InnerTest}, \texttt{OrthoTest})}
\label{app:gradient_tests}

\citet{Bollapragada2017} and \citet{Byrd2012} propose batch size adaptation
schemes based on the gradient noise. They formulate geometric
constraints between population and mini-batch gradient and accessible approximations that can be probed to decide
whether the mini-batch size should be increased. Because mini-batches are
i.i.d.\,from $P$, it holds that
\begin{subequations}
  \label{eq:iid-sampling-expectation}
  \begin{align}
    \E\left[\vg_{\gB}(\vtheta) \right]
    &=
      \vg_{P}(\vtheta),
    \\
    \E\left[ \vg_{\gB}(\vtheta)^\top \vg_{P}(\vtheta)  \right]
    &=
      \lVert \vg_{P}(\vtheta)  \rVert^2.
  \end{align}
\end{subequations}

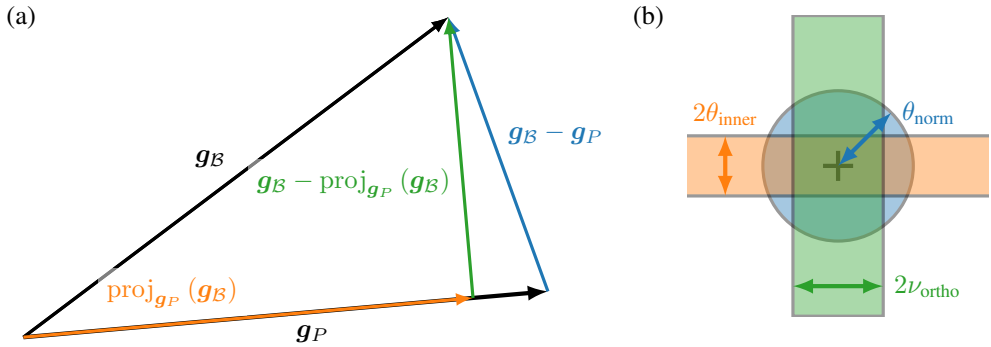
\begin{figure}[ht]
  \begin{minipage}[t]{0.59\linewidth}
    \centering
    \begin{flushleft}
      (a)
      \vspace{-\baselineskip}
    \end{flushleft}
    \begin{tikzpicture}[rotate=5, >=latex, very thick, xscale = 1., yscale =
      1.25]

      \tikzset{my label style/.style={fill=white, fill opacity=0.5, text
          opacity=1}}

      \draw[->, ultra thick] (0,0) to node [midway, anchor=north west, my label
      style] {$\vg_{P}$} (7,0);

      \draw[->] (0,0) to node [midway, anchor=south east, my label style]
      {$\vg_{\gB}$} (6,3);
      \draw[->, >=latex] (0,0) -- (6,3);

      \draw[->, sns_blue] (7,0) to node [midway, right, anchor=south west, my label
      style] {$\vg_{\gB} - \vg_{P}$} (6,3);

      \draw[->, sns_orange] (0,0) to node [midway, above, anchor=south east, my label
      style] {$\mathrm{proj}_{\vg_{P}}\left(\vg_{\gB}\right)$} (6,0);

      \draw[->, sns_green] (6,0) to node [midway, left, anchor=north east, my label
      style] {$\vg_{\gB} - \mathrm{proj}_{\vg_{P}}\left(\vg_{\gB}\right)$} (6,3);
    \end{tikzpicture}
  \end{minipage}
  \begin{minipage}[t]{0.39\linewidth}
    \centering
    \begin{flushleft}
      (b)
      \vspace{-\baselineskip}
    \end{flushleft}
    \begin{tikzpicture}[>=latex, very thick, xscale = 2, yscale = 2]
      \clip (-1,0) rectangle (1,2);

      \draw[ultra thick] (0,0.9) to (0,1.1);
      \draw[ultra thick] (-0.1,1) to (0.1,1);

      \pgfmathsetmacro{\normTestRadius}{0.5}
      \filldraw [fill=sns_blue, opacity=0.4] (0,1) circle (\normTestRadius);

      \pgfmathsetmacro{\innerProductWidth}{0.2}
      \filldraw [fill=sns_orange, opacity=0.4] (-2,1 - \innerProductWidth) rectangle
      (2,1 + \innerProductWidth);

      \pgfmathsetmacro{\orthogonalityTestWidth}{0.3}
      \filldraw [fill=sns_green, opacity=0.4] (-\orthogonalityTestWidth, 0) rectangle
      (\orthogonalityTestWidth, 2);

      \draw[ultra thick, <->, sns_blue] (0,1) to ++(45:\normTestRadius) node [above, right] {$\theta_{\text{norm}}$};

      \draw[ultra thick, <->, sns_orange] (-0.75, 1 - \innerProductWidth) to ++(0, 2 *\innerProductWidth) node [above] {$2 \theta_{\text{inner}}$};

      \draw[ultra thick, <->, sns_green] (-\orthogonalityTestWidth0, 0.2) to ++(2 *\orthogonalityTestWidth, 0.) node [right] {$2 \nu_{\text{ortho}}$};

    \end{tikzpicture}
  \end{minipage}
  \vspace{\baselineskip}
  \caption{\textbf{Conceptual sketch for gradient tests.} \textit{(a)} Relevant
    vectors to formulate the geometric constraints between population and
    mini-batch gradient probed by the gradient tests. \textit{(b)} Gradient test
    visualization in \cockpit.}
  \label{fig:gradient-tests-sketch}
\end{figure}

The above works propose enforcing other weaker similarity in expectation during
optimization. These geometric constraints reduce to basic vector geometry (see
\Cref{fig:gradient-tests-sketch} (a) for an overview of the relevant vectors).
We recall their formulation here for consistency and derive the practical
versions, which can be computed from training observables and are used in
\cockpit (consult \Cref{fig:gradient-tests-sketch} (b) for the visualization).

\paragraph{Usage:} All three gradient tests describe the noise level of
the gradients.
\citet{Bollapragada2017} and \citet{Byrd2012} adapt the batch size so that
the proposed geometric constraints are fulfilled.
Practitioners can use the combined gradient test plot, \ie top center plot in
\Cref{fig:showcase}, to monitor gradient noise during training
and adjust hyperparameters such as the batch size.

\subsubsection{Norm test (\texttt{NormTest})}
\label{app:norm-test}
The norm test \citep{Byrd2012} constrains the residual norm $\lVert
\vg_{\gB}(\vtheta) - \vg_{P}(\vtheta) \rVert$, rescaled by $\lVert
\vg_{P}(\vtheta) \rVert$. This gives rise to a
standardized ball of radius $\theta_{\text{norm}} \in (0, \infty)$ around the
population gradient, where the mini-batch gradient should reside.
\citet{Byrd2012} set $\theta_{\text{norm}} = 0.9$ in their
experiments and increase the batch size if (in the practical version, see below) the following constraint is not fulfilled
\begin{subequations}
  \begin{align}
    \label{eq:norm-test-constraint}
    \E\left[ \frac{ \left\lVert \vg_{\gB}(\vtheta) - \vg_{P}(\vtheta)
    \right\rVert^2 }{\left\lVert \vg_{P}(\vtheta) \right\rVert^2} \right]
    \le
    \theta_{\text{norm}}^2\,.
  \end{align}
  Instead of taking the expectation over mini-batches, \citet{Byrd2012} note that
  the above will be satisfied if
  \begin{equation}
    \label{eq:norm-test-individual}
    \frac{1}{|\gB|} \E\left[ \frac{ \left\lVert \vg_n(\vtheta) - \vg_{P}(\vtheta)
        \right\rVert^2 }{\left\lVert \vg_{P}(\vtheta) \right\rVert^2} \right]
    \le \theta_{\text{norm}}^2\,.
  \end{equation}
\end{subequations}
They propose a practical form of this test,
\begin{subequations}
  \begin{equation}
    \label{eq:norm-test-practical-proposed}
    \frac{1}{|\gB| (|\gB| - 1)} \frac{\sum_{n \in \gB} \left\lVert \vg_n(\vtheta) -
        \vg_{\gB}(\vtheta) \right\rVert^2}{\left\lVert
        \vg_{\gB}(\vtheta)\right\rVert^2} \le \theta_{\text{norm}}^2\,,
  \end{equation}
  which can be computed from mini-batch statistics. Rearranging
  \begin{align}
    \sum_{n \in \gB} \left\lVert \vg_n(\vtheta) - \vg_{\gB}(\vtheta) \right\rVert^2
      &= \left( \sum_{n \in \gB} \left\lVert \vg_n(\vtheta) \right\rVert^2 \right) - |\gB|
        \left\lVert \vg_{\gB}(\vtheta) \right\rVert^2\,,
  \end{align}
  we arrive at
  \begin{align}
    \label{eq:norm-test-feature-table}
    \frac{1}{|\gB| (|\gB| - 1)} \left[ \frac{ \sum_{n \in \gB} \left\lVert
    \vg_n(\vtheta) \right\rVert^2 }{\left\lVert
    \vg_{\gB}(\vtheta)\right\rVert^2} - |\gB| \right] &\le
                                                        \theta_{\text{norm}}^2
  \end{align}
\end{subequations}
that leverages the norm of both the mini-batch and the individual gradients,
which can be aggregated over parameters during a backward pass.
\cockpit's \texttt{NormTest} corresponds to the maximum radius
$\theta_{\text{norm}}$ for which the above inequality holds.

\subsubsection{Inner product test (\texttt{InnerTest})}
\label{app:inner-product-test}
The inner product test \citep{Bollapragada2017} constrains the projection of
$\vg_{\gB}(\vtheta)$ onto $\vg_{P}(\vtheta)$ (compare
\Cref{fig:gradient-tests-sketch} (a)),
\begin{align}
  \label{eq:inner-product-projection}
  \mathrm{proj}_{\vg_{P}(\vtheta)}\left(\vg_{\gB}(\vtheta)\right)
  =
  \frac{\vg_{\gB}(\vtheta)^\top \vg_{P}(\vtheta)}{\left\lVert
  \vg_{P}(\vtheta) \right\rVert^2} \vg_{P}(\vtheta)\,,
\end{align}
rescaled by $\lVert \vg_{P}(\vtheta) \rVert$. This restricts the mini-batch
gradient to reside in a standardized band of relative width
$\theta_{\text{inner}}\in (0, \infty)$ around the population risk gradient.
\citet{Bollapragada2017} use $\theta_{\text{inner}} = 0.9$ (in the practical
version, see below)
to adapt the batch size if the parallel component's variance does not satisfy
the condition
\begin{subequations}
  \begin{align}
    \label{eq:inner-product-test}
    \Var\left( \frac{\vg_{\gB}(\vtheta)^\top \vg_{P}(\vtheta)}{\left\lVert
    \vg_{P}(\vtheta) \right\rVert^2} \right)
    &= \E\left[ \left( \frac{\vg_{\gB}(\vtheta)^\top \vg_{P}(\vtheta)}{\left\lVert
      \vg_{P}(\vtheta) \right\rVert^2}  -1 \right)^2 \right]
      \le \theta_{\text{inner}}^2
  \end{align}
  (note that by \Cref{eq:iid-sampling-expectation} we have
  $\E\left[ \frac{\vg_{\gB}(\vtheta)^\top \vg_{P}(\vtheta)}{\left\lVert
        \vg_{P}(\vtheta) \right\rVert^2} \right] = 1 $).
  \citet{Bollapragada2017} bound \Cref{eq:inner-product-test} by the individual
  gradient variance,
  \begin{align}
    \label{eq:inner-product-test-individual}
    \frac{1}{|\gB|}\Var\left( \frac{\vg_n(\vtheta)^\top \vg_{P}(\vtheta)}{\left\lVert
    \vg_{P}(\vtheta) \right\rVert^2}\right)
    =
    \frac{1}{|\gB|} \E \left[ \left( \frac{\vg_n(\vtheta)^\top \vg_{P}(\vtheta) }{\left\lVert
    \vg_{P}(\vtheta) \right\rVert^2} - 1    \right)^2  \right] \le \theta_{\text{inner}}^2\,.
  \end{align}
\end{subequations}
They then propose a practical form of \Cref{eq:inner-product-test-individual},
which uses the mini-batch sample variance, %
\begin{subequations}
  \begin{align}
    \label{eq:inner-product-test-practical}
    \frac{1}{|\gB|} \Var\left( \frac{\vg_n(\vtheta)^\top \vg_{\gB}(\vtheta)}{\left\lVert
    \vg_{\gB}(\vtheta)\right\rVert^2}\right)
    = \frac{1}{|\gB| (|\gB| - 1)}\left[  \sum_{n\in \gB}  \left( \frac{\vg_n(\vtheta)^\top \vg_{\gB}(\vtheta)}{\left\lVert
    \vg_{\gB}(\vtheta)\right\rVert^2} - 1    \right)^2  \right]
    \le \theta_{\text{inner}}^2\,.
  \end{align}
  Expanding
  \begin{align}
    \label{eq:inner-product-test-practical-rewrite}
    \sum_{n\in \gB}  \left( \frac{\vg_n(\vtheta)^\top \vg_{\gB}(\vtheta)}{\left\lVert
    \vg_{\gB}(\vtheta)\right\rVert^2} - 1    \right)^2
    &=
      \frac{\sum_{n\in \gB}  \left( \vg_n(\vtheta)^\top \vg_{\gB}(\vtheta)\right)^2}{\left\lVert
      \vg_{\gB}(\vtheta)\right\rVert^4} - |\gB|
  \end{align}
  and inserting \Cref{eq:inner-product-test-practical-rewrite} into
  \Cref{eq:inner-product-test-practical} yields
  \begin{align}
    \label{eq:inner-product-test-feature-table}
    \frac{1}{|\gB| (|\gB| - 1)}
    \left[   \frac{\sum_{n\in \gB}  \left( \vg_n(\vtheta)^\top \vg_{\gB}(\vtheta)\right)^2}{\left\lVert
    \vg_{\gB}(\vtheta)\right\rVert^4} - |\gB|\right]
        &\le \theta_{\text{inner}}^2\,.
  \end{align}
\end{subequations}
It relies on pairwise scalar products between individual gradients, which can be
aggregated over layers during backpropagation. \cockpit's \texttt{InnerTest}
quantity computes the maximum band width $\theta_{\text{inner}}$ that satisfies
\Cref{eq:inner-product-test-feature-table}.

\subsubsection{Orthogonality test (\texttt{OrthoTest})}
\label{app:orthogonality-test}
In contrast to the inner product test (\Cref{app:inner-product-test}) which
constrains the projection (\Cref{eq:inner-product-projection}), the
orthogonality test \citep{Bollapragada2017} constrains the orthogonal part (see
\Cref{fig:gradient-tests-sketch} (a))
\begin{align}
  \label{eq:orthogonality-projection}
  \vg_{\gB}(\vtheta)
  -
  \mathrm{proj}_{\vg_{P}(\vtheta)}\left(\vg_{\gB}(\vtheta)\right)\,,
\end{align}
rescaled by $\lVert \vg_{P}(\vtheta) \rVert$. This restricts the mini-batch
gradient to a standardized band of relative width $\nu_{\text{ortho}} \in (0,
\infty)$ parallel to the population gradient. \citet{Bollapragada2017} use $\nu =
\tan(80^{\circ}) \approx 5.84$ (in the practical version, see below) to adapt
the batch size if the following condition is violated,
\begin{subequations}
  \begin{align}
    \label{eq:orthogonality-test-constraint}
    \E\left[
    \left\lVert
    \frac{
    \vg_{\gB}(\vtheta)
    -
    \mathrm{proj}_{\vg_{P}(\vtheta)}\left(\vg_{\gB}(\vtheta)\right)
    }{
    \lVert \vg_{P}(\vtheta) \rVert
    }
    \right\rVert^2
    \right]
    \le \nu^2_{\text{ortho}}\,.
  \end{align}
  Expanding the norm, and inserting \Cref{eq:inner-product-projection}, this simplifies to
  \begin{align}
    \begin{split}
      \E
      \left[
        \left\lVert
          \frac{
            \vg_{\gB}(\vtheta)
          }{
            \lVert
            \vg_{P}(\vtheta)
            \rVert
          }
          -
          \frac{
            \vg_{\gB}(\vtheta)^\top
            \vg_{P}(\vtheta)
          }{
            \lVert
            \vg_{P}(\vtheta)
            \rVert^2
          }
          \frac{
            \vg_{P}(\vtheta)
          }{
            \lVert
            \vg_{P}(\vtheta)
            \rVert}
        \right\rVert^2
      \right]
      &\le \nu^2_{\text{ortho}}\,,
      \\
      \E\left[
        \frac{
          \lVert
          \vg_{\gB}(\vtheta)
          \rVert^2
        }{
          \lVert
          \vg_{P}(\vtheta)
          \rVert^2
        }
        -
        \frac{
          \left(
            \vg_{\gB}(\vtheta)^\top
            \vg_{P}(\vtheta)
          \right)^2
        }{
          \lVert
          \vg_{P}(\vtheta)
          \rVert^4
        }
      \right]
      &\le \nu^2_{\text{ortho}}\,.
    \end{split}
  \end{align}
  \citet{Bollapragada2017} bound this inequality using individual gradients instead,
  \begin{align}
    \frac{1}{|\gB|}
    \E \left[
    \left\lVert
    \frac{
    \vg_n(\vtheta)
    }{
    \lVert
    \vg_{P}(\vtheta)
    \rVert^2
    }
    -
    \frac{
    \vg_n(\vtheta)^\top
    \vg_{P}(\vtheta)
    }{
    \lVert
    \vg_{P}(\vtheta)
    \rVert^2
    }
    \frac{
    \vg_{P}(\vtheta)
    }{
    \left\lVert
    \vg_{P}(\vtheta)
    \right\rVert}
    \right\rVert^2
    \right]
    &\le \nu^2_{\text{ortho}}\,.
  \end{align}
\end{subequations}
They propose the practical form
\begin{subequations}
  \begin{align}
    \frac{1}{|\gB|(|\gB|-1)}
    \E\left[
    \left\lVert
    \frac{
    \vg_n(\vtheta)
    }{
    \lVert
    \vg_{\gB}(\vtheta)
    \rVert
    }
    -
    \frac{
    \vg_n(\vtheta)^\top
    \vg_{\gB}(\vtheta)
    }{
    \lVert
    \vg_{\gB}(\vtheta)
    \rVert^2
    }
    \frac{
    \vg_{\gB}(\vtheta)
    }{
    \left\lVert
    \vg_{\gB}(\vtheta)
    \right\rVert}
    \right\rVert^2
    \right]
    &\le \nu^2_{\text{ortho}}\,,
  \end{align}
  which simplifies to
  \begin{align}
    \label{eq:orthogonality-test-feature-table}
    \frac{1}{|\gB| (|\gB| - 1)}
    \sum_{n \in \gB }
    \left(
    \frac{
    \lVert
    \vg_n(\vtheta)
    \rVert^2
    }{
    \lVert
    \vg_{\gB}(\vtheta)
    \rVert^2
    }
    -
    \frac{
    \left(
    \vg_n(\vtheta)^\top
    \vg_{\gB}(\vtheta)
    \right)^2
    }{
    \lVert
    \vg_{\gB}(\vtheta)
    \rVert^4
    }
    \right)
    &\le \nu^2_{\text{ortho}}\,.
  \end{align}
\end{subequations}
It relies on pairwise scalar products between individual gradients which can be
aggregated over layers during a backward pass.\cockpit's \texttt{OrthTest}
quantity computes the maximum band width $\nu_{\text{ortho}}$ which satisfies
\Cref{eq:orthogonality-test-feature-table}.

\paragraph{Relation to acute angle test:} Recently, a novel ``acute angle test''
was proposed by \citet{Bahamou2019}. While the theoretical constraint between
$\vg_{\gB}(\vtheta)$ and $\vg_{P}(\vtheta)$ differs from the orthogonality test,
the practical versions coincide. Hence, we do not incorporate the acute angle here.

\subsection{Hessian maximum eigenvalue (\texttt{HessMaxEV})}
\label{app:max-ev}

The Hessian's maximum eigenvalue $\lambda_{\text{max}}(\mH_{\gB}(\vtheta))$ is
computed with an iterative eigensolver from Hessian-vector products through
\pytorch's automatic differentiation \citep{Pearlmutter1994}.
Like \citet{Yao2020}, we employ power iterations with similar
\href{https://github.com/amirgholami/PyHessian/blob/0f7e0f63a0f132998608013351ba19955fc9d861/pyhessian/hessian.py#L111-L158}{default
  stopping parameters} (stop after at most 100 iterations, or if the iterate
does converged with a relative and absolute tolerance of $10^{-3}, 10^{-6}$,
respectively) to compute $\lambda_{\text{max}}(\mH_{\gB}(\vtheta))$ through the
\texttt{HessMaxEV} quantity in \cockpit.

In principle, more sophisticated eigensolvers (for example Arnoldi's method)
could be applied to converge in fewer iterations or compute eigenvalues other
than the leading ones. \citet{Warsa2004} empirically demonstrate that the FLOP
ratio between power iteration and implicitly restarted Arnoldi method can reach
values larger than $100$. While we can use such a beneficial method on a CPU through
\href{https://docs.scipy.org/doc/scipy/reference/generated/scipy.sparse.linalg.eigsh.html}
{\texttt{scipy.sparse.linalg.eigsh}} we are restricted to the GPU-compatible
power iteration for GPU training. We expect that extending the support of
popular machine learning libraries like \pytorch for such iterative eigensolvers
on GPUs can help to save computation time.

\begin{equation}
  \label{eq:hess-max-ev-feature-table}
  \lambda_{\text{max}}(\mH_{\gB}(\vtheta)) = \max_{\lVert \vv \rVert = 1} \lVert \mH_{\gB}(\vtheta)\vv \rVert
  =
  \max_{\vv \in \sR^D} \frac{\vv^\top \mH_{\gB}(\vtheta) \vv}{\vv^\top \vv}.
\end{equation}

\paragraph{Usage:} The Hessian's maximum eigenvalue describes the
loss surface's sharpest direction and thus provides an understanding of
the current loss landscape.
Additionally, in convex optimization, the largest Hessian eigenvalue
crucially determines the appropriate step size \citep{Schmidt2014}.
In \Cref{sec:showcase}, we can observe that although training seems
stuck in the very first few iterations progress is visible when looking
at the maximum Hessian eigenvalue.

\subsection{Hessian trace (\texttt{HessTrace})}
\label{app:hessian-trace}

In comparison to \citet{Yao2020}, who leverage Hessian-vector products
\citep{Pearlmutter1994} to estimate the Hessian trace, we compute
the exact value $\Tr(\mH_{\gB}(\vtheta))$ with the \texttt{HessTrace} quantity in
\cockpit by aggregating the output of \backpack's
\href{https://docs.backpack.pt/en/master/extensions.html#backpack.extensions.DiagHessian}{\texttt{DiagHessian}}
extension, which computes the diagonal entries of $\mH_{\gB}(\vtheta)$.
Alternatively, the trace can also be estimated from the generalized Gauss-Newton
matrix, or an MC-sampled approximation thereof.

\paragraph{Usage:} The Hessian trace equals the sum of the eigenvalues and
thus provides a notion of ``average curvature'' of the current loss landscape. It
has long been theorized and discussed that curvature and generalization
performance may be linked \citep[\citeeg]{Hochreiter1997}.

\subsection{Takeuchi Information Criterion (TIC) (\texttt{TICDiag},
  \texttt{TICTrace})}
\label{app:tic}

Recent work by \citet{Thomas2020} suggests that optimizer convergence speed and
generalization is mainly influenced by curvature and gradient noise; and hence their
interaction is crucial to understand the generalization and optimization
behavior of deep neural networks. They reinvestigate the Takeuchi Information criterion
\citep{Takeuchi1976}, an estimator for the generalization gap in
overparameterized maximum likelihood estimation. At a local minimum
$\vtheta^\star$, the generalization gap is estimated by the TIC
\begin{equation}
  \label{eq:tic-theory}
  \frac{1}{|\gD|}
  \Tr
  \left(
    \mH_P(\vtheta^\star)^{-1}
    \mC_P(\vtheta^\star)
    \right)\,,
\end{equation}
where $\mH_P(\vtheta^\star)$ is the population Hessian and
$\mC_P(\vtheta^\star)$ is the gradient's uncentered second moment,
\begin{equation*}
  \mC_P(\vtheta^\star)
  = \int
    \nabla_\vtheta\ell(f(\vtheta^\star, \vx), \vy)
    \left(
    \nabla_\vtheta\ell(f(\vtheta^\star, \vx), \vy)
  \right)^\top P(\vx, \vy)
  \:d\vx\:d\vy.
\end{equation*}
Both matrices are inaccessible in practice. In their experiments,
\citet{Thomas2020} propose the approximation $\Tr(\mC) / \Tr(\mH)$ for
$\Tr(\mH^{-1} \mC)$. They also replace the Hessian by the Fisher as it is easier
to compute. With these practical simplifications, they investigate the TIC of
trained neural networks where the curvature and noise matrix are evaluated on a
large data set.

The TIC provided in \cockpit differs from this setting, since by design we want
to observe quantities during training, while avoiding additional model
predictions. Also, \backpack provides access to the Hessian; hence we don't need
to use the Fisher. We propose the following two approximations of the TIC from a
mini-batch:
\begin{itemize}
\item \texttt{TICTrace}: Uses the approximation of \citet{Thomas2020} which
  replaces the matrix-product trace by the product of traces,
  \begin{align}
  	\label{eq:tic-trace-feature-table}
    \frac{
      \Tr\left(
        \mC_{\gB}(\vtheta)
      \right)
    }{
      \Tr\left(
        \mH_{\gB}(\vtheta)
      \right)
    }
    =
    \frac{
      \frac{1}{|\gB|}
      \sum_{n\in\gB}
      \lVert
        \vg_{n}(\vtheta)
      \rVert^2
    }{
      \Tr\left(
        \mH_{\gB}(\vtheta)
      \right)
    }\,.
  \end{align}
\item \texttt{TICDiag}: Uses a diagonal approximation of the Hessian, which
  is cheap to invert,
  \begin{align}
  	\label{eq:tic-diag-feature-table}
    \Tr\left(
      \mathrm{diag}\left(
        \mH_{\gB}(\vtheta)
      \right)^{-1}
      \mC_{\gB}(\vtheta)
    \right)
    =
    \frac{1}{|\gB|}
    \sum_{j=1}^D
    \left[
      \mH_\gB(\vtheta)
      \right]_{j,j}^{-1}
      \left[
        \sum_{n\in\gB}
        \vg_{n}(\vtheta)^{\odot 2}
      \right]_{j}\,.
  \end{align}
\end{itemize}

\paragraph{Usage:} The TIC is a proxy for the model's generalization gap, see
\citet{Thomas2020}.

\subsection{Gradient signal-to-noise-ratio (\texttt{MeanGSNR})}
\label{app:mean-gsnr}
The gradient signal-to-noise-ratio $\mathrm{GSNR}(\left[ \vtheta \right]_j) \in \sR$ for
a single parameter $\left[ \vtheta \right]_j$ is defined as
\begin{align}
  \label{eq:gsnr-definition}
  \mathrm{GSNR}(\left[ \vtheta \right]_j)
  =
  \frac{
  \E_{(\vx, \vy)\sim P}\left[
  \left[
  \nabla_\vtheta\ell(f(\vtheta, \vx), \vy)
  \right]_j
  \right]^2
  }{
  \Var_{(\vx, \vy)\sim P}\left[
  \left[
  \nabla_\vtheta \ell(f(\vtheta, \vx), \vy)
  \right]_j
  \right]
  }
  =
  \frac{
  \left[
  \vg_{P}(\vtheta)
  \right]^2_j
  }{
  \left[
  \mSigma_P(\vtheta)
  \right]_{j,j}
  }\,.
\end{align}
\begin{subequations}
\citet{Liu2020} use it to explain generalization properties of models
in the early training phase. We apply their estimation to mini-batches,
  \begin{align}
    \label{eq:gsnr-mini-batch}
    \mathrm{GSNR}(\left[ \vtheta \right]_j)
    &\approx
      \frac{
      \left[
      \vg_{\gB}(\vtheta)
      \right]_j^2
      }{
      \frac{|\gB| - 1}{|\gB|}
      \left[
      \hat{\mSigma}_{\gB}(\vtheta)
      \right]_{j,j}
      }
      =
      \frac{
      \left[
      \vg_{\gB}(\vtheta)
      \right]_j^2
      }{
      \frac{1}{|\gB|}
      \left(
      \sum_{n\in \gB}
      \left[
      \vg_n(\vtheta)
      \right]_j^2
      \right)
      -
      \left[
      \vg_{\gB}(\vtheta)
      \right]_j^2
      }\,.
  \end{align}
  Inspired by \citet{Liu2020}, \cockpit's \texttt{MeanGSNR} computes the
  average GSNR over all parameters,
  \begin{equation}
    \label{eq:mean-gsnr-feature-table}
    \frac{1}{D}\sum_{j=1}^D \mathrm{GSNR}(\left[ \vtheta \right]_j)\,.
  \end{equation}
\end{subequations}

\paragraph{Usage:} The \gsnr describes the gradient noise level which is
influenced, among other things, by the batch size. Using the \gsnr, perhaps in
combination with the gradient tests or the \cabs criterion could provide
practitioners a clearer picture of suitable batch sizes for their particular
problem. As shown by \citet{Liu2020}, the \gsnr is also linked to generalization
of neural networks.

\section{Additional experiments}
\label{app:experiments}

In this section, we present additional experiments and use cases that showcase
\cockpit's utility. \Cref{app:misscaled_data_exp_imagenet} shows that \cockpit
is able to scale to larger data sets by running the experiment with incorrectly
scaled data (see \Cref{sec:misscaled_data_exp}) on \imagenet instead of
\cifarten. \Cref{app:implicit_regularization_exp} provides another concrete use
case similar to \Cref{fig:LINE}: detecting regularization during training.

\subsection{Incorrectly scaled data for \textsc{ImageNet}}
\label{app:misscaled_data_exp_imagenet}

We repeat the experiment of \Cref{sec:misscaled_data_exp} on the \imagenet
\citep{Deng2009} data set instead of \cifarten. We also use a larger neural
network model, switching from \threecthreed to \vgg \citep{Simonyan2015}. This
demonstrates that \cockpit is able to scale to both larger models and data sets.
The input size of the images is almost fifty times larger ($224 \times 224$
instead of $32 \times 32$). The model size increased by roughly a factor of 150
(\vgg for \imagenet has roughly 138 million parameters, \threecthreed has less
than a million).

\begin{figure}[h]
	\pgfkeys{/pgfplots/preprocessingexperimentdefault/.style={
			width=\linewidth,
			height=1.4\linewidth,
			every axis plot/.append style={line width = 1.2pt},
			tick pos = left,
			xmajorticks = true,
			ymajorticks = true,
			ylabel near ticks,
			xlabel near ticks,
			xtick align = inside,
			ytick align = inside,
			legend cell align = left,
			legend columns = 1,
			legend pos = south east,
			legend style = {
				fill opacity = 0.9,
				text opacity = 1,
				font = \small,
			},
			xticklabel style = {font = \small, inner xsep = -5ex},
			xlabel style = {font = \small},
			axis line style = {black},
			yticklabel style = {font = \small, inner ysep = -4ex},
			ylabel style = {font = \small},
			title style = {font = \small, inner ysep = -3ex},
			grid = major,
			grid style = {dashed}
		}
	}

	\centering
	\begin{subfigure}[t]{0.46\textwidth}
		\begin{minipage}{.49\textwidth}
			\Cshadowbox{\includegraphics[width = .35\textwidth]{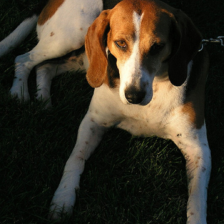}}
			\Cshadowbox{\includegraphics[width = .35\textwidth]{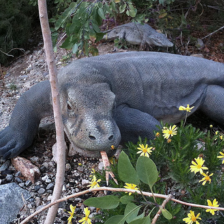}}

			\Cshadowbox{\includegraphics[width = .35\textwidth]{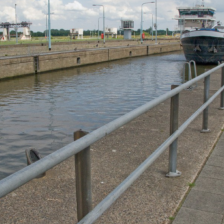}}
			\Cshadowbox{\includegraphics[width = .35\textwidth]{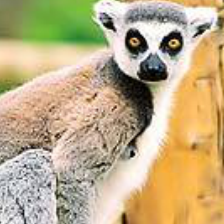}}
		\end{minipage}
		\begin{minipage}{.49\textwidth}
			\centering
			\pgfkeys{/pgfplots/zmystyle/.style={preprocessingexperimentdefault,
					ylabel={Gradient Element}
			}}
			\vspace{1.4\baselineskip}
\begin{tikzpicture}

\definecolor{color0}{rgb}{0.12156862745098,0.466666666666667,0.705882352941177}

\begin{axis}[
axis line style={white},
log basis x={10},
tick align=outside,
xmajorticks=false,
xmin=0.9, xmax=315917612.324796,
xmode=log,
xtick style={color=white!15!black},
ymajorticks=false,
ymin=-1.5, ymax=1.5,
zmystyle
]
\draw[draw=white,fill=color0,line width=0.04pt] (axis cs:0.9,-1.5) rectangle (axis cs:0.9,-1.42499995231628);
\draw[draw=white,fill=color0,line width=0.04pt] (axis cs:0.9,-1.42500007152557) rectangle (axis cs:0.9,-1.35000002384186);
\draw[draw=white,fill=color0,line width=0.04pt] (axis cs:0.9,-1.35000002384186) rectangle (axis cs:0.9,-1.27499997615814);
\draw[draw=white,fill=color0,line width=0.04pt] (axis cs:0.9,-1.27499997615814) rectangle (axis cs:0.9,-1.19999992847443);
\draw[draw=white,fill=color0,line width=0.04pt] (axis cs:0.9,-1.20000004768372) rectangle (axis cs:0.9,-1.125);
\draw[draw=white,fill=color0,line width=0.04pt] (axis cs:0.9,-1.125) rectangle (axis cs:0.9,-1.04999995231628);
\draw[draw=white,fill=color0,line width=0.04pt] (axis cs:0.9,-1.04999995231628) rectangle (axis cs:0.9,-0.974999904632568);
\draw[draw=white,fill=color0,line width=0.04pt] (axis cs:0.9,-0.975000023841858) rectangle (axis cs:0.9,-0.899999976158142);
\draw[draw=white,fill=color0,line width=0.04pt] (axis cs:0.9,-0.899999976158142) rectangle (axis cs:0.9,-0.824999928474426);
\draw[draw=white,fill=color0,line width=0.04pt] (axis cs:0.9,-0.825000047683716) rectangle (axis cs:0.9,-0.75);
\draw[draw=white,fill=color0,line width=0.04pt] (axis cs:0.9,-0.75) rectangle (axis cs:0.9,-0.674999952316284);
\draw[draw=white,fill=color0,line width=0.04pt] (axis cs:0.9,-0.674999952316284) rectangle (axis cs:0.9,-0.599999904632568);
\draw[draw=white,fill=color0,line width=0.04pt] (axis cs:0.9,-0.600000023841858) rectangle (axis cs:0.9,-0.524999976158142);
\draw[draw=white,fill=color0,line width=0.04pt] (axis cs:0.9,-0.524999976158142) rectangle (axis cs:0.9,-0.449999928474426);
\draw[draw=white,fill=color0,line width=0.04pt] (axis cs:0.9,-0.449999988079071) rectangle (axis cs:0.9,-0.374999940395355);
\draw[draw=white,fill=color0,line width=0.04pt] (axis cs:0.9,-0.374999970197678) rectangle (axis cs:0.9,-0.299999922513962);
\draw[draw=white,fill=color0,line width=0.04pt] (axis cs:0.9,-0.299999982118607) rectangle (axis cs:0.9,-0.224999934434891);
\draw[draw=white,fill=color0,line width=0.04pt] (axis cs:0.9,-0.224999964237213) rectangle (axis cs:3.9,-0.149999916553497);
\draw[draw=white,fill=color0,line width=0.04pt] (axis cs:0.9,-0.149999968707561) rectangle (axis cs:102.9,-0.0749999210238457);
\draw[draw=white,fill=color0,line width=0.04pt] (axis cs:0.9,-0.075000025331974) rectangle (axis cs:123780803.9,2.23517417907715e-08);
\draw[draw=white,fill=color0,line width=0.04pt] (axis cs:0.9,-8.19563865661621e-08) rectangle (axis cs:14580628.9,0.0749999657273293);
\draw[draw=white,fill=color0,line width=0.04pt] (axis cs:0.9,0.0749999210238457) rectangle (axis cs:98.9,0.149999968707561);
\draw[draw=white,fill=color0,line width=0.04pt] (axis cs:0.9,0.149999916553497) rectangle (axis cs:6.9,0.224999964237213);
\draw[draw=white,fill=color0,line width=0.04pt] (axis cs:0.9,0.224999934434891) rectangle (axis cs:0.9,0.299999982118607);
\draw[draw=white,fill=color0,line width=0.04pt] (axis cs:0.9,0.299999922513962) rectangle (axis cs:0.9,0.374999970197678);
\draw[draw=white,fill=color0,line width=0.04pt] (axis cs:0.9,0.374999940395355) rectangle (axis cs:0.9,0.449999988079071);
\draw[draw=white,fill=color0,line width=0.04pt] (axis cs:0.9,0.449999928474426) rectangle (axis cs:0.9,0.524999976158142);
\draw[draw=white,fill=color0,line width=0.04pt] (axis cs:0.9,0.524999976158142) rectangle (axis cs:0.9,0.600000023841858);
\draw[draw=white,fill=color0,line width=0.04pt] (axis cs:0.9,0.599999904632568) rectangle (axis cs:0.9,0.674999952316284);
\draw[draw=white,fill=color0,line width=0.04pt] (axis cs:0.9,0.674999952316284) rectangle (axis cs:0.9,0.75);
\draw[draw=white,fill=color0,line width=0.04pt] (axis cs:0.9,0.75) rectangle (axis cs:0.9,0.825000047683716);
\draw[draw=white,fill=color0,line width=0.04pt] (axis cs:0.9,0.824999928474426) rectangle (axis cs:0.9,0.899999976158142);
\draw[draw=white,fill=color0,line width=0.04pt] (axis cs:0.9,0.899999976158142) rectangle (axis cs:0.9,0.975000023841858);
\draw[draw=white,fill=color0,line width=0.04pt] (axis cs:0.9,0.974999904632568) rectangle (axis cs:1.9,1.04999995231628);
\draw[draw=white,fill=color0,line width=0.04pt] (axis cs:0.9,1.04999995231628) rectangle (axis cs:0.9,1.125);
\draw[draw=white,fill=color0,line width=0.04pt] (axis cs:0.9,1.125) rectangle (axis cs:0.9,1.20000004768372);
\draw[draw=white,fill=color0,line width=0.04pt] (axis cs:0.9,1.19999992847443) rectangle (axis cs:0.9,1.27499997615814);
\draw[draw=white,fill=color0,line width=0.04pt] (axis cs:0.9,1.27499997615814) rectangle (axis cs:0.9,1.35000002384186);
\draw[draw=white,fill=color0,line width=0.04pt] (axis cs:0.9,1.35000002384186) rectangle (axis cs:0.9,1.42500007152557);
\draw[draw=white,fill=color0,line width=0.04pt] (axis cs:0.9,1.42499995231628) rectangle (axis cs:0.9,1.5);
\end{axis}

\end{tikzpicture}
 		\end{minipage}
		\caption{Normalized Data}
		\label{fig:data-pre-processing_norm_imagenet}
	\end{subfigure}
	\hfill
	\begin{subfigure}[t]{0.46\textwidth}
		\begin{minipage}{.49\textwidth}
			\Cshadowbox{\includegraphics[width = .35\textwidth]{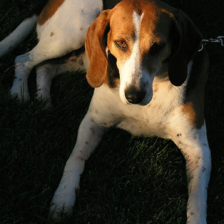}}
			\Cshadowbox{\includegraphics[width = .35\textwidth]{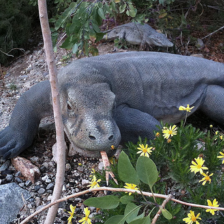}}

			\Cshadowbox{\includegraphics[width = .35\textwidth]{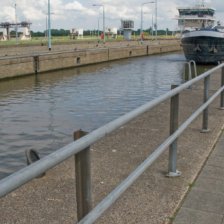}}
			\Cshadowbox{\includegraphics[width = .35\textwidth]{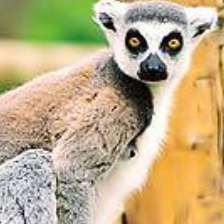}}
		\end{minipage}
		\begin{minipage}{.49\textwidth}
			\centering
			\pgfkeys{/pgfplots/zmystyle/.style={preprocessingexperimentdefault,
					ylabel={Gradient Element}
			}}
			\vspace{1.4\baselineskip}
\begin{tikzpicture}

\definecolor{color0}{rgb}{1,0.498039215686275,0.0549019607843137}

\begin{axis}[
axis line style={white},
log basis x={10},
tick align=outside,
xmajorticks=false,
xmin=0.9, xmax=315354991.80789,
xmode=log,
xtick style={color=white!15!black},
ymajorticks=false,
ymin=-1.5, ymax=1.5,
zmystyle
]
\draw[draw=white,fill=color0,line width=0.04pt] (axis cs:0.9,-1.5) rectangle (axis cs:0.9,-1.42499995231628);
\draw[draw=white,fill=color0,line width=0.04pt] (axis cs:0.9,-1.42500007152557) rectangle (axis cs:0.9,-1.35000002384186);
\draw[draw=white,fill=color0,line width=0.04pt] (axis cs:0.9,-1.35000002384186) rectangle (axis cs:0.9,-1.27499997615814);
\draw[draw=white,fill=color0,line width=0.04pt] (axis cs:0.9,-1.27499997615814) rectangle (axis cs:0.9,-1.19999992847443);
\draw[draw=white,fill=color0,line width=0.04pt] (axis cs:0.9,-1.20000004768372) rectangle (axis cs:0.9,-1.125);
\draw[draw=white,fill=color0,line width=0.04pt] (axis cs:0.9,-1.125) rectangle (axis cs:0.9,-1.04999995231628);
\draw[draw=white,fill=color0,line width=0.04pt] (axis cs:0.9,-1.04999995231628) rectangle (axis cs:0.9,-0.974999904632568);
\draw[draw=white,fill=color0,line width=0.04pt] (axis cs:0.9,-0.975000023841858) rectangle (axis cs:0.9,-0.899999976158142);
\draw[draw=white,fill=color0,line width=0.04pt] (axis cs:0.9,-0.899999976158142) rectangle (axis cs:0.9,-0.824999928474426);
\draw[draw=white,fill=color0,line width=0.04pt] (axis cs:0.9,-0.825000047683716) rectangle (axis cs:0.9,-0.75);
\draw[draw=white,fill=color0,line width=0.04pt] (axis cs:0.9,-0.75) rectangle (axis cs:0.9,-0.674999952316284);
\draw[draw=white,fill=color0,line width=0.04pt] (axis cs:0.9,-0.674999952316284) rectangle (axis cs:0.9,-0.599999904632568);
\draw[draw=white,fill=color0,line width=0.04pt] (axis cs:0.9,-0.600000023841858) rectangle (axis cs:0.9,-0.524999976158142);
\draw[draw=white,fill=color0,line width=0.04pt] (axis cs:0.9,-0.524999976158142) rectangle (axis cs:0.9,-0.449999928474426);
\draw[draw=white,fill=color0,line width=0.04pt] (axis cs:0.9,-0.449999988079071) rectangle (axis cs:0.9,-0.374999940395355);
\draw[draw=white,fill=color0,line width=0.04pt] (axis cs:0.9,-0.374999970197678) rectangle (axis cs:0.9,-0.299999922513962);
\draw[draw=white,fill=color0,line width=0.04pt] (axis cs:0.9,-0.299999982118607) rectangle (axis cs:40.9,-0.224999934434891);
\draw[draw=white,fill=color0,line width=0.04pt] (axis cs:0.9,-0.224999964237213) rectangle (axis cs:872.9,-0.149999916553497);
\draw[draw=white,fill=color0,line width=0.04pt] (axis cs:0.9,-0.149999968707561) rectangle (axis cs:45198.9,-0.0749999210238457);
\draw[draw=white,fill=color0,line width=0.04pt] (axis cs:0.9,-0.075000025331974) rectangle (axis cs:123570849.9,2.23517417907715e-08);
\draw[draw=white,fill=color0,line width=0.04pt] (axis cs:0.9,-8.19563865661621e-08) rectangle (axis cs:14693460.9,0.0749999657273293);
\draw[draw=white,fill=color0,line width=0.04pt] (axis cs:0.9,0.0749999210238457) rectangle (axis cs:48872.9,0.149999968707561);
\draw[draw=white,fill=color0,line width=0.04pt] (axis cs:0.9,0.149999916553497) rectangle (axis cs:1580.9,0.224999964237213);
\draw[draw=white,fill=color0,line width=0.04pt] (axis cs:0.9,0.224999934434891) rectangle (axis cs:137.9,0.299999982118607);
\draw[draw=white,fill=color0,line width=0.04pt] (axis cs:0.9,0.299999922513962) rectangle (axis cs:91.9,0.374999970197678);
\draw[draw=white,fill=color0,line width=0.04pt] (axis cs:0.9,0.374999940395355) rectangle (axis cs:74.9,0.449999988079071);
\draw[draw=white,fill=color0,line width=0.04pt] (axis cs:0.9,0.449999928474426) rectangle (axis cs:73.9,0.524999976158142);
\draw[draw=white,fill=color0,line width=0.04pt] (axis cs:0.9,0.524999976158142) rectangle (axis cs:62.9,0.600000023841858);
\draw[draw=white,fill=color0,line width=0.04pt] (axis cs:0.9,0.599999904632568) rectangle (axis cs:44.9,0.674999952316284);
\draw[draw=white,fill=color0,line width=0.04pt] (axis cs:0.9,0.674999952316284) rectangle (axis cs:58.9,0.75);
\draw[draw=white,fill=color0,line width=0.04pt] (axis cs:0.9,0.75) rectangle (axis cs:54.9,0.825000047683716);
\draw[draw=white,fill=color0,line width=0.04pt] (axis cs:0.9,0.824999928474426) rectangle (axis cs:45.9,0.899999976158142);
\draw[draw=white,fill=color0,line width=0.04pt] (axis cs:0.9,0.899999976158142) rectangle (axis cs:27.9,0.975000023841858);
\draw[draw=white,fill=color0,line width=0.04pt] (axis cs:0.9,0.974999904632568) rectangle (axis cs:21.9,1.04999995231628);
\draw[draw=white,fill=color0,line width=0.04pt] (axis cs:0.9,1.04999995231628) rectangle (axis cs:23.9,1.125);
\draw[draw=white,fill=color0,line width=0.04pt] (axis cs:0.9,1.125) rectangle (axis cs:12.9,1.20000004768372);
\draw[draw=white,fill=color0,line width=0.04pt] (axis cs:0.9,1.19999992847443) rectangle (axis cs:11.9,1.27499997615814);
\draw[draw=white,fill=color0,line width=0.04pt] (axis cs:0.9,1.27499997615814) rectangle (axis cs:12.9,1.35000002384186);
\draw[draw=white,fill=color0,line width=0.04pt] (axis cs:0.9,1.35000002384186) rectangle (axis cs:6.9,1.42500007152557);
\draw[draw=white,fill=color0,line width=0.04pt] (axis cs:0.9,1.42499995231628) rectangle (axis cs:20.9,1.5);
\end{axis}

\end{tikzpicture}
 		\end{minipage}
		\caption{Raw Data}
		\label{fig:data-pre-processing_raw_imagenet}
	\end{subfigure}
	\caption{\textbf{Same inputs, different gradients on \textsc{ImageNet}.} This
    is structurally the same plot as \Cref{fig:data-pre-processing}, but using
    \imagenet and \vgg. (a) \emph{normalized} ($[0, 1]$) and (b) \emph{raw}
    $([0, 255])$ images look identical in auto-scaled front-ends like
    \matplotlib's \texttt{imshow}. The gradient distribution on the \vgg model,
    however, is affected by this scaling.}
	\label{fig:data-pre-processing_imagenet}
\end{figure}

Similar to the example shown in the main text, the gradients are affected by the
scaling introduced via the input images, albeit less drastically (see
\Cref{fig:data-pre-processing_imagenet}). Due to the gradient scaling, default
optimization hyperparameters might not work well anymore for the model using the
raw input data.

\subsection{Detecting implicit regularization of the optimizer}
\label{app:implicit_regularization_exp}

In non-convex optimization, optimizers can converge to local minima with
different properties. Here, we illustrate this by investigating the effect of
sub-sampling noise on a simple task from \cite{Mulayoff2020, Ginsburg2020}.

We generate synthetic data $\gD = \{(x_n, y_n) \in \sR \times \sR
\}_{n=1}^{N=100}$ for a regression task with $x \sim \gN(0; 1)$ with noisy
observations $y = 1.4 x + \epsilon$ where $\epsilon \sim \gN(0;1)$. The model is
a scalar network with parameters $\vtheta = \begin{pmatrix} w_1 &
  w_2 \end{pmatrix}^\top \in \sR^2$, initialized at $\vtheta_0 = \begin{pmatrix}
  0.1 & 1.7 \end{pmatrix}^\top$, that produces predictions via $f(\vtheta, x) =
w_2 w_1 x$. We seek to minimize the mean squared error
\begin{equation*}
  \gL_\gD(\vtheta) = \frac{1}{N} \sum_{n=1}^{N} \left( f(\theta, x_n) - y_n \right)^2
\end{equation*}
and compare \sgd ($|\gB|=95$) with \gd ($|\gB|= N =100$) at a learning rate of
$0.1$ (see \Cref{fig:implicit-regularization}).

We observe that the loss of both \sgd and \gd is almost identical. Using a noisy
gradient regularizes the Hessian's maximum eigenvalue though. It decreases in
later stages where the loss curve suggests that training has converged. This
regularization effect constitutes an important phenomenon that cannot be
observed by monitoring only the loss.

\pgfkeys{/pgfplots/regularizationdefault/.style={
    width=1.0\linewidth,
    height=0.8\linewidth,
    every axis plot/.append style={line width = 1.2pt},
    every axis background/.style={fill=white},
    ymajorticks=true,
    xmajorticks=true,
    tick pos = left,
    ylabel near ticks,
    xlabel near ticks,
    xtick align = inside,
    ytick align = inside,
    legend cell align = left,
    legend columns = 1,
    legend pos = north east,
    legend style = {
      fill opacity = 0.7,
      text opacity = 1,
      font = \small,
    },
    xticklabel style = {font = \small, inner xsep = -5ex},
    xlabel style = {font = \small},
    axis line style = {black},
    yticklabel style = {font = \small, inner ysep = -4ex},
    ylabel style = {font = \small},
    grid = major,
    grid style = {dashed}
  }
}

\begin{figure}
  \centering
	\begin{subfigure}[t]{0.495\textwidth}
		\pgfkeys{/pgfplots/zmystyle/.style={regularizationdefault, ymin=0.6, ymax=1.1}}
\begin{tikzpicture}

\definecolor{color0}{rgb}{0.12156862745098,0.466666666666667,0.705882352941177}
\definecolor{color1}{rgb}{1,0.498039215686275,0.0549019607843137}

\begin{axis}[
axis line style={white},
legend style={fill opacity=0.8, draw opacity=1, text opacity=1, draw=white!80!black},
log basis x={10},
tick align=outside,
xlabel={Iteration},
xmajorticks=false,
xmin=0.563970595937194, xmax=167345.603972783,
xmode=log,
xtick style={color=white!15!black},
ylabel={Mini-Batch Loss},
ymajorticks=false,
ymin=0.614918631315231, ymax=1.72726913094521,
zmystyle
]
\addplot [, color0]
table {%
0 1.65519058704376
1 1.01682019233704
2 0.780193150043488
3 0.775982201099396
4 0.737731337547302
5 0.771629452705383
6 0.738641858100891
7 0.745266854763031
8 0.772922933101654
9 0.748066663742065
10 0.77786260843277
11 0.769368886947632
12 0.775993466377258
13 0.769854426383972
14 0.780099034309387
15 0.732372939586639
16 0.76972508430481
17 0.77314555644989
18 0.787652790546417
19 0.731954038143158
20 0.753742933273315
21 0.760406374931335
22 0.772047460079193
24 0.752689599990845
25 0.768256425857544
27 0.772502541542053
28 0.745130240917206
30 0.73753297328949
32 0.775065541267395
34 0.7575803399086
36 0.748904526233673
38 0.78914475440979
40 0.774977564811707
42 0.770744025707245
45 0.771463930606842
48 0.724553644657135
51 0.750621318817139
54 0.786653280258179
57 0.740645587444305
60 0.780794203281403
64 0.765990436077118
68 0.772936582565308
72 0.764917433261871
76 0.75068473815918
81 0.763052582740784
86 0.766409575939178
91 0.779118478298187
96 0.779287576675415
102 0.694868266582489
108 0.753131628036499
114 0.745236694812775
121 0.769624173641205
128 0.733922481536865
136 0.763625741004944
144 0.771767735481262
153 0.778103351593018
162 0.76782751083374
172 0.770959913730621
182 0.778098523616791
193 0.747863173484802
204 0.767393231391907
217 0.747031450271606
230 0.761284053325653
243 0.788995981216431
258 0.78480863571167
273 0.725325763225555
289 0.788911044597626
307 0.757820010185242
325 0.701482534408569
344 0.786840677261353
365 0.75093948841095
387 0.793320834636688
410 0.733399331569672
434 0.790173828601837
460 0.789791882038116
488 0.772913932800293
517 0.747723639011383
547 0.767966389656067
580 0.769767463207245
615 0.777678191661835
651 0.732484042644501
690 0.765304982662201
731 0.779313802719116
775 0.752284646034241
821 0.751649916172028
870 0.763818562030792
922 0.763033270835876
977 0.76042252779007
1035 0.794110774993896
1096 0.770817458629608
1162 0.719465613365173
1231 0.773622274398804
1304 0.777041494846344
1382 0.752636194229126
1464 0.746034681797028
1552 0.756577551364899
1644 0.710349500179291
1742 0.788471639156342
1846 0.749858379364014
1956 0.751055300235748
2072 0.77653980255127
2196 0.743179202079773
2327 0.777839779853821
2465 0.765402734279633
2612 0.784416019916534
2768 0.675363481044769
2933 0.767980337142944
3107 0.77206939458847
3292 0.761081337928772
3489 0.780567407608032
3696 0.759546518325806
3917 0.772010743618011
4150 0.731160163879395
4397 0.778384268283844
4659 0.792823255062103
4937 0.780224800109863
5231 0.716187834739685
5542 0.732295513153076
5872 0.757798075675964
6222 0.783280313014984
6593 0.774591028690338
6985 0.76332038640976
7401 0.73084568977356
7842 0.744604051113129
8309 0.77698802947998
8804 0.770139813423157
9329 0.7605100274086
9884 0.745865881443024
10473 0.772988736629486
11097 0.784264206886292
11758 0.73901504278183
12458 0.665480017662048
13200 0.74626225233078
13987 0.765305399894714
14820 0.788454294204712
15702 0.783531785011292
16638 0.790920555591583
17629 0.723659753799438
18679 0.79127299785614
19791 0.776141047477722
20970 0.772833287715912
22219 0.73369562625885
23542 0.771220922470093
24945 0.785363912582397
26430 0.72164660692215
28005 0.788875877857208
29673 0.772274553775787
31440 0.764169812202454
33312 0.723604738712311
35297 0.758412897586823
37399 0.769568562507629
39626 0.782877564430237
41987 0.774317562580109
44487 0.74730920791626
47137 0.759216368198395
49945 0.75877833366394
52919 0.77678370475769
56071 0.770803034305573
59411 0.757794678211212
62949 0.775205314159393
66699 0.787786245346069
70671 0.761527419090271
74881 0.740764439105988
79340 0.766364872455597
84066 0.771571159362793
89073 0.774646580219269
94378 0.766631364822388
};
\addlegendentry{SGD}
\addplot [, color1]
table {%
0 1.67670774459839
1 1.00288474559784
2 0.814482688903809
3 0.769742786884308
4 0.761073589324951
5 0.759601593017578
6 0.759367525577545
7 0.759331345558167
8 0.759325861930847
9 0.759325087070465
10 0.759324848651886
11 0.759324848651886
12 0.759324908256531
13 0.759324848651886
14 0.759324848651886
15 0.759324848651886
16 0.759324848651886
17 0.759324848651886
18 0.759324908256531
19 0.759324848651886
20 0.759324848651886
21 0.759324848651886
22 0.759324848651886
24 0.759324789047241
25 0.759324848651886
27 0.759324848651886
28 0.759324848651886
30 0.759324848651886
32 0.759324848651886
34 0.759324848651886
36 0.759324908256531
38 0.759324848651886
40 0.759324848651886
42 0.759324848651886
45 0.759324908256531
48 0.759324848651886
51 0.759324848651886
54 0.759324848651886
57 0.759324848651886
60 0.759324848651886
64 0.759324848651886
68 0.759324848651886
72 0.759324848651886
76 0.759324848651886
81 0.759324908256531
86 0.759324789047241
91 0.759324848651886
96 0.759324908256531
102 0.759324848651886
108 0.759324848651886
114 0.759324908256531
121 0.759324848651886
128 0.759324848651886
136 0.759324789047241
144 0.759324908256531
153 0.759324848651886
162 0.759324848651886
172 0.759324908256531
182 0.759324908256531
193 0.759324848651886
204 0.759324908256531
217 0.759324908256531
230 0.759324908256531
243 0.759324848651886
258 0.759324848651886
273 0.759324908256531
289 0.759324848651886
307 0.759324848651886
325 0.759324848651886
344 0.759324789047241
365 0.759324848651886
387 0.759324848651886
410 0.759324908256531
434 0.759324848651886
460 0.759324848651886
488 0.759324848651886
517 0.759324848651886
547 0.759324848651886
580 0.759324908256531
615 0.759324908256531
651 0.759324848651886
690 0.759324848651886
731 0.759324848651886
775 0.759324789047241
821 0.759324848651886
870 0.759324848651886
922 0.759324848651886
977 0.759324848651886
1035 0.759324908256531
1096 0.759324908256531
1162 0.759324789047241
1231 0.759324848651886
1304 0.759324848651886
1382 0.759324848651886
1464 0.759324848651886
1552 0.759324848651886
1644 0.759324848651886
1742 0.759324908256531
1846 0.759324848651886
1956 0.759324908256531
2072 0.759324908256531
2196 0.759324908256531
2327 0.759324789047241
2465 0.759324908256531
2612 0.759324848651886
2768 0.759324848651886
2933 0.759324848651886
3107 0.759324848651886
3292 0.759324908256531
3489 0.759324908256531
3696 0.759324908256531
3917 0.759324848651886
4150 0.759324848651886
4397 0.759324848651886
4659 0.759324848651886
4937 0.759324848651886
5231 0.759324848651886
5542 0.759324848651886
5872 0.759324908256531
6222 0.759324908256531
6593 0.759324848651886
6985 0.759324908256531
7401 0.759324908256531
7842 0.759324848651886
8309 0.759324848651886
8804 0.759324908256531
9329 0.759324848651886
9884 0.759324908256531
10473 0.759324848651886
11097 0.759324908256531
11758 0.759324848651886
12458 0.759324848651886
13200 0.759324848651886
13987 0.759324908256531
14820 0.759324908256531
15702 0.759324848651886
16638 0.759324848651886
17629 0.759324848651886
18679 0.759324908256531
19791 0.759324908256531
20970 0.759324848651886
22219 0.759324848651886
23542 0.759324848651886
24945 0.759324848651886
26430 0.759324908256531
28005 0.759324848651886
29673 0.759324848651886
31440 0.759324789047241
33312 0.759324848651886
35297 0.759324848651886
37399 0.759324789047241
39626 0.759324908256531
41987 0.759324848651886
44487 0.759324848651886
47137 0.759324848651886
49945 0.759324848651886
52919 0.759324848651886
56071 0.759324848651886
59411 0.759324908256531
62949 0.759324789047241
66699 0.759324848651886
70671 0.759324848651886
74881 0.759324848651886
79340 0.759324848651886
84066 0.759324848651886
89073 0.759324848651886
94378 0.759324908256531
};
\addlegendentry{GD}
\end{axis}

\end{tikzpicture}
 	\end{subfigure}
	\hfill
	\begin{subfigure}[t]{0.495\textwidth}
		\pgfkeys{/pgfplots/zmystyle/.style={regularizationdefault,
        legend style = {
          opacity = 0,
          fill opacity = 0,
          text opacity = 0,
          font = \small,
        },
        ylabel = {Max.\,Hessian eigenvalue}
      }}
\begin{tikzpicture}

\definecolor{color0}{rgb}{0.12156862745098,0.466666666666667,0.705882352941177}
\definecolor{color1}{rgb}{1,0.498039215686275,0.0549019607843137}

\begin{axis}[
axis line style={white},
legend style={fill opacity=0.8, draw opacity=1, text opacity=1, draw=white!80!black},
log basis x={10},
tick align=outside,
xlabel={Iteration},
xmajorticks=false,
xmin=0.563970595937194, xmax=167345.603972783,
xmode=log,
xtick style={color=white!15!black},
ylabel={Maximum Hessian eigenvalue},
ymajorticks=false,
ymin=3.85510742664337, ymax=6.68519914150238,
zmystyle
]
\addplot [, color0]
table {%
0 5.01858520507812
1 4.91631031036377
2 5.16143560409546
3 5.60115337371826
4 5.79127502441406
5 6.0561056137085
6 6.26061916351318
7 6.19245529174805
8 6.1329493522644
9 6.05357074737549
10 5.97355842590332
11 6.05519437789917
12 6.31369209289551
13 6.22404289245605
14 6.33904695510864
15 5.68460321426392
16 6.16733026504517
17 6.17319679260254
18 6.14776849746704
19 6.2695107460022
20 6.03493785858154
21 6.15226316452026
22 6.19907712936401
24 5.59337615966797
25 6.36163806915283
27 6.35177993774414
28 6.22747230529785
30 5.91255331039429
32 6.19828081130981
34 6.39521312713623
36 6.0371265411377
38 6.43193244934082
40 6.14503479003906
42 6.25523281097412
45 6.05292510986328
48 5.67981719970703
51 6.36472225189209
54 6.06869220733643
57 6.18863964080811
60 5.93778562545776
64 6.12580823898315
68 6.18363857269287
72 6.01109552383423
76 6.02523136138916
81 6.21799850463867
86 6.18191480636597
91 6.55655860900879
96 6.31975746154785
102 6.05267763137817
108 6.0778636932373
114 6.01467657089233
121 6.24698925018311
128 6.04640293121338
136 6.24112749099731
144 6.31688690185547
153 6.28657817840576
162 6.13792037963867
172 5.65208387374878
182 6.30600261688232
193 5.98742771148682
204 5.71659326553345
217 5.84029388427734
230 6.0068564414978
243 6.3081259727478
258 6.39043521881104
273 6.14346885681152
289 6.18961191177368
307 6.13053369522095
325 6.17688846588135
344 5.90940189361572
365 5.89127063751221
387 5.98564434051514
410 6.25674438476562
434 5.93166160583496
460 6.01510143280029
488 6.29282379150391
517 6.3133373260498
547 5.89875364303589
580 5.78323602676392
615 6.22842979431152
651 6.12653160095215
690 6.13379669189453
731 6.08839702606201
775 6.19680023193359
821 6.05588006973267
870 6.21298170089722
922 6.0586724281311
977 5.70560264587402
1035 6.05281639099121
1096 6.12884950637817
1162 5.72979116439819
1231 5.75020885467529
1304 5.78489637374878
1382 6.24677515029907
1464 5.72618103027344
1552 6.16928005218506
1644 5.99916076660156
1742 6.31236791610718
1846 5.42526912689209
1956 6.01460218429565
2072 6.17887306213379
2196 5.93650770187378
2327 6.21088695526123
2465 5.92963600158691
2612 5.98927402496338
2768 6.04936790466309
2933 5.9457950592041
3107 6.15447568893433
3292 6.04442024230957
3489 5.88113212585449
3696 6.0181360244751
3917 5.82596063613892
4150 5.97462749481201
4397 5.84421348571777
4659 5.84129619598389
4937 5.80539751052856
5231 6.00247669219971
5542 5.8760838508606
5872 5.92244338989258
6222 5.69848346710205
6593 5.64358997344971
6985 5.6120719909668
7401 5.73087453842163
7842 5.67744302749634
8309 5.43394136428833
8804 5.7706823348999
9329 5.68305492401123
9884 5.22680282592773
10473 5.09199523925781
11097 5.7391529083252
11758 5.49506330490112
12458 5.29003858566284
13200 5.21850109100342
13987 5.15451717376709
14820 5.25626420974731
15702 5.2664647102356
16638 5.09290647506714
17629 5.06191682815552
18679 5.22023582458496
19791 4.99110126495361
20970 5.13607454299927
22219 4.82361221313477
23542 4.96726989746094
24945 4.517502784729
26430 4.89909362792969
28005 4.69721984863281
29673 4.44070720672607
31440 4.71606779098511
33312 4.43321752548218
35297 4.52913951873779
37399 4.35525035858154
39626 4.48494052886963
41987 4.43455600738525
44487 4.4298300743103
47137 4.41794347763062
49945 4.16174602508545
52919 4.41815376281738
56071 4.30182790756226
59411 4.18179321289062
62949 4.38689374923706
66699 4.22925329208374
70671 4.26896953582764
74881 4.2410717010498
79340 4.02750778198242
84066 3.98374795913696
89073 4.10416793823242
94378 4.25959587097168
};
\addlegendentry{SGD}
\addplot [, color1]
table {%
0 5.19412231445312
1 4.89158582687378
2 5.36561346054077
3 5.75625705718994
4 5.95861196517944
5 6.04718065261841
6 6.08332633972168
7 6.09766292572021
8 6.10328483581543
9 6.10547971725464
10 6.10633707046509
11 6.10666942596436
12 6.10679912567139
13 6.106849193573
14 6.10686922073364
15 6.10687732696533
16 6.10687875747681
17 6.1068811416626
18 6.1068811416626
19 6.10688161849976
20 6.10688161849976
21 6.10688161849976
22 6.10688161849976
24 6.10688161849976
25 6.1068811416626
27 6.10688209533691
28 6.1068811416626
30 6.10688161849976
32 6.10688161849976
34 6.10688161849976
36 6.10688161849976
38 6.10688161849976
40 6.1068811416626
42 6.10688161849976
45 6.10688161849976
48 6.10688257217407
51 6.10688209533691
54 6.10688161849976
57 6.10688209533691
60 6.10688161849976
64 6.10688209533691
68 6.1068811416626
72 6.10688161849976
76 6.10688161849976
81 6.10688257217407
86 6.10688161849976
91 6.1068811416626
96 6.10688209533691
102 6.10688257217407
108 6.10688161849976
114 6.10688209533691
121 6.1068811416626
128 6.10688257217407
136 6.10688161849976
144 6.10688209533691
153 6.10688161849976
162 6.10688209533691
172 6.10688161849976
182 6.10688161849976
193 6.1068811416626
204 6.10688161849976
217 6.10688257217407
230 6.1068811416626
243 6.10688209533691
258 6.1068811416626
273 6.1068811416626
289 6.10688161849976
307 6.10688161849976
325 6.1068811416626
344 6.10688161849976
365 6.10688161849976
387 6.10688161849976
410 6.10688209533691
434 6.10688161849976
460 6.10688161849976
488 6.1068811416626
517 6.10688161849976
547 6.10688161849976
580 6.10688209533691
615 6.10688161849976
651 6.1068811416626
690 6.1068811416626
731 6.1068811416626
775 6.1068811416626
821 6.1068811416626
870 6.10688209533691
922 6.10688161849976
977 6.10688161849976
1035 6.10688161849976
1096 6.10688161849976
1162 6.10688161849976
1231 6.1068811416626
1304 6.10688209533691
1382 6.10688161849976
1464 6.10688161849976
1552 6.1068811416626
1644 6.1068811416626
1742 6.10688161849976
1846 6.10688209533691
1956 6.10688161849976
2072 6.1068811416626
2196 6.10688161849976
2327 6.1068811416626
2465 6.10688161849976
2612 6.10688161849976
2768 6.10688161849976
2933 6.10688209533691
3107 6.10688257217407
3292 6.10688161849976
3489 6.10688257217407
3696 6.10688161849976
3917 6.1068811416626
4150 6.10688209533691
4397 6.1068811416626
4659 6.10688161849976
4937 6.10688209533691
5231 6.10688161849976
5542 6.1068811416626
5872 6.10688161849976
6222 6.10688209533691
6593 6.1068811416626
6985 6.10688161849976
7401 6.10688209533691
7842 6.10688161849976
8309 6.10688209533691
8804 6.10688161849976
9329 6.10688161849976
9884 6.1068811416626
10473 6.1068811416626
11097 6.10688161849976
11758 6.10688161849976
12458 6.10688161849976
13200 6.1068811416626
13987 6.1068811416626
14820 6.1068811416626
15702 6.10688161849976
16638 6.10688161849976
17629 6.10688161849976
18679 6.10688161849976
19791 6.1068811416626
20970 6.1068811416626
22219 6.1068811416626
23542 6.10688209533691
24945 6.10688161849976
26430 6.10688209533691
28005 6.10688209533691
29673 6.1068811416626
31440 6.10688161849976
33312 6.10688209533691
35297 6.10688161849976
37399 6.10688161849976
39626 6.10688209533691
41987 6.10688161849976
44487 6.10688161849976
47137 6.10688161849976
49945 6.1068811416626
52919 6.10688209533691
56071 6.10688209533691
59411 6.10688161849976
62949 6.10688161849976
66699 6.10688161849976
70671 6.10688161849976
74881 6.10688161849976
79340 6.10688161849976
84066 6.10688161849976
89073 6.1068811416626
94378 6.10688161849976
};
\addlegendentry{GD}
\end{axis}

\end{tikzpicture}
 	\end{subfigure}
  \caption{\textbf{Observing implicit regularization of the optimizer with
      \cockpittitle} through a comparison of \sgd and \gd on a synthetic problem
    inspired by \cite{Mulayoff2020, Ginsburg2020} (details in the text).
    \textit{Left:} The mini-batch loss of both optimizers looks similar.
    \textit{Right:} Noise due to mini-batching regularizes the Hessian's maximum
    eigenvalue in stages where the loss suggests that training has converged.}
	  \label{fig:implicit-regularization}
\end{figure}
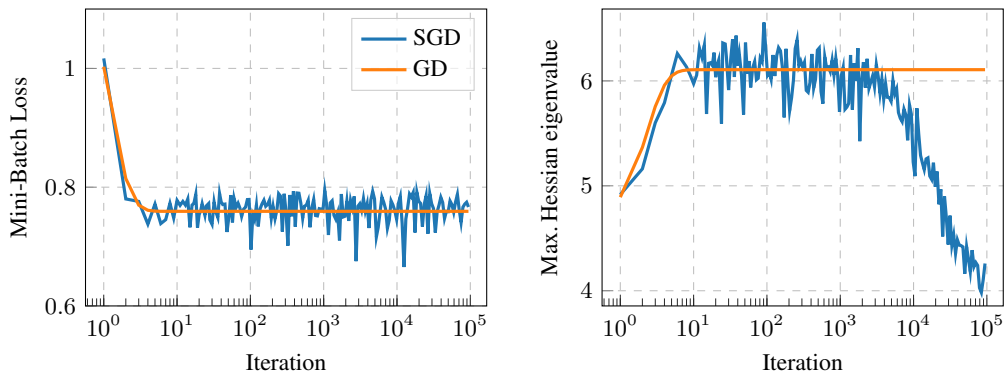

\clearpage

\section{Implementation details and additional benchmarks}
\label{app:benchmarks}

In this section, we provide more details about our implementation
(\Cref{app:hooks_benchmarks}) to access the desired quantities with as little
overhead as possible.
Additionally, we present more benchmarks for individual instruments
(\Cref{app:benchmark-instruments}) and \cockpit
configurations (\Cref{app:benchmark-configuration}).
These are similar but extended versions of the ones presented in
\Cref{fig:benchmark-instruments,fig:benchmark_heatmap} in the main
text.
Lastly, we benchmark different implementations of computing the two-dimensional
gradient histogram (\Cref{app:histograms}), identifying a computational
bottleneck for its current GPU implementation.

\paragraph{Hardware details:} Throughout this paper, we conducted benchmarks on
the following setup
\begin{itemize}
\item \textbf{CPU:} Intel Core i7-8700K CPU @ 3.70\,GHz × 12 (32\,GB)
\item \textbf{GPU:} NVIDIA GeForce RTX 2080 Ti (11\,GB)
\end{itemize}

\paragraph{Test problem details:} The experiments in this paper rely mostly on 
optimization problems provided by the \deepobs benchmark suite \citep{Schneider2019}.
If not stated otherwise, we use the default training details suggested by \deepobs,
that are summarized below. For more details see the original paper.
\begin{itemize}
	\item \textbf{Quadratic Deep:} A stochastic quadratic problem with an eigenspectrum 
	similar to what has been reported for neural nets. Default batch size $128$,
	default number of epochs $100$.
	\item \textbf{\mnist Log. Reg.:} Multinomial logistic regression on \mnist \citep{Lecun1998}.
	Default batch size $128$, default number of epochs $50$.
	\item \textbf{\mnist \mlp:} Multi-layer perceptron neural network on \mnist.
	Default batch size $128$, default number of epochs $100$.
	\item \textbf{\fmnist \mlp:} Multi-layer perceptron neural network on \fmnist \citep{Xiao2017}.
	Default batch size $128$, default number of epochs $100$.
	\item \textbf{\fmnist \twoctwod:} A two convolutional and two dense layered neural network on \fmnist.
	Default batch size $128$, default number of epochs $100$.
	\item \textbf{\cifarten \threecthreed:} A three convolutional and three dense layered neural network on \cifarten \citep{Krizhevsky2009}. Default batch size $128$, default number of epochs $100$.
	\item \textbf{\cifarhun \allcnnc:} All Convolutional Neural Network C (\allcnnc \citep{Springenberg2015}) on \cifarhun \citep{Krizhevsky2009}. Default batch size $256$, default number of epochs $350$.
	\item \textbf{\svhn \threecthreed:}	A three convolutional and three dense layered neural network on \svhn \citep{Netzer2011}. Default batch size $128$, default number of epochs $100$.
\end{itemize}

\subsection{Hooks \& Memory benchmarks}
\label{app:hooks_benchmarks}

To improve memory consumption, we compact information during the backward pass
by adding hooks to the neural network's layers. These are executed after
\backpack extensions and have access to the quantities computed therein. They
compress information to what is requested by a quantity and free the memory
occupied by \backpack buffers. Such savings primarily depend on the
parameter distribution over layers, and are bigger for more balanced
architectures (compare \Cref{fig:memory-benchmark}).

\paragraph{Example:} Say, we want to compute a histogram over the $|\gB| \times
D$ individual gradient elements of a network. Suppose that $|\gB| = 128$ and the
model is \deepobs's \cifarten \threecthreed test problem with $895,210$
parameters. Given that every parameter is stored in single precision, the
model requires $895,210 \times 4\,\text{Bytes} \approx 3.41\,\text{MB}$.
Storing the individual gradients will require $128 \times 895,210 \times
4\,\text{Bytes} \approx 437\,\text{MB}$ (for larger networks this quickly
exceeds the available memory as the individual gradients occupy $|\gB|$ times
the model size). If instead, the layer-wise individual gradients are condensed
into histograms of negligible size and immediately freed afterwards during
backpropagation, the maximum memory overhead reduces to storing the individual
gradients of the largest layer. For our example, the largest layer has
$589,824$ parameters, and the associated individual gradients will require $128
\times 589,824\times 4\,\text{Bytes} \approx 288\,\text{MB}$, saving roughly
$149\,\text{MB}$ of RAM. In practice, we observe these expected savings, see
\Cref{fig:memory-benchmark-cifar10}.

\pgfkeys{/pgfplots/memorybenchmarkdefault/.style={
    enlarge x limits=-0.05,
    width=1.0\linewidth,
    height=0.3\linewidth,
    every axis plot/.append style={line width = 1.2pt},
    ymin=0,
    tick pos = left,
    ylabel near ticks,
    xlabel near ticks,
    xtick align = inside,
    ytick align = inside,
    legend cell align = left,
    legend columns = 1,
    legend pos = south east,
    legend style = {
      fill opacity = 0.7,
      text opacity = 1,
      font = \small,
    },
    xticklabel style = {font = \small, inner xsep = -5ex},
    xlabel style = {font = \small},
    axis line style = {black},
    yticklabel style = {font = \small, inner ysep = -4ex},
    ylabel style = {font = \small},
    title style = {font = \small, inner ysep = -3ex},
    grid = major,
    grid style = {dashed}
  }
}

\captionsetup[subfigure]{justification=justified,singlelinecheck=false}

\begin{figure}
	\begin{subfigure}[t]{0.99\textwidth}
		\pgfkeys{/pgfplots/zmystyle/.style={ memorybenchmarkdefault, xlabel = {},
				legend pos = north west}}
		\caption{\qquad \fmnist \twoctwod}

 		\label{fig:memory-benchmark-fmnist}
	\end{subfigure}
	\hfill
	\begin{subfigure}[t]{0.99\textwidth}
		\pgfkeys{/pgfplots/zmystyle/.style={ memorybenchmarkdefault, xlabel = {},
				legend pos = north west}}
		\caption{\qquad \mnist \mlp}
%
 		\label{fig:memory-benchmark-mnist}
	\end{subfigure}
	\hfill
	\begin{subfigure}[t]{0.99\textwidth}
		\pgfkeys{/pgfplots/zmystyle/.style={ memorybenchmarkdefault, xlabel = {},
				legend pos = north west}}
		\caption{\qquad \cifarten \threecthreed}
%
 		\label{fig:memory-benchmark-cifar10}
	\end{subfigure}
	\hfill
	\begin{subfigure}[t]{0.99\textwidth}
		\pgfkeys{/pgfplots/zmystyle/.style={ memorybenchmarkdefault,
				legend pos = north west}}
		\caption{\qquad \cifarhun \allcnnc}
%
 		\label{fig:memory-benchmark-cifar100}
	\end{subfigure}
	  \caption{\textbf{Memory consumption and savings with hooks} during one
	    forward-backward step on a CPU for different \deepobs problems. We compare
	    three settings; i) without \cockpit (baseline); ii) \cockpit with
	    \texttt{GradHist1d} with \backpack (expensive); iii) \cockpit with
	    \texttt{GradHist1d} with \backpack and additional hooks (optimized). Peak
	    memory consumptions are highlighted by horizontal dashed bars and shown in
	    the legend. Shaded areas, if visible, fill two standard deviations above and
	    below the mean value, all of them result from ten independent runs. Dotted
	    lines indicate individual runs. Our optimized approach allows to free
	    obsolete tensors during backpropagation and thereby reduces memory
	    consumption. From top to bottom: the effect is less pronounced for
	    architectures that concentrate the majority of parameters in a single layer
	    ((\subref{fig:memory-benchmark-fmnist}) $3,274,634$ total, $3,211,264$ largest layer) and
	    increases for more balanced networks ((\subref{fig:memory-benchmark-mnist}) $1,336,610$ total,
	    $784,000$ largest layer, (\subref{fig:memory-benchmark-cifar10}): $895,210$
	    total, $589,824$ largest layer).}
	  \label{fig:memory-benchmark}
\end{figure}

\captionsetup[subfigure]{justification=centering, singlelinecheck=true}

 \clearpage
\subsection{Additional run time benchmarks}
\label{app:run-time-benchmarks}

\subsubsection{Individual instrument overhead}
\label{app:benchmark-instruments}

To estimate the computational overhead for individual instruments, we run
\cockpit with that instrument for $32$ iterations, tracking at every step.
Training proceeds with the default batch size specified by the \deepobs
problem and uses \sgd with learning rate $10^{-3}$. We measure the time
between iterations $1$ and $32$, and average for the overhead per step. Every
such estimate is repeated over $10$ random seeds to obtain mean and error bars
as reported in \Cref{fig:benchmark-instruments}.

Note that this protocol does \textit{not} include initial overhead for setting
up data loading and also does \textit{not} include the time for evaluating
train/test loss on a larger data set, which is usually done by practitioners.
Hence, we even expect the shown overheads to be smaller in a conventional
training loop which includes the above steps.

\begin{figure}[p]
	\vfill
	\begin{subfigure}[t]{0.99\textwidth}
		\caption{\qquad Computational overhead for \mnist Log.\,Reg.\,(GPU)}
		\includegraphics[width=\textwidth]{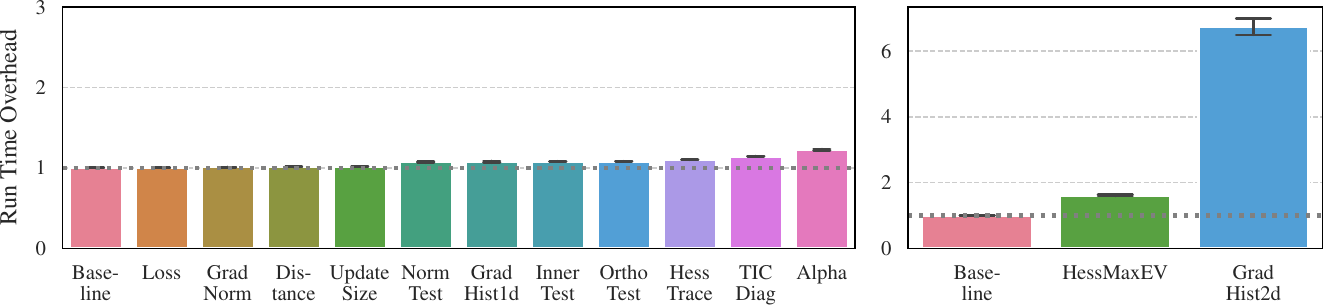}
		\label{fig:app_benchmark_instruments_cuda-mnist_logreg}
		\vspace{0.5cm}
	\end{subfigure}
	\vfill
	\begin{subfigure}[t]{0.99\textwidth}
		\caption{\qquad Computational overhead for \mnist \mlp\,(GPU)}
		\includegraphics[width=\textwidth]{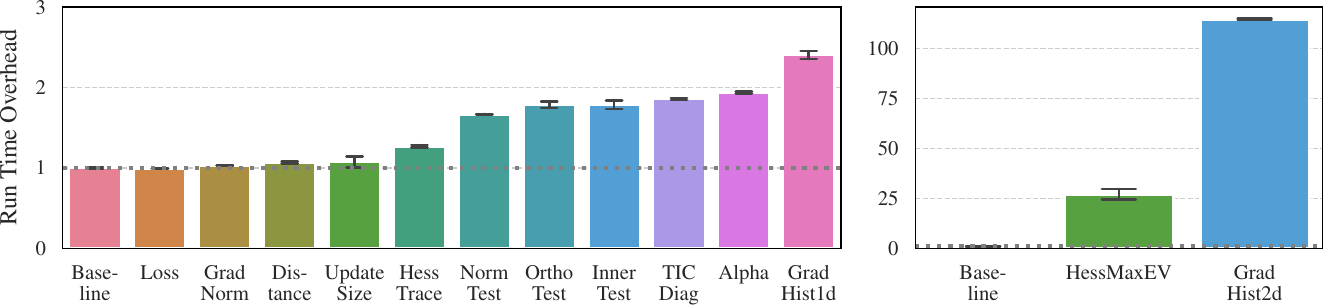}
		\label{fig:app_benchmark_instruments_cuda-mnist_mlp}
		\vspace{0.5cm}
	\end{subfigure}
	\vfill
	\begin{subfigure}[t]{0.99\textwidth}
		\caption{\qquad Computational overhead for \cifarten \threecthreed\,(GPU)}
		\includegraphics[width=\textwidth]{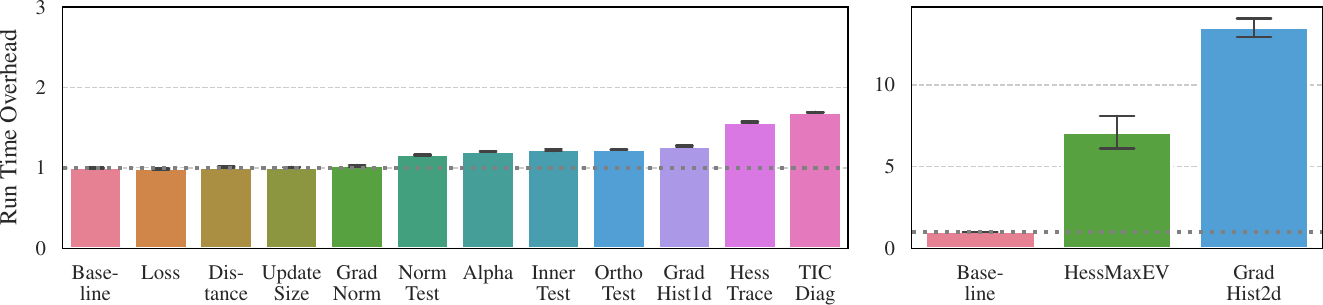}
		\label{fig:app_benchmark_instruments_cuda-cifar10}
		\vspace{0.5cm}
	\end{subfigure}
	\vfill
	\begin{subfigure}[t]{0.99\textwidth}
		\caption{\qquad Computational overhead for \fmnist \twoctwod\,(GPU)}
		\includegraphics[width=\textwidth]{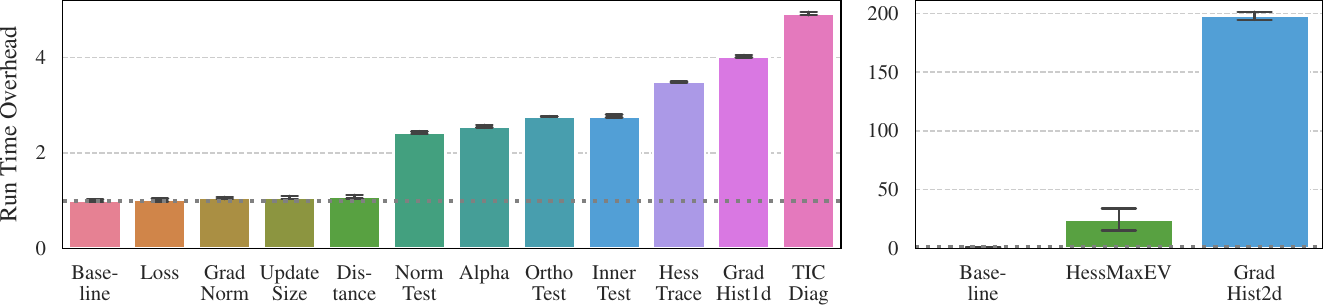}
		\label{fig:app_benchmark_instruments_cuda-fmnist}
		\vspace{0.5cm}
	\end{subfigure}
	\vfill
	\caption{\textbf{Individual overhead of \cockpittitle's instruments on GPU for
		four different problems.} All run times are shown as multiples of
	the \emph{baseline} without tracking. Expensive quantities are displayed in
	separate panels on the right. Experimental details in the text.}
	\label{fig:app_benchmark_instruments_cuda}
\end{figure}

\begin{figure}[p]
	\vfill
	\begin{subfigure}[t]{0.99\textwidth}
		\caption{\qquad Computational overhead for \mnist Log.\,Reg.\,(CPU)}
		\includegraphics[width=\textwidth]{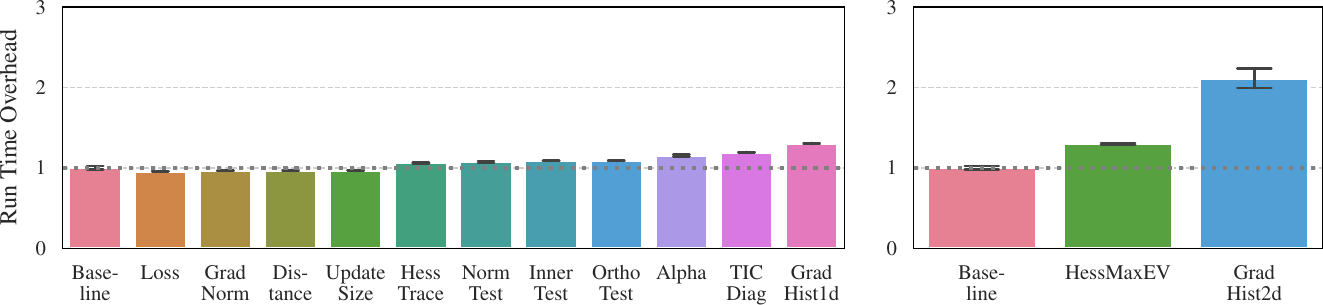}
		\label{fig:app_benchmark_instruments_cpu-mnist_logreg}
		\vspace{0.5cm}
	\end{subfigure}
	\vfill
	\begin{subfigure}[t]{0.99\textwidth}
		\caption{\qquad Computational overhead for \mnist \mlp\,(CPU)}
		\includegraphics[width=\textwidth]{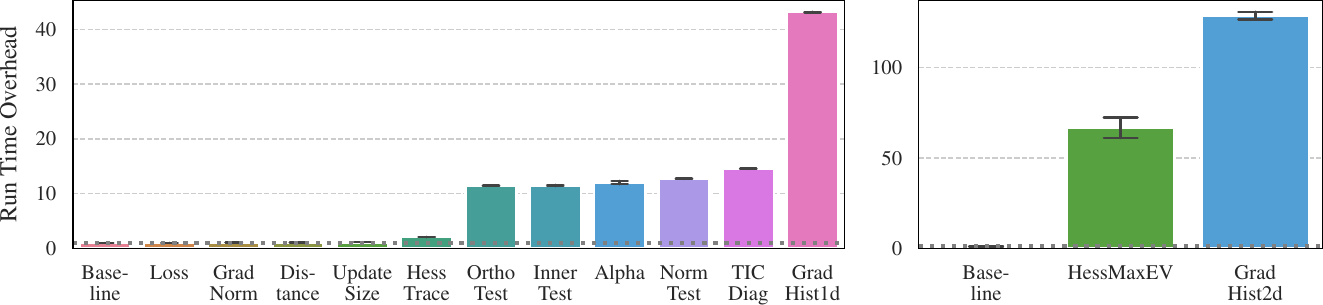}
		\label{fig:app_benchmark_instruments_cpu-mnist_mlp}
		\vspace{0.5cm}
	\end{subfigure}
	\vfill
	\begin{subfigure}[t]{0.99\textwidth}
		\caption{\qquad Computational overhead for \cifarten \threecthreed\,(CPU)}
		\includegraphics[width=\textwidth]{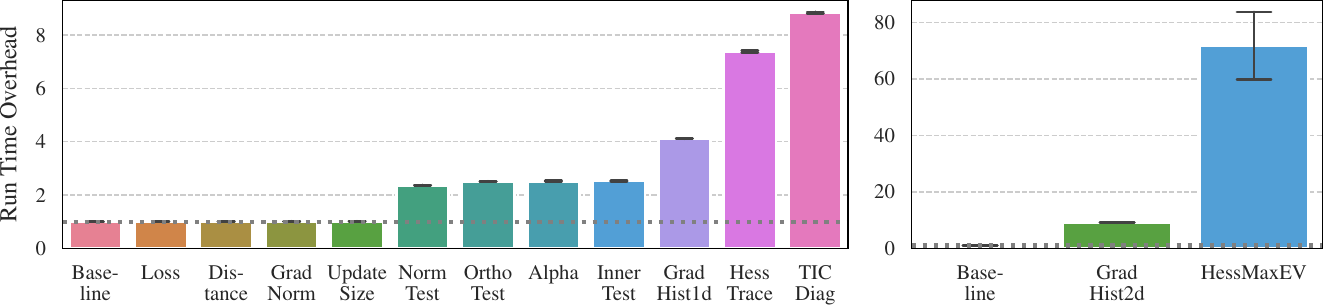}
		\label{fig:app_benchmark_instruments_cpu-cifar10}
		\vspace{0.5cm}
	\end{subfigure}
	\vfill
	\begin{subfigure}[t]{0.99\textwidth}
		\caption{\qquad Computational overhead for \fmnist \twoctwod\,(CPU)}
		\includegraphics[width=\textwidth]{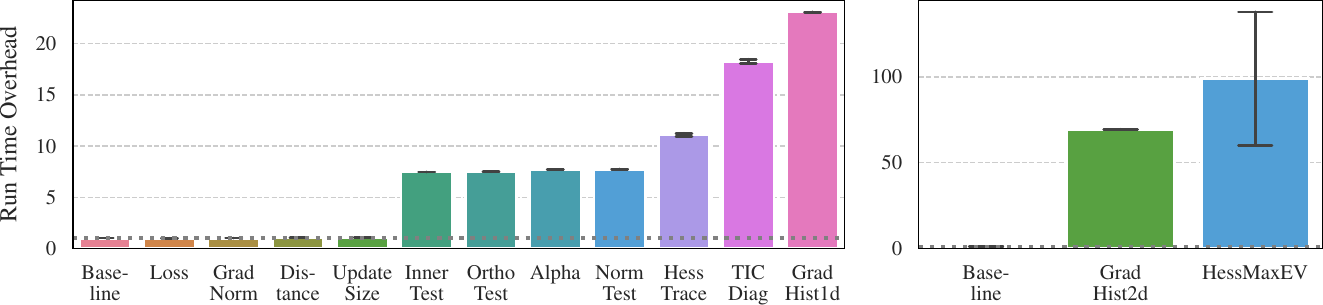}
		\label{fig:app_benchmark_instruments_cpu-fmnist}
		\vspace{0.5cm}
	\end{subfigure}
	\vfill
	\caption{\textbf{Individual overhead of \cockpittitle's instruments on CPU for
			four different problems.} All run times are shown as multiples of
		the \emph{baseline} without tracking. Expensive quantities are displayed in
		separate panels on the right. Experimental details in the text.}
	\label{fig:app_benchmark_instruments_cpu}
\end{figure}

\paragraph{Individual overhead on GPU versus CPU:}
\Cref{fig:app_benchmark_instruments_cuda} and
\Cref{fig:app_benchmark_instruments_cpu} show the individual overhead for four
different \deepobs problems on GPU and CPU, respectively. The left part of
\Cref{fig:app_benchmark_instruments_cuda} (c) corresponds to
\Cref{fig:benchmark-instruments}. Right panels show the expensive quantities,
which we omitted in the main text as they were expected to be expensive due to
their computational work (\texttt{HessMaxEV}) or bottlenecks in the
implementation (\texttt{GradHist2d}, see \Cref{app:histograms} for details). We
see that they are in many cases equally or more expensive than computing all
other instruments. Another expected feature of the GPU-to-CPU comparison is that
parallelism on the CPU is significantly less pronounced. Hence, we observe an
increased overhead for all quantities that contain non-linear transformations
and contractions of the high-dimensional individual gradients, or require
additional backpropagations (curvature).

\captionsetup[subfigure]{justification=justified,singlelinecheck=false}

\captionsetup[subfigure]{justification=centering, singlelinecheck=true}

\captionsetup[subfigure]{justification=justified,singlelinecheck=false}

\captionsetup[subfigure]{justification=centering, singlelinecheck=true}

\subsubsection{Configuration overhead}
\label{app:benchmark-configuration}

For the estimation of different \cockpit configuration overheads, we use
almost the same setting as described above, training for $512$ iterations and
tracking only every specified interval.

\begin{figure}[t]
	\centering
	\begin{subfigure}[t]{0.4\textwidth}
		\caption{\mnist Log.\,Reg.\,(GPU)}
		\includegraphics[width=\linewidth]{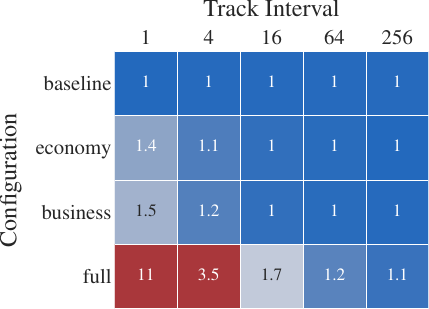}
		\label{fig:app_benchmark_configurations_cuda-mnist_logreg}
	\end{subfigure}
	\hspace{0.06\textwidth}
	\begin{subfigure}[t]{0.4\textwidth}
		\caption{\mnist \mlp\,(GPU)}
		\includegraphics[width=\linewidth]{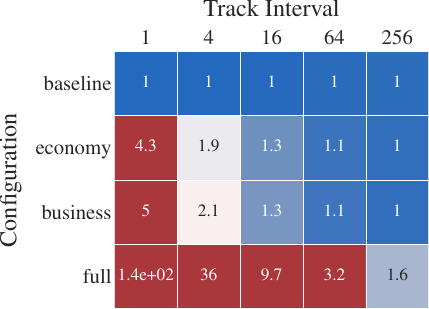}
		\label{fig:app_benchmark_configurations_cuda-mnist_mlp}
		\vspace{0.25cm}
	\end{subfigure}
	\begin{subfigure}[t]{0.4\textwidth}
		\caption{\cifarten \threecthreed\,(GPU)}
		\includegraphics[width=\linewidth]{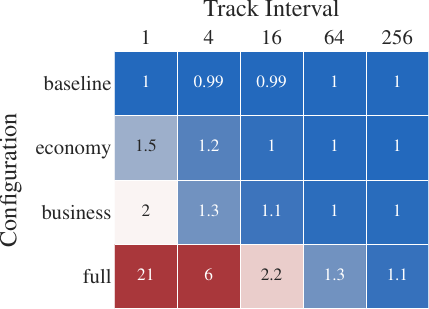}
		\label{fig:app_benchmark_configurations_cuda-cifar}
	\end{subfigure}
	\hspace{0.06\textwidth}
	\begin{subfigure}[t]{0.4\textwidth}
		\caption{\fmnist \twoctwod\,(GPU)}
		\includegraphics[width=\linewidth]{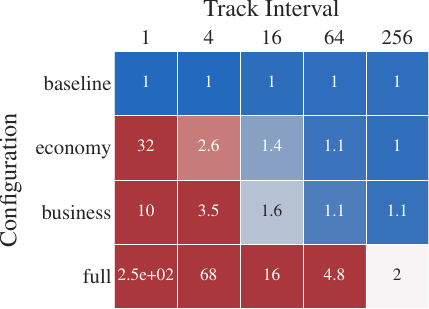}
		\label{fig:app_benchmark_configurations_cuda-fmnist}
	\end{subfigure}
  \caption{\textbf{Overhead of \cockpittitle configurations on GPU for four
      different problems with varying tracking interval.} Color bar is the same as in \autoref{fig:benchmark}.}
  \label{fig:app_benchmark_configurations_cuda}
\end{figure}

\begin{figure}[t]
	\centering
	\begin{subfigure}[t]{0.4\textwidth}
		\caption{\mnist Log.\,Reg.\,(CPU)}
		\includegraphics[width=\linewidth]{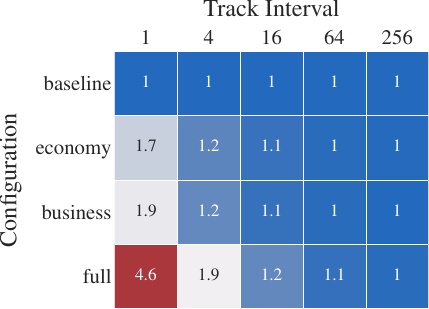}
		\label{fig:app_benchmark_configurations_cpu-mnist_logreg}
	\end{subfigure}
	\hspace{0.06\textwidth}
	\begin{subfigure}[t]{0.4\textwidth}
		\caption{\mnist \mlp\,(CPU)}
		\includegraphics[width=\linewidth]{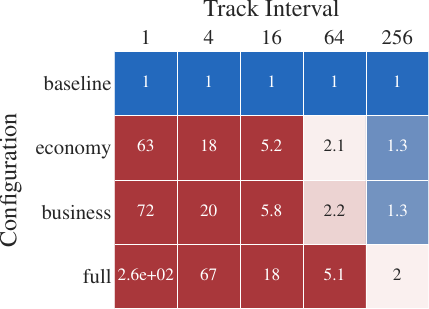}
		\label{fig:app_benchmark_configurations_cpu-mnist_mlp}
		\vspace{0.25cm}
	\end{subfigure}
	\begin{subfigure}[t]{0.4\textwidth}
		\caption{\cifarten \threecthreed\,(CPU)}
		\includegraphics[width=\linewidth]{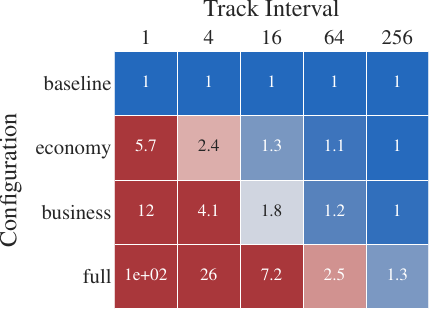}
		\label{fig:app_benchmark_configurations_cpu-cifar}
	\end{subfigure}
	\hspace{0.06\textwidth}
	\begin{subfigure}[t]{0.4\textwidth}
		\caption{\fmnist \twoctwod\,(GPU)}
		\includegraphics[width=\linewidth]{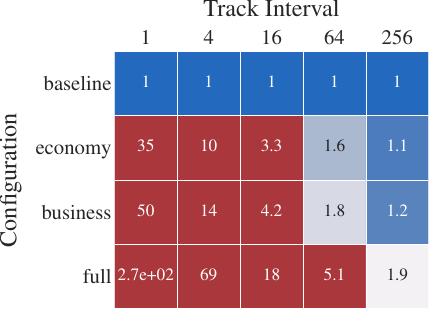}
		\label{fig:app_benchmark_configurations_cpu-fmnist}
	\end{subfigure}
	\caption{\textbf{Overhead of \cockpittitle configurations on CPU for four
			different problems with varying tracking interval.} Color bar is the same as in \autoref{fig:benchmark}.}
	\label{fig:app_benchmark_configurations_cpu}
\end{figure}

\paragraph{Configuration overhead on GPU versus CPU:}
\Cref{fig:app_benchmark_configurations_cuda} and
\Cref{fig:app_benchmark_configurations_cpu} show the configuration overhead for
four different \deepobs problems. The bottom left part of
\Cref{fig:app_benchmark_configurations_cuda} corresponds to
\Cref{fig:benchmark_heatmap}. In general, we observe that increased
parallelism can be exploited on a GPU, leading to smaller overheads in
comparison to a CPU.

\begin{figure}[t]
	\centering
	\includegraphics[width=0.59\linewidth]{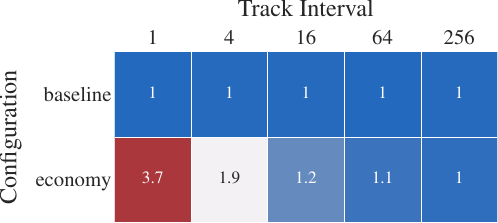}
	
	\caption{\textbf{Overhead of \cockpittitle configurations on GPU for \resnetfifty on
			\imagenet.} \cockpit's instruments scale efficiently even to very large problems
			(here: $1000$ classes, $(3, 224, 224)$-sized inputs, and a batch size of $64$.
			For individual gradients to be defined, we replaced the batch norm layers of the 
			\resnetfifty model with identities.) Color bar is the same as in \autoref{fig:benchmark}.}
	\label{fig:app_benchmark_configurations_gpu_imagenet}
\end{figure}

\cockpit can even scale to significantly larger problems, such as a \resnetfifty
on \imagenet-like data. \Cref{fig:app_benchmark_configurations_gpu_imagenet} shows
the computational overhead for different tracking intervals on such a large-scale
problem. Using the \textit{economy} configuration, we can achieve our self-imposed
goal of at most doubling the run time even when tracking every fourth step.
More extensive configurations (such as the \textit{full} set) would indeed have
almost prohibitively large costs associated. However, these costs could be dramatically
reduced when one decides to only inspect a part of the network using \cockpit.
Note, individual gradients are not properly defined when using batch norm, therefore,
we replaced these batch norm layers with identity layers when using the \resnetfifty.

\captionsetup[subfigure]{justification=justified,singlelinecheck=false}

\captionsetup[subfigure]{justification=centering, singlelinecheck=true}

\captionsetup[subfigure]{justification=justified,singlelinecheck=false}

\captionsetup[subfigure]{justification=centering, singlelinecheck=true}

 \clearpage
\subsection{Performance of two-dimensional histograms:}
\label{app:histograms}

Both one- and two-dimensional histograms require $|\gB| \times D$ elements be
accessed, and hence perform similarly. However, we observed different behavior
on GPU and decided to omit the two-dimensional histogram's run time in the main
text. As explained here, this performance lack is not fundamental, but a
shortcoming of the GPU implementation. \pytorch provides built-in functionality
for computing one-dimensional histograms at the time of writing, but is not yet
featuring multi-dimensional histograms. We experimented with three implementations:
\begin{itemize}
\item \textbf{\pytorch (third party):} A third party
  implementation\footnote{Permission granted by the authors of \href{https://github.com/miranov25/RootInteractive/blob/7019e4c2b9f291551aeeb8677a969cfcfde690d1/RootInteractive/Tools/Histograms/histogramdd_pytorch.py}{\texttt{github.com/miranov25/.../histogramdd\_pytorch.py}}.}
  under review for being integrated into \pytorch\footnote{See
    \url{https://github.com/pytorch/pytorch/pull/44485}.}. It relies on
  \texttt{torch.bincount},
  which uses \texttt{atomicAdd}s that represent a bottleneck for histograms
  where most counts are contained in one bin.\footnote{See \mbox{\url{https://discuss.pytorch.org/t/torch-bincount-1000x-slower-on-cuda/42654}}}
  This occurs often for over-parameterized deep models, as most of the
  gradient elements are zero.

\item \textbf{\pytorch (\cockpittitle):} Our implementation uses a
  suggested workaround, computes bin indices and scatters the counts into their
  associated bins with \texttt{torch.Tensor.put\_}. This circumvents
  \texttt{atomicAdd}s, but has poor memory locality.

\item \textbf{\numpy:} The single-threaded \texttt{numpy.histogram2d} serves as
  baseline, but does not run on GPUs.
\end{itemize}

\pgfkeys{/pgfplots/histogrambenchmarkdefault/.style={
    enlarge x limits=-0.05,
    width=1.0\linewidth,
    height=0.7\linewidth,
    every axis plot/.append style={line width = 1.5pt},
    tick pos = left,
    ylabel near ticks,
    xlabel near ticks,
    xtick align = inside,
    ytick align = inside,
    legend cell align = left,
    legend columns = 1,
    legend style = {
      fill opacity = 0.7,
      text opacity = 1,
      font = \small,
    },
    xticklabel style = {font = \small, inner xsep = -5ex},
    xlabel style = {font = \small},
    axis line style = {black},
    yticklabel style = {font = \small, inner ysep = -4ex},
    ylabel style = {font = \small},
    title style = {font = \small, inner ysep = -3ex},
    grid = major,
    grid style = {dashed}
  }
}

\begin{figure}
	\begin{subfigure}[t]{0.49\textwidth}
		\pgfkeys{/pgfplots/zmystyle/.style={histogrambenchmarkdefault,
				xlabel={Histogram Balance $b$}
			}}
		\caption{\qquad GPU}
\begin{tikzpicture}

\definecolor{color0}{rgb}{0.12156862745098,0.466666666666667,0.705882352941177}
\definecolor{color1}{rgb}{1,0.498039215686275,0.0549019607843137}

\begin{axis}[
axis line style={white!80!black},
legend style={fill opacity=0.8, draw opacity=1, text opacity=1, draw=white!80!black},
tick pos=left,
xlabel={Histogram Balance},
xmin=-0.039, xmax=1.039,
ylabel={Run Time [s]},
ymin=0.0647793941403482, ymax=344.025767453251,
ymode=log,
zmystyle
]
\addplot [, color0, dashed]
table {%
0.01 4.54533004760742
0.0615789473684211 2.08099224567413
0.113157894736842 0.744264936447144
0.164736842105263 0.445815587043762
0.216315789473684 0.335660696029663
0.267894736842105 0.282471227645874
0.319473684210526 0.253187036514282
0.371052631578947 0.232630848884583
0.422631578947368 0.219761872291565
0.474210526315789 0.210704207420349
0.52578947368421 0.206429862976074
0.577368421052632 0.218602991104126
0.628947368421053 0.214805579185486
0.680526315789474 0.20776104927063
0.732105263157895 0.20455629825592
0.783684210526316 0.20727813243866
0.835263157894737 0.206504940986633
0.886842105263158 0.203398704528809
0.9 0.20402295589447
0.938421052631579 0.204293870925903
0.99 0.204171013832092
};
\addlegendentry{\textsc{PyTorch} (\textsc{Cockpit})}
\addplot [, color1, dashed]
table {%
0.01 232.951647591591
0.0615789473684211 67.0944833517075
0.113157894736842 12.7715488433838
0.164736842105263 5.53443562984467
0.216315789473684 2.58965411186218
0.267894736842105 1.4656699180603
0.319473684210526 0.907050752639771
0.371052631578947 0.633884406089783
0.422631578947368 0.44161102771759
0.474210526315789 0.332058143615723
0.52578947368421 0.243267822265625
0.577368421052632 0.193850588798523
0.628947368421053 0.160495829582214
0.680526315789474 0.137686920166016
0.732105263157895 0.118721318244934
0.783684210526316 0.105425000190735
0.835263157894737 0.106805515289307
0.886842105263158 0.108340263366699
0.9 0.105959153175354
0.938421052631579 0.103169775009155
0.99 0.0956669807434082
};
\addlegendentry{\textsc{PyTorch} (third party)}
\end{axis}

\end{tikzpicture}
 		\label{fig:app-histogram2d-benchmark-gpu}
	\end{subfigure}
	\hfill
	\begin{subfigure}[t]{0.49\textwidth}
		\pgfkeys{/pgfplots/zmystyle/.style={histogrambenchmarkdefault,
				xlabel={Histogram Balance $b$}
		}}
		\caption{\qquad CPU}
\begin{tikzpicture}

\definecolor{color0}{rgb}{0.12156862745098,0.466666666666667,0.705882352941177}
\definecolor{color1}{rgb}{1,0.498039215686275,0.0549019607843137}
\definecolor{color2}{rgb}{0.172549019607843,0.627450980392157,0.172549019607843}

\begin{axis}[
axis line style={white!80!black},
legend style={fill opacity=0.8, draw opacity=1, text opacity=1, at={(0.91,0.5)}, anchor=east, draw=white!80!black},
tick pos=left,
xlabel={Histogram Balance},
xmin=-0.039, xmax=1.039,
ylabel={Run Time [s]},
ymin=0.919387552518064, ymax=8.28341609861724,
ymode=log,
zmystyle
]
\addplot [, color0, dashed]
table {%
0.01 1.04204399585724
0.0615789473684211 1.03900332450867
0.113157894736842 1.07859117984772
0.164736842105263 1.06876609325409
0.216315789473684 1.04880547523499
0.267894736842105 1.04529702663422
0.319473684210526 1.05313646793365
0.371052631578947 1.06466472148895
0.422631578947368 1.06266655921936
0.474210526315789 1.05389556884766
0.52578947368421 1.06146202087402
0.577368421052632 1.04773416519165
0.628947368421053 1.06101129055023
0.680526315789474 1.04648265838623
0.732105263157895 1.0483683347702
0.783684210526316 1.04395830631256
0.835263157894737 1.04497628211975
0.886842105263158 1.04231572151184
0.9 1.04766194820404
0.938421052631579 1.04687905311584
0.99 1.0437038898468
};
\addlegendentry{\textsc{PyTorch} (\textsc{Cockpit})}
\addplot [, color1, dashed]
table {%
0.01 1.02336490154266
0.0615789473684211 1.02175369262695
0.113157894736842 1.02237765789032
0.164736842105263 1.02118575572968
0.216315789473684 1.01996810436249
0.267894736842105 1.02348549365997
0.319473684210526 1.0248743057251
0.371052631578947 1.02083847522736
0.422631578947368 1.01863925457001
0.474210526315789 1.02599215507507
0.52578947368421 1.0236154794693
0.577368421052632 1.02260437011719
0.628947368421053 1.03915128707886
0.680526315789474 1.03184995651245
0.732105263157895 1.0255656003952
0.783684210526316 1.0171329498291
0.835263157894737 1.02414612770081
0.886842105263158 1.02574753761292
0.9 1.01600201129913
0.938421052631579 1.01842267513275
0.99 1.02955875396729
};
\addlegendentry{\textsc{PyTorch} (third party)}
\addplot [, color2, dashed]
table {%
0.01 3.44840505123138
0.0615789473684211 3.95587799549103
0.113157894736842 4.67446537017822
0.164736842105263 4.98263554573059
0.216315789473684 5.3127799987793
0.267894736842105 5.58865821361542
0.319473684210526 5.89873046875
0.371052631578947 6.09700152873993
0.422631578947368 6.21813678741455
0.474210526315789 6.48546359539032
0.52578947368421 6.69456965923309
0.577368421052632 6.80199146270752
0.628947368421053 6.89959270954132
0.680526315789474 7.07993612289429
0.732105263157895 7.23664236068726
0.783684210526316 7.27940900325775
0.835263157894737 7.2930465221405
0.886842105263158 7.35840466022491
0.938421052631579 7.44740259647369
0.99 7.4957230091095
};
\addlegendentry{\textsc{NumPy} (single thread)}
\end{axis}

\end{tikzpicture}
 		\label{fig:app-histogram2d-benchmark-cpu}
	\end{subfigure}
  \caption{\textbf{Performance of two-dimensional histogram GPU implementations
      depends on the data.} (\subref{fig:app-histogram2d-benchmark-gpu}) 
    Run time for two different GPU
    implementations with histograms of different imbalance. \cockpit's
    implementation outperforms the third party solution by more than one order
    of magnitude in the deep learning regime ($b \ll 1$). 
    (\subref{fig:app-histogram2d-benchmark-cpu}) On CPU,
    performance is robust to histogram balance. The run time difference between
    \numpy and \pytorch is due to multi-threading. Data has the same size as
    \deepobs's \cifarten \threecthreed problem ($D =895,210, |\gB| = 128$).
    Curves represent averages over 10 independent runs. Error bars are omitted
    to improve legibility.}
  \label{fig:app-histogram2d-benchmark}
\end{figure}

To demonstrate the strong performance dependence on the data, we generate data
from a uniform distribution over $[0, b]\times[0, b]$, where $b \in (0, 1)$ parametrizes the
histogram's balance, and compute two-dimensional histograms on $[0,1]\times [0,
1]$. \Cref{fig:app-histogram2d-benchmark} (a) shows a clear increase in
run time of both GPU implementations for more imbalanced histograms. Note that
even though our implementation outperforms the third party by more than one
order of magnitude in the deep neural network regime ($b \ll 1$), it is still
considerably slower than a one-dimensional histogram (see
\Cref{fig:app_benchmark_instruments_cuda} (c)), and even slower on GPU than on CPU
(\Cref{fig:app-histogram2d-benchmark} (b)). As expected, the CPU implementations
do not significantly depend on the data (\Cref{fig:app-histogram2d-benchmark}
(b)). The performance difference between \pytorch and \numpy is likely due to
multi-threading versus single-threading.

Although a carefully engineered histogram GPU implementation is currently not
available, we think it will reduce the computational overhead to that of
a one-dimensional histogram in future releases.

\clearpage

\section{\cockpittitle view of convex stochastic problems}
\label{app:convex-problems}

\begin{figure*}[h]
  \centering
  \Cshadowbox{\includegraphics[width=.97\textwidth, trim={7cm 2.5cm 5cm
      0.5cm}, clip]{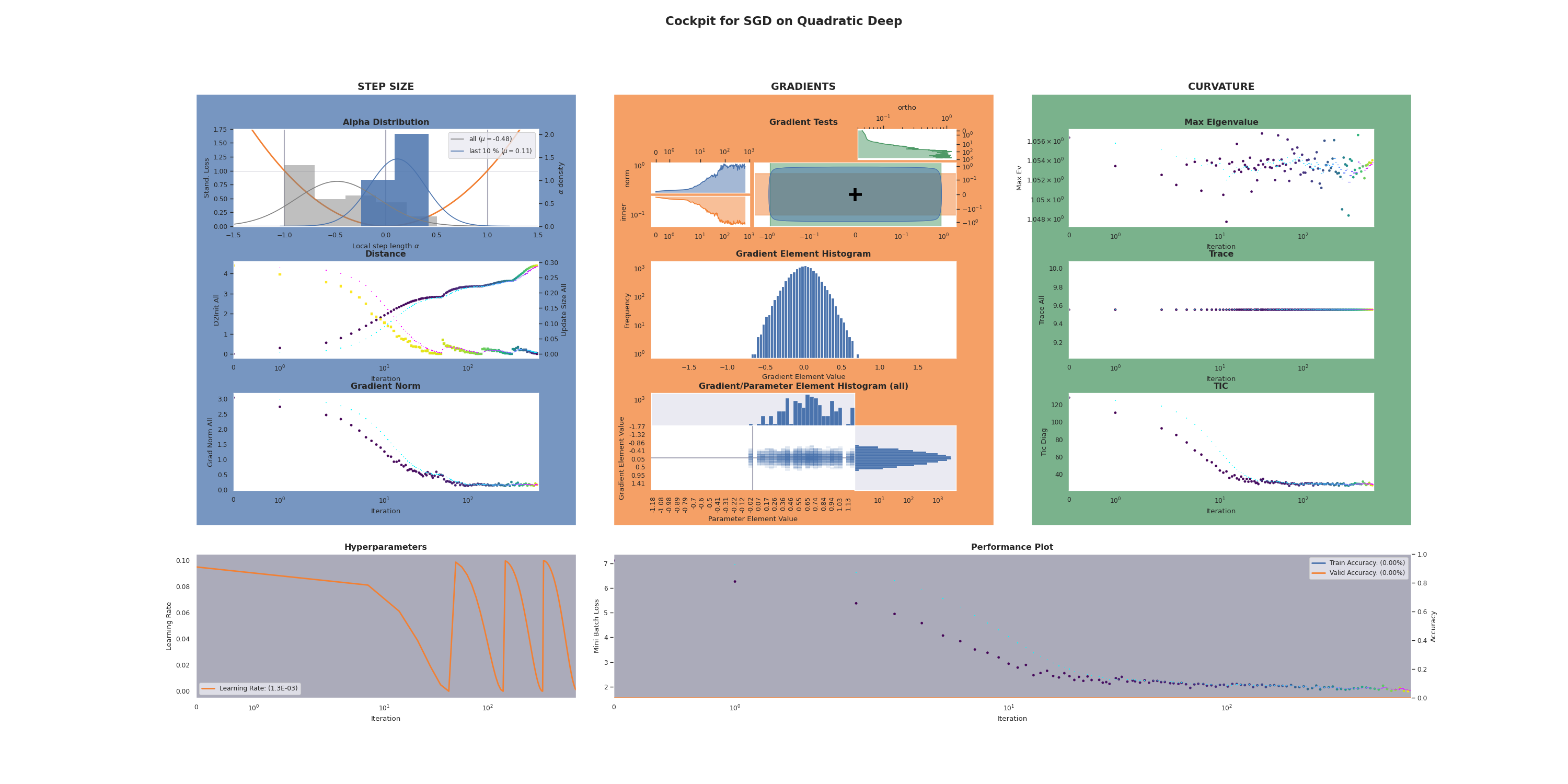}}

  \Cshadowbox{\includegraphics[width=.97\textwidth, trim={7cm 2.5cm 5cm
      0.5cm}, clip]{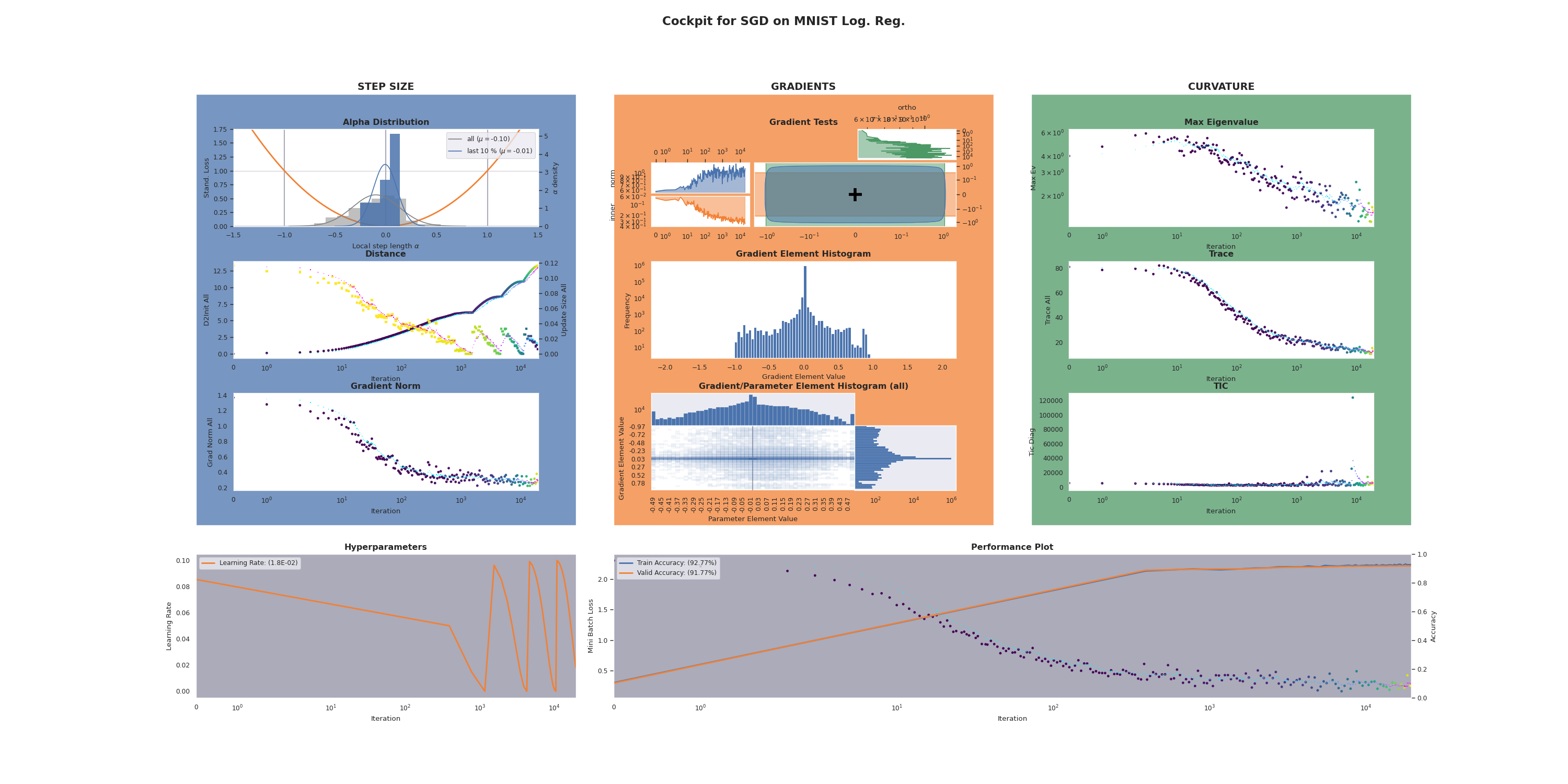}}

  \vspace{0.5\baselineskip}

  \caption{\textbf{Screenshot of \cockpittitle's full view for convex \deepobs
      problems.} Top \cockpit shows training on a noisy quadratic loss function.
    Bottom shows training on logistic regression on \mnist . Figure and labels are
    not meant to be legible. It is evident, that there is a fundamental difference
    in the optimization process, compared to training deep networks, \ie
    \Cref{fig:showcase}. This is, for example, visible when comparing the gradient
    norms, which converge to zero for convex problems but not for deep learning.}
  \label{fig:convex-problems}
\end{figure*}

\end{document}